\documentclass[nowfntplain,biber]{nowfnt} 

\usepackage[utf8]{inputenc}
\makeatletter
\DeclareOldFontCommand{\rm}{\normalfont\rmfamily}{\mathrm}
\DeclareOldFontCommand{\sf}{\normalfont\sffamily}{\mathsf}
\DeclareOldFontCommand{\tt}{\normalfont\ttfamily}{\mathtt}
\DeclareOldFontCommand{\bf}{\normalfont\bfseries}{\mathbf}
\DeclareOldFontCommand{\it}{\normalfont\itshape}{\mathit}
\DeclareOldFontCommand{\sl}{\normalfont\slshape}{\@nomath\sl}
\DeclareOldFontCommand{\sc}{\normalfont\scshape}{\@nomath\sc}
\makeatother

\usepackage{arabtex}
\renewcommand{\R}{\mathbb{R}}

\usepackage[round]{natbib}
\usepackage{amsmath}
\usepackage{amssymb}
\usepackage{mathtools}
\usepackage{hyperref} 
\usepackage{mathpazo} 
\usepackage[T1]{fontenc}
\usepackage{titlesec} 
\usepackage{epigraph}
\usepackage{blindtext} 
\usepackage[dvipsnames]{xcolor}
\usepackage{tikz}
\usepackage{dsfont}
\usepackage{pgfplots}
\usepackage{float}
\usepackage[capitalize,noabbrev]{cleveref}
\usepackage{bm}
\usepackage{csquotes}
\usepackage{stmaryrd}
\usepackage{wasysym}
\usepackage{amsthm}
\usepackage{mathtools}
\usepackage{paralist}
\usepackage{enumitem}
\usepackage[ruled,noend]{algorithm2e}
\usepackage{listings}
\usepackage[defaultsans,scale=0.95]{opensans} 
\hypersetup{
  linkcolor  = Black,
  citecolor  = RoyalBlue,
  urlcolor   = RoyalBlue,
  colorlinks = true,
}

\SetProcNameSty{textsc}
\SetProcArgSty{textsc}

\newcounter{example}[chapter]

\newenvironment{example}[1][]{\vspace{1em}\refstepcounter{example}\par\medskip
\noindent {\small{{\hlv Example}~{\hlvr\theexample}  #1}} \rmfamily}{\hfill\scriptsize\ensuremath{\blacksquare}\medskip\vspace{1em}}

\renewcommand*{\theexample}{\thechapter.\Alph{example}}

\newcounter{theorem}[chapter]

\renewcommand*{\thetheorem}{\thechapter.\Alph{theorem}}

\renewcommand{\vec}[1]{\bm{#1}}
\renewcommand{\geq}{\geqslant}
\renewcommand{\leq}{\leqslant}
\renewcommand{\iff}{\Leftrightarrow}

\DeclareMathOperator*{\argmin}{argmin}

\setlrmargins{*}{*}{1}
\checkandfixthelayout

\newcommand{\hlv}{\sffamily\bfseries}
\newcommand{\hlvr}{\normalfont\sffamily}

\renewcommand\printtoctitle[1]{\Huge\hlv Table of Contents}

\cftpagenumbersoff{part} 

\nobibintoc

\makepagestyle{headings}

\makeatletter      
\makeevenhead{headings}{\footnotesize\hlvr\thepage}{}{\footnotesize\hlvr\MakeUppercase{\leftmark}}
\makeoddhead{headings}{\footnotesize\hlvr\MakeUppercase{\rightmark}}{}{\footnotesize\hlvr\thepage}

\makeatother

\makepagestyle{chapterstyle}

\makeatletter      
\makeevenhead{chapterstyle}{}{}{}
\makeoddhead{chapterstyle}{}{}{\footnotesize\hlvr\thepage}

\makeatother   %

\aliaspagestyle{chapter}{chapterstyle}
\aliaspagestyle{part}{empty} 
\aliaspagestyle{cft}{headings}

\titleformat{\chapter}[display]
{\hlv\huge}
{\hlvr\Large\chaptertitlename~\thechapter}
{0ex}
{
}
[\vspace{0ex}]

\titleformat{\section}[hang]{}{\large\hlvr\thesection}{0.5em}{\large\hlvr\bfseries}

\titleformat{\subsection}[hang]{}{}{0pt}{\hlv}

\titlespacing*{\section}
{0pt}{5.5ex plus 1ex minus .2ex}{2.3ex plus .2ex}
\titlespacing*{\subsection}
{0pt}{3ex plus 1ex minus .2ex}{1.5ex plus .2ex}

\makeatletter
\renewcommand\paragraph{
  \@startsection{paragraph}
                {4}
                {\z@}
                {2ex \@plus1ex \@minus.2ex}
                {-1em}
                {\hlvr\small\bfseries}}
\makeatother

\captionnamefont{\small\hlv}
\captiontitlefont{\small\hlvr}
\captiondelim{\hspace{0.5em}}

\epigraphfontsize{\small}
\newcommand\blfootnote[1]{%
  \begingroup
  \renewcommand\thefootnote{}\footnote{#1}%
  \addtocounter{footnote}{-1}%
  \endgroup
}

\title{Introduction to Neural Network Verification}
\author{Aws Albarghouthi}

\usetikzlibrary{calc,arrows,backgrounds,decorations.markings}
\usepgflibrary{shapes.multipart}
\usetikzlibrary{patterns}

\makeatletter
\newcommand{\pgfplotsdrawaxis}{\pgfplots@draw@axis}
\makeatother

\pgfplotsset{only axis on top/.style={axis on top=false, after end axis/.code={
             \pgfplotsset{axis line style=opaque, ticklabel style=opaque, tick style={thick,opaque},
                          grid=none}\pgfplotsdrawaxis}}}

\tikzset{>=latex} 

\tikzstyle{oper}=[circle, draw=black, thick, minimum size = 8mm]
\tikzstyle{input}=[rounded corners, draw=Maroon, thick, minimum size = 6mm]
\tikzstyle{output}=[rounded corners, thick, draw=RoyalBlue, minimum size = 6mm]
\tikzstyle{empty}=[minimum size = 4mm]

\setlist[description]{
    labelindent=.5cm,
    style=unboxed,
    leftmargin=.5cm,
    font=\bfseries,
}

\Crefname{algocf}{Algorithm}{Algorithms} 
\newcommand{\acomment}[1]{ {\color{MidnightBlue} $\triangleright$ #1}}

\lstdefinestyle{customc}{
  belowcaptionskip=1\baselineskip,
  breaklines=true,
  xleftmargin=\parindent,
  language=C,
  showstringspaces=false,
  basicstyle=\ttfamily,
  keywordstyle=\bfseries\ttfamily\color{MidnightBlue},
  commentstyle=\itshape\color{black},
  identifierstyle=\ttfamily\color{black},
  stringstyle=\itshape\color{blue},
  keywords={ map,  while, flatMap, reduce, then, in, return,  for, if, else, reduceByKey, filter, partition},
}

\newcommand{\node}{v}
\newcommand{\nodes}{V}
\newcommand{\inodes}{V^\textsf{in}}
\newcommand{\onodes}{V^\textsf{o}}
\newcommand{\edges}{E}

\newcommand{\outs}{\textsf{out}}

\newcommand{\relu}{\textrm{relu}}

\newcommand{\mlp}{\textsc{mlp}\xspace}
\newcommand{\cnn}{\textsc{cnn}\xspace}
\newcommand{\rnn}{\textsc{rnn}\xspace}
\newcommand{\dagg}{\textsc{dag}\xspace}
\newcommand{\cpu}{\textsc{cpu}\xspace}

\newcommand{\pre}[1]{\color{Blue}{\{ ~ #1 ~ \}}}
\newcommand{\post}[1]{\color{Blue}{\{ ~ #1 ~ \}}}
\newcommand{\class}{\mathsf{class}}
\newcommand{\true}{\mathsf{true}}
\newcommand{\false}{\mathsf{false}}

\renewcommand{\varphi}{F}
\newcommand{\fv}{\mathit{fv}}
\newcommand{\eval}{\mathsf{eval}}
\newcommand{\sat}{\textsc{sat}\xspace}
\newcommand{\unsat}{\textsc{unsat}\xspace}
\newcommand{\lra}{\textsc{lra}\xspace}
\newcommand{\milp}{\textsc{milp}\xspace}
\newcommand{\smt}{\textsc{smt}\xspace}
\newcommand{\cnf}{\textsc{cnf}\xspace}
\newcommand{\fol}{\textsc{fol}\xspace}

\newcommand{\iorel}{R}
\newcommand{\inv}[2]{{#1}^{\textsf{in},#2}}
\newcommand{\outv}[1]{{#1}^\textsf{o}}
\newcommand{\nodesf}{\varphi_\nodes}
\newcommand{\edgesf}[1]{\varphi_{\circ\rightarrow#1}}
\newcommand{\edgesfall}{\varphi_{\edges}}
\newcommand{\graphf}[1]{\varphi_{#1}}

\newcommand{\dpll}{\textsc{dpll}\xspace}
\newcommand{\dpllt}{\textsc{dpll}$^T$\xspace}
\newcommand{\bcp}{\textsc{bcp}\xspace}

\newcommand{\lb}{l}
\newcommand{\ub}{u}

\newcommand{\ps}[1]{\mathcal{P}(#1)}
\newcommand{\setf}[1]{#1^s}

\newcommand{\xset}{X}

\newcommand{\mse}{\textsc{mse}\xspace}
\newcommand{\sgd}{\textsc{sgd}\xspace}

\newcommand{\gpu}{\textsc{gpu}}
\newcommand\norm[1]{\left\lVert#1\right\rVert}

\newcommand{\R}{\mathbb{R}}

\title{Introduction to Neural Network Verification}

\maintitleauthorlist{
Aws Albarghouthi
}

\author{Albarghouthi, Aws}
\affil{University of Wisconsin--Madison; aws@cs.wisc.edu}

\begin{document}

\makeabstracttitle

\begin{abstract}
    Deep learning has transformed the way we think of software and what it can do. But deep neural networks are fragile and their behaviors are often surprising. In many settings, we need to provide formal guarantees on the safety, security, correctness, or robustness of neural networks. This book covers foundational ideas from formal verification and their adaptation to reasoning about neural networks and deep learning.
\end{abstract}

\vfill
\noindent
The author's name in native alphabet is\\
\includegraphics[scale=.07]{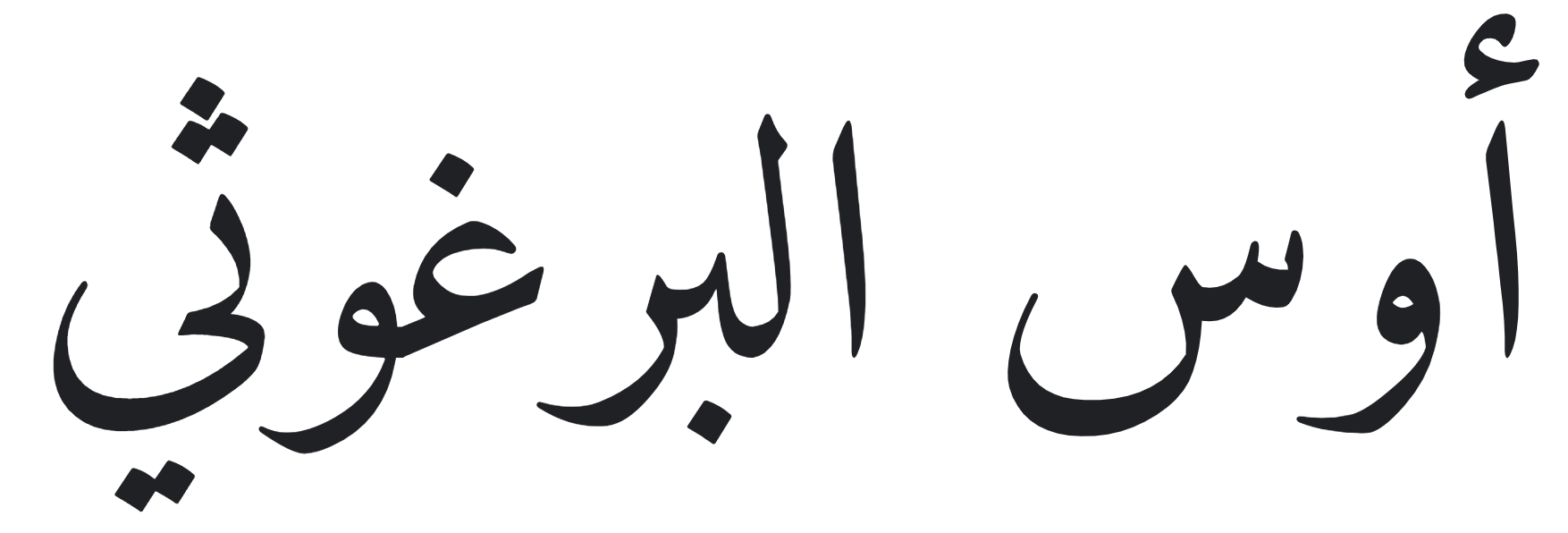}

\begin{acknowledgements}
Thanks to the best focus group ever: the CS 839 students and TA, Swati Anand, at the University of Wisconsin--Madison.
A number of insightful people sent me comments that radically improved the presentation: Frantisek Plasil, Georg Weissenbacher, Sayan Mitra, Benedikt B\"oing, Vivek Garg, 
Guy Van den Broeck, Matt Fredrikson, in addition to some anonymous folks.

\end{acknowledgements}

\chapter*{About This Book}

\section*{Why This Book?}
Over the past decade, a number of hardware and software advances
have conspired to thrust deep learning and neural networks to the forefront
of computing.
Deep learning has created a qualitative shift in our conception 
of what software is and what it can do: Every day we're seeing new applications of deep learning,
from healthcare to art, and it feels like we're only scratching the surface of a universe of new possibilities.

It is thus safe to say that deep learning is here to stay, in one form or another.
The line between software 1.0 (that is, manually written code) and software 2.0 (learned neural networks) is getting fuzzier and fuzzier,
and  neural networks are participating in safety-critical, security-critical, and socially critical tasks.
Think, for example, healthcare, self-driving cars, malware detection, etc.
But neural networks are fragile and so we need to prove that they are well-behaved when applied in critical settings.

Over the past few decades, the formal methods community has developed a plethora of techniques for automatically proving properties of programs, and, well, neural networks are programs.
So there is a great opportunity to port verification ideas to the software 2.0 setting.
This book offers the first introduction of foundational ideas from automated verification as applied to deep neural networks and deep learning.
I hope that it will inspire verification researchers to explore correctness in deep learning and deep learning researchers to adopt verification technologies.

\section*{Who Is This Book For?}
Given that the book's subject matter sits at the intersection of two pretty much disparate areas of computer science, one of my main design goals was to make it as self-contained as possible.
This way the book can serve as an introduction to the field for first-year graduate students or senior undergraduates, even if they have not been exposed to deep learning or verification.
For a comprehensive survey of verification algorithms for neural networks, along with implementations, I direct the reader to \cite{DBLP:journals/ftopt/LiuALSBK21}.

\section*{What Does This Book Cover?}
The book is divided into three parts:
\begin{description}
    \item[Part 1] defines neural networks as data-flow graphs of operators over real-valued inputs. This formulation will serve as our basis for the rest of the book. Additionally, we will survey a number of correctness properties that are desirable of neural networks and place them in a formal framework.
    \item[Part 2] discusses \emph{constraint-based} techniques for verification. As the name suggests, we construct a system of constraints and solve it to prove (or disprove) that a neural network satisfies some properties of interest. Constraint-based verification techniques are also referred to as \emph{complete verification}  in the literature.
    \item[Part 3] discusses \emph{abstraction-based} techniques for verification.
    Instead of executing a neural network on a single input, we can actually execute it on an \emph{infinite} set and show that all of those inputs satisfy desirable correctness properties. 
    Abstraction-based techniques are also referred to as \emph{approximate verification}  in the literature. 

\end{description}
Parts 2 and 3 are disjoint; the reader may go directly from Part 1 to Part 3 without losing context.


\part{Neural Networks \& Correctness}
  
    \chapter{A New Beginning}\label{ch:beginning}

\epigraph{He had become so caught up in building sentences that he had almost forgotten the barbaric days when thinking was like a splash of color landing on a page.}{---Edward St. Aubyn, \emph{Mother's Milk}}\blfootnote{Quote found in William Finnegan's \emph{Barbarian Days}.}

\section{It Starts With Turing}
This book is about \emph{verifying} that a \emph{neural network} behaves according to some set of desirable properties. These fields of study, verification and neural networks, have been two distinct areas of computing research with almost no bridges connecting them, until very recently. Intriguingly, however, both fields trace their genesis to a two-year period of Alan Turing's tragically short life.

In 1949, Turing wrote a little-known paper titled \emph{Checking a Large Routine} \citep{turing1989checking}.
It was a truly forward-looking piece of work. In it, Turing asks how can we prove that the programs we write do what they are supposed to do? Then, he proceeds to provide a proof of correctness of a program implementing the factorial function. 
Specifically, Turing proved that his little piece of code always terminates and always produces the factorial of its input.
The proof is elegant; it breaks down the program into single instructions, proves a lemma for every instruction, and finally stitches the lemmas together to prove correctness of the full program.  
Until this day, proofs of programs very much follow Turing's proof style from 1949. And, as we shall see in this book, proofs of neural networks will, too. 

Just a year before Turing's proof of correctness of factorial, in 1948, Turing wrote a perhaps even more farsighted paper, \emph{Intelligent Machinery}, in which he proposed \emph{unorganized machines}.\footnote{\emph{Intelligent Machinery} is reprinted in \citet{turing1948intelligent}.} These machines, Turing argued, mimic the infant human cortex, and he showed how they can \emph{learn} using what we now call a genetic algorithm.
Unorganized machines are a very simple form of what we now know as neural networks.

\section{The Rise of Deep Learning}
The topic of training neural networks continued to be studied since Turing's 1948 paper. 
But it has only exploded in popularity over the past decade, thanks to a combination algorithmic insights, hardware developments, and a flood of data for training.

Modern neural networks are called \emph{deep} neural networks, and the approach to training these neural networks is \emph{deep learning}. 
Deep learning has enabled incredible improvements in complex computing tasks, most notably in computer vision and natural-language processing, for example, in recognizing objects and people in an image and translating between languages. 
Everyday, a growing research community is exploring ways to extend and apply deep learning to more challenging problems, from music generation to proving mathematical theorems and beyond.

The advances in deep learning have changed the way we think of what software is, what it can do, and how we build it. Modern software is increasingly becoming a menagerie of traditional, manually written code and automatically trained---sometimes constantly learning---neural networks.
But deep neural networks can be fragile and produce unexpected results.
As deep learning becomes used more and more in sensitive settings, like autonomous cars, it is imperative that we verify these systems and provide formal guarantees on their behavior.
Luckily, we have decades of research on program verification that we can build upon,
but what exactly do we verify?

\section{What do We Expect of Neural Networks?}
In Turing's proof of correctness of his factorial program, 
Turing was concerned that we will be programming computers to perform mathematical operations, but we could be getting them wrong.
So in his proof he showed that his implementation of factorial is indeed equivalent to the mathematical definition. This notion of program correctness is known as \emph{functional correctness}, meaning that a program is a faithful implementation of some mathematical function.
Functional correctness is incredibly important in many settings---think of the disastrous effects of a buggy implementation of a cryptographic primitive
or an aircraft controller.

In the land of deep learning, proving functional correctness is an unrealistic task.
What does it mean to correctly recognize cats in an image or correctly translate English to Hindi? 
We cannot mathematically define such tasks.
The whole point of using deep learning to do tasks like translation or image recognition is because we cannot mathematically 
capture what exactly they entail. 

So what now? Is verification out of the question for deep neural networks?
No! While we cannot precisely capture what a deep neural network should do, we can often characterize some of its desirable or undesirable properties. Let's look at some examples of such properties.

\subsection*{Robustness}
The most-studied correctness property of neural networks is \emph{robustness}, because it is generic in nature and deep learning models are infamous for their fragility \citep{SzegedyZSBEGF13}.
Robustness means that small perturbations to inputs should not result in changes to the output of the neural network. For example, changing a small number of pixels in my photo should not make the network think that I am a cupboard instead of a person,
or adding inaudible noise to a  recording of my lecture should not make the network think it is a lecture about the Ming dynasty in the 15th century.
Funny examples aside, lack of robustness can be a safety and security risk.
Take, for instance, an autonomous vehicle following traffic signs using cameras. It has been shown that a light touch of vandalism to a stop sign can cause the vehicle to miss it, potentially causing an accident \citep{eykholt2018robust}.
Or consider the case of a neural network for detecting malware. We do not want a minor tweak to the malware's binary to cause the detector to suddenly deem it safe to install.

\subsection*{Safety}
Safety is a broad class of correctness properties stipulating that a program should not get to a \emph{bad state}. The definition of \emph{bad} depends on the task at hand. 
Consider a neural-network-operated robot working in some kind of plant. We might be interested in ensuring that the robot does not exceed certain speed limits, to avoid endangering human workers, or that it does not go to a dangerous part of the plant.
Another well-studied example is a neural network implementing a collision avoidance system for aircrafts \citep{katz2017reluplex}. One property of interest is that if an intruding aircraft is approaching from the left, the neural network should decide to turn the aircraft right.  

\subsection*{Consistency}
Neural networks learn about our world via examples, like images. As such, they may sometimes  miss basic axioms, like physical laws, and assumptions about realistic scenarios.
For instance, a neural network recognizing objects in an image and their relationships might say that object A is on top of object B, B is on top of C, and C is on top of A. But this cannot be! (At least not in the world as we know it.) 

For another example, consider a neural network tracking players on the soccer field using a camera. It should not in one frame of video say that Ronaldo is on the right side of the pitch and then in the next frame say that Ronaldo is on the left side of the pitch---Ronaldo is fast, yes, but he has slowed down in the last couple of seasons.

\section*{Looking Ahead}

I hope that I have convinced you of the importance of verifying properties of neural networks. 
In the next two chapters, we will formally define what neural networks look like (spoiler: they are ugly programs) and then build a language for formally specifying correctness properties of neural networks, paving the way for verification algorithms to prove these properties.

    \chapter{Neural Networks as Graphs}\label{ch:semantics}

There is no rigorous definition of what deep learning is and what it is not.
In fact, at the time of writing this, there is a raging debate in the artificial intelligence community about a clear definition.
In this chapter, we will define neural networks generically as graphs of operations over real numbers.
In practice, the shape of those graphs, called the \emph{architecture}, is not arbitrary: 
Researchers and practitioners carefully construct new architectures to suit various tasks.
For example, at the time of writing, neural networks for image recognition typically look different from those for natural language tasks.

First, we will informally introduce graphs and look at some popular architectures.
Then, we will formally define graphs and their semantics. 

\section{The Neural Building Blocks}\label{sec:blocks}
A neural network is a graph where each node performs an operation.
Overall, the graph represents a function from vectors of real numbers to vectors of real numbers, that is, 
a function in $\R^n \rightarrow \R^m$.
Consider the following very simple graph.
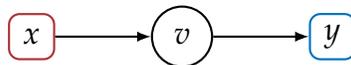
\begin{figure}[h!]
    \begin{center}
        \begin{tikzpicture}

            \draw node at (0, 0) [input] (in) {$x$};
            \draw node at (2, 0) [oper] (op) {$\node$};
            \draw node at (4, 0) [output] (out) {$y$};

            \draw[->,thick] (in) -- (op);
            \draw[->,thick] (op) -- (out);

        \end{tikzpicture}
        \caption{A very simple neural network}
        \label{fig:simple}
    \end{center}
\end{figure}

\noindent
The red node is an \emph{input} node; it just passes input $x$, a real number, to node $\node$.
Node $\node$ performs some operation on $x$ and spits out a value that goes to the \emph{output} node $y$.
For example, $\node$ might simply return $2x+1$, which we will denote as the function $f_\node:\R\to\R$:
$$f_\node(x) = 2x + 1$$
In our model, the output node may also perform some operation, for example, $$f_y(x) = \max(0,x)$$
Taken together, this simple graph encodes the following function $f : \R\to\R$:
$$f(x) =  f_y(f_\node(x))  = \max(0,2x+1)$$

\subsection*{Transformations and Activations}
\begin{figure}[h!]
\begin{center}
    \begin{tikzpicture}
    \draw[->] (-1,0) -- (2,0) node[right] {$x$};
    \draw[->] (0,0) -- (0,2) node[above] {$\textrm{relu}(x)$};
    \draw[scale=0.5,domain=-3:3,variable=\x,RoyalBlue,thick] plot ({\x},{max(0,\x)});
  \end{tikzpicture}
  \caption{Rectified linear unit}  \label{fig:relu}

\end{center}
\end{figure}
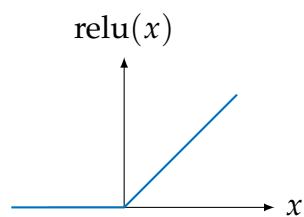

The function $f_\node$ in our example above is \emph{affine}:
simply, it multiplies inputs by constant values (in this case, $2x$) and adds constant values (in this case, $1$).
The function $f_y$ is an \emph{activation} function, because it turns \emph{on} or \emph{off} depending on its input. 
When its input is negative, $f_y$ outputs $0$ (off), 
otherwise it outputs its input (on).
Specifically, $f_y$, illustrated in \cref{fig:relu}, is called a \emph{rectified linear unit} (ReLU),
and it is a very popular activation function in modern deep neural networks~\citep{DBLP:conf/icml/NairH10}.
Activation functions are used to add non-linearity into a neural network.

\begin{figure}[t]
    \centering
    \begin{tikzpicture}
        \begin{axis}[
            axis lines=middle,
            xmax=5,
            xmin=-5,
            ymin=-0.05,
            ymax=1.2,
            ytick={.5},
            xtick={},
            width=5cm
        ]

        \addplot [domain=-10:10, samples=100,
                  thick, MidnightBlue] {1/(1+exp(-x)};
    
    \end{axis}
    \end{tikzpicture}
\caption{Sigmoid function}\label{fig:sigmoid}
\end{figure}
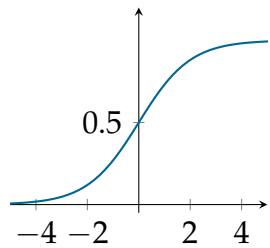

There are other popular activation functions, for example, sigmoid,
$$\sigma(x) = \frac{1}{1+\exp(-x)}$$
whose output is bounded between 0 and 1, as shown in \cref{fig:sigmoid}.

Often, in the literature and practice, the affine functions and the activation function are composed into a single operation.
Our graph model of neural networks can capture that, but we usually prefer to separate the two operations on to two different nodes of the graph, as it will simplify our life in later chapters when we start analyzing those graphs.

\subsection*{Universal Approximation}
What is so special about these activation functions?
The short answer is they work in practice, in that they result in neural networks that are able to learn complex tasks. 
It is also very interesting to point out that you can construct a neural network comprised of ReLUs or sigmoids and affine functions to approximate any continuous function. This is known as the \emph{universal approximation theorem}~\citep{DBLP:journals/nn/HornikSW89}, and in fact the result is way more general than ReLUs and sigmoids---nearly any activation function you can think of works, as long as it is not polynomial~\citep{DBLP:journals/nn/LeshnoLPS93}!
For an interactive illustration of universal approximation, I highly recommend
\citet[Ch.4]{nielsen}.

\section{Layers and Layers and Layers}\label{ssec:layers}
\begin{figure}[h]
    \begin{center}
        \begin{tikzpicture}            
            \foreach \x in {1,...,3}
                \draw node at (0, -\x*1.25) [input] (first_\x) {$x_\x$};

            \foreach \x in {1,...,3}
                \node at (4, -\x*1.25) [oper] (second_\x){$a_\x$};

        \foreach \x in {1,...,3}
                \node at (8, -\x*1.25) [output] (fourth_\x){$y_\x$};

            \foreach \i in {1,...,3}
                \foreach \j in {1,...,3}
                    \draw [->,thick] (first_\i) to (second_\j);

            \foreach \i in {1,...,3}
            \foreach \j in {1,...,3}
                    \draw [->,thick] (second_\i) to (fourth_\j);

        \end{tikzpicture}
        \caption{A multilayer perceptron}
        \label{fig:ffn}
    \end{center}
\end{figure}
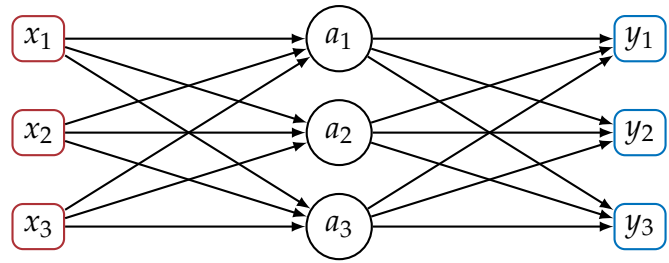

\begin{figure}[t]
    \begin{center}
        \begin{tikzpicture}            
            \foreach \x in {1,...,3}
                \draw node at (0, -\x*1.25) [input] (first_\x) {$x_\x$};

            \foreach \x in {1,...,3}
                \node at (4, -\x*1.25) [oper] (second_\x){$v_\x$};

            \foreach \x in {4,...,6}
                \node at (8, -\x*1.25 +3*1.25) [oper] (third_\x){$\node_\x$};

            \foreach \x in {1,...,3}
                \node at (12, -\x*1.25) [output] (fourth_\x){$y_\x$};

            \foreach \i in {1,...,3}
                \foreach \j in {1,...,3}
                    \draw [->,thick] (first_\i) to (second_\j);

            \foreach \i in {1,...,3}
            \foreach \j in {4,...,6}
                    \draw [->,thick] (second_\i) to (third_\j);

            \foreach \i in {4,...,6}
            \foreach \j in {1,...,3}
                    \draw [->,thick] (third_\i) to (fourth_\j);

        \end{tikzpicture}
        \caption{A multilayer perceptron with two hidden layers}
        \label{fig:mlp}
    \end{center}
\end{figure}
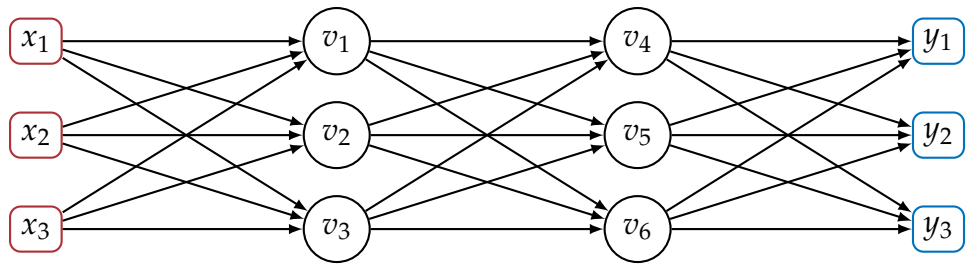

In general, a neural network can be a crazy graph, with nodes and arrows pointing all over the place. In practice, networks are usually \emph{layered}.
Take the graph in \cref{fig:ffn}.   
Here we have 3 inputs and 3 outputs, denoting a function in $\R^3 \to \R^3$.
Notice that the nodes of the graph form layers,
the input layer, the output layer, and the layer in the middle which is called the \emph{hidden} layer.
This form of graph---or architecture---has the grandiose name of \emph{multilayer perceptron} (\mlp).
Usually, we have a bunch of hidden layers in an \mlp;  \cref{fig:mlp} shows a \mlp with two hidden layers.
Layers in an \mlp are called \emph{fully connected} layers, since each node receives all outputs from the preceding layer.

Neural networks are typically used as \emph{classifiers}:
they take an input, e.g., pixels of an image, and predict
 what the image is about (the image's class).
When we are doing classification, the output layer of the \mlp represents the probability of each class, for example, $y_1$ is the probability of the input being a chair, $y_2$ is the probability of a TV, and $y_3$ of a couch.
To ensure that the probabilities are normalized, that is, between $0$ and $1$ and sum up to $1$, the final layer employs a \emph{softmax} function. 
Softmax, generically, looks like this for an output node $y_i$, where $n$ is the number of classes:
$$f_{y_i}(x_1,\ldots,x_n) = \frac{\exp(x_i)}{\sum_{k=1}^n \exp(x_k)}$$

Why does this work?
Imagine that we have two classes, i.e., $n=2$.
First, we can verify that $$f_{y_1}, f_{y_2} \in [0,1]$$
This is because the numerators and denominators are both positive,
and the numerator is $\leq$ than the denominator.
Second, we can see that $f_{y_1}(x_1,x_2) + f_{y_2}(x_1,x_2) = 1$,
because
$$f_{y_1}(x_1,x_2) + f_{y_2}(x_1,x_2) = \frac{e^{x_1}}{e^{x_1} + e^{x_2}}  + \frac{e^{x_2}}{e^{x_1} + e^{x_2}} = 1$$
Together, these two facts mean that we have a probability distribution.
For an interactive visualization of softmax,  
please see the excellent online book by \citet[Chapter 3]{nielsen}.

Given some outputs $(y_1,\ldots,y_n)$ of the neural network,
we will use $$\class(y_1,\ldots,y_n)$$ to denote the index 
of the largest element (we assume no ties), i.e., the class
with the largest probability.
For example, $\class(0.8, 0.2) = 1$, while $\class(0.3,0.7) = 2$.

\section{Convolutional Layers}

\begin{figure}[h]
    \begin{center}
        \begin{tikzpicture}            
            \foreach \x in {1,...,4}
                \draw node at (0, -\x*1) [input] (first_\x) {$x_\x$};

            \foreach \x in {1,...,3}
                \node at (4, -\x*1.25) [oper] (second_\x){$\node_\x$};

            \foreach \x in {1,...,3}
                \node at (8, -\x*1.25) [output] (fourth_\x){$y_\x$};

            \draw [->,thick] (first_1) to (second_1);
            \draw [->,thick] (first_2) to (second_1);

            \draw [->,thick] (first_2) to (second_2);
            \draw [->,thick] (first_3) to (second_2);
            
            \draw [->,thick] (first_3) to (second_3);
            \draw [->,thick] (first_4) to (second_3);

            \foreach \i in {1,...,3}
                    \draw [->,thick] (second_\i) to (fourth_\i);

        \end{tikzpicture}
        \caption{1-dimensional convolution}
        \label{fig:cnn}
    \end{center}
\end{figure}
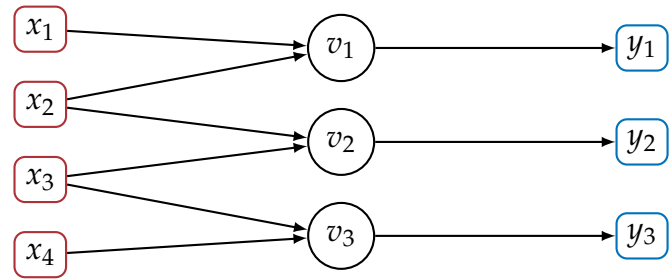

Another kind of layer that you will find in a neural network is a \emph{convolutional} layer.
This kind of layer is widely used in computer-vision tasks, but also has uses in natural-language processing.
The rough intuition is that if you are looking at an image, you want to scan it looking for patterns.
The convolutional layer gives you that: it defines an operation, a \emph{kernel}, that is applied to every region of pixels in an image or every sequence of words in a sentence.
For illustration, let's consider an input layer of size 4, perhaps each input defines a word in a 4-word sentence,
as shown in \cref{fig:cnn}.
Here we have a kernel, nodes $\{\node_1,\node_2,\node_3\}$, that is applied to every pair of consecutive words, $(x_1,x_2), (x_2,x_3),$ and  $(x_3,x_4)$.
We say that this kernel has size 2, since it takes an input in $\R^2$.
This kernel is 1-dimensional, since its input is a vector of real numbers.
In practice, we work with 2-dimensional kernels or more; 
for instance, to scan blocks of pixels of a gray-scale image where every pixel is a real number,
we can use kernels that are functions in $\R^{10\times 10} \to \R$, 
meaning that the kernel is applied to every $10\times 10$ sub-image in the input.

Typically, a \emph{convolutional neural network} (\cnn) will apply a bunch of kernels to an input---%
and many layers of them---and aggregate (\emph{pool}) the information from each kernel.
We will meet these operations in later chapters when we verify properties of such networks.\footnote{Note that there are many parameters that are used to construct a \cnn, e.g., how many kernels are applied, how many inputs a kernel applies to, the \emph{stride} or step size of a kernel, etc. These are not of interest to us in this book. We're primarily concerned with the core building blocks of the neural network, which will dictate the verification challenges.}

\section{Where are the Loops?}
All of the neural networks we have seen so far seem to be a composition of a number mathematical functions,
one after the other.
So what about loops? Can we have loops in neural networks?
In practice, neural network graphs are really just directed acyclic graphs (\dagg).
This makes training the neural network possible using the \emph{backpropagation} algorithm.

That said, there are popular classes of neural networks that appear to have loops, but they are very simple,
in the sense that the number of iterations of the loop is just the size of the input.
\emph{Recurrent neural networks} (\rnn) is the canonical class of such networks,
which are usually used for sequence data, like text.
You will often see the graph of an \rnn rendered as follows, with the self loop on node $\node$.
\begin{figure}[h]
    \begin{center}
        \begin{tikzpicture}
            \draw node at (0, 0) [input] (in) {$x$};
            \draw node at (0, 1.5) [oper] (op) {$v$};
            \draw node at (0, 3) [output] (out) {$y$};

            \draw[->,thick] (in) -- (op);
            \draw[->,thick] (op) -- (out);
            \draw[->,thick] (op)  to[in=135,out=230,loop] (op);
        \end{tikzpicture}
        \caption{Recurrent neural network}
        \label{fig:rnn}
    \end{center}
\end{figure}
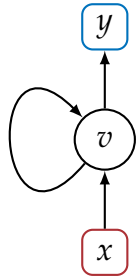

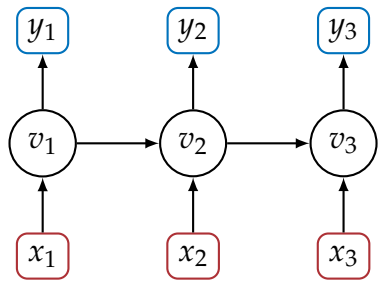
\begin{figure}[h]
    \begin{center}
        \begin{tikzpicture}

            \draw node at (0, 0) [input] (in) {$x_1$};
            \draw node at (0, 1.5) [oper] (op) {$v_1$};
            \draw node at (0, 3) [output] (out) {$y_1$};

            \draw node at (2, 0) [input] (in1) {$x_2$};
            \draw node at (2, 1.5) [oper] (op1) {$v_2$};
            \draw node at (2, 3) [output] (out1) {$y_2$};

            \draw node at (4, 0) [input] (in2) {$x_3$};
            \draw node at (4, 1.5) [oper] (op2) {$v_3$};
            \draw node at (4, 3) [output] (out2) {$y_3$};

            \draw[->,thick] (in) -- (op);
            \draw[->,thick] (op) -- (out);
            \draw[->,thick] (op)  -- (op1);

            \draw[->,thick] (in1) -- (op1);
            \draw[->,thick] (op1) -- (out1);
            \draw[->,thick] (op1)  -- (op2);

            \draw[->,thick] (in2) -- (op2);
            \draw[->,thick] (op2) -- (out2);
        \end{tikzpicture}
        \caption{Unrolled recurrent neural network}
        \label{fig:rnn3}
    \end{center}
\end{figure}

Effectively, this graph represents an infinite family of acyclic graphs 
that unroll this loop a finite number of times.
For example, \cref{fig:rnn3} is an unrolling of length 3.
Notice that this is an acyclic graph that takes 3 inputs and produces 3 outputs.
The idea is that if you receive a sentence, say, with $n$ words,
you unroll the \rnn to length $n$ and apply it to the sentence.

Thinking of it through a programming lens,
given an input, we can easily statically determine---i.e., without executing the network---how many loop iterations it will require.
This is in contrast to, say, a program where the number of loop iterations is a complex function of its input,
and therefore we do not know how many loop iterations it will take until we actually run it.
That said, in what follows, we will formalize neural networks as acyclic graphs.

\section{Structure and Semantics of Neural Networks}\label{sec:dags}

We're done with looking at pretty graphs. Let's now look at pretty symbols.
We will now formally define neural networks as directed acyclic graphs and discuss some of their properties.

\subsection*{Neural Networks as DAGs}
A neural network is a directed acyclic graph $G = (\nodes, \edges)$,
where
\begin{itemize}
    \item $\nodes$ is a finite set of nodes,
    \item $\edges \subseteq \nodes \times \nodes$ is a set of edges,
    \item $\inodes \subset \nodes$ is a non-empty set of input nodes,
    \item $\onodes \subset \nodes$ is a non-empty set of output nodes, and
    \item each non-input node $\node$ is associated with a function $f_\node : \R^{n_\node} \to \R$, 
    where $n_\node$ is the number of edges whose target is node $\node$.  
    The vector of real values $\R^{n_\node}$ that $\node$ takes as input is all of the outputs of nodes $\node'$ such that $(\node',\node) \in \edges$.
    Notice that we assume, for simplicity but without loss of generality, that a node $\node$ only outputs a single real value.
  
\end{itemize}

To make sure that a graph $G$ does not have any dangling nodes and that semantics are clearly defined,
we will assume the following structural properties:
\begin{itemize} 
    \item All nodes are reachable, via directed edges, from some input node. 
    \item Every node can reach an output node.
    \item There is fixed total ordering on edges $\edges$ and another one on nodes $\nodes$.  

\end{itemize}

We will use $\vec{x} \in \R^n$ to denote an $n$-ary (row) vector,
which we represent as a tuple of scalars $(x_1,\ldots,x_n)$,
where $x_i$ is the $i$th element of $\vec{x}$.

\subsection*{Semantics of DAGs}
A neural network $G  = (\nodes, \edges)$ defines a function in $\R^{n} \to \R^{m}$
where $$n = |\inodes| ~\text{ and }~ m = |\onodes|$$
That is, $G$ maps the values of the input nodes to those of the output nodes.

Specifically, for every non-input node $\node \in \nodes$, we recursively define the value in $\R$ that it produces as follows.
Let $(\node_1,\node), \ldots, (\node_{n_\node},\node)$ be an ordered sequence of all edges whose target is node $\node$ (remember that we've assumed an order on edges). Then, we define the output of node $\node$ as
$$\outs(\node) = f_\node(x_1, \ldots, x_{n_\node})$$
where $x_i = \outs(\node_i)$, for $i \in \{1,\ldots,{n_\node}\}$.

The base case of the above definition (of $\outs$) is input nodes, since they have no edges incident on them.
Suppose that we're given an input vector $\vec{x} \in \R^n$.
Let $\node_1,\ldots,\node_n$ be an ordered sequence of all input nodes.
Then, $$\outs(\node_i) = x_i$$

\subsection*{A Simple Example}
Let's look at an example graph $G$:
    \begin{center}
        \begin{tikzpicture}

            \draw node at (0, 0) [input] (in) {$\node_1$};
            \draw node at (0, -1) [input] (in2) {$\node_2$};

            \draw node at (2, -0.5) [output] (op) {$\node_3$};

            \draw[->,thick] (in) -- (op);
            \draw[->,thick] (in2) -- (op);

        \end{tikzpicture}
    \end{center}
We have $\inodes = \{\node_1,\node_2\}$  and $\onodes = \{\node_3\}$.
Now assume that $$f_{\node_3}(x_1,x_2) = x_1 + x_2$$
and that we're given the input vector $(11,79)$ to the network,
where node $\node_1$ gets the value $11$ and $\node_2$ the value $79$.
Then, we have 
\begin{align*}
\outs(\node_1) &= 11\\
\outs(\node_2) &= 79\\
\outs(\node_3) &= f_{\node_3}(\outs(\node_1),\outs(\node_2)) = 11 + 79 = 90    
\end{align*}

\subsection*{Data Flow and Control Flow}
The graphs we have defined are known in the field of compilers and program analysis as \emph{data-flow} graphs;
this is in contrast to \emph{control-flow} graphs.\footnote{In deep learning frameworks like TensorFlow, they call data-flow graphs \emph{computation graphs}.}
Control-flow graphs dictate the \emph{order} in which operations need be performed---the flow of who has \emph{control} of the \cpu.
Data-flow graphs, on the other hand, only tell us what node needs what data to perform its computation,
but not how to order the computation.
This is best seen through a small example.

Consider the following graph
\begin{center}
    \begin{tikzpicture}

        \draw node at (0, 0) [input] (in) {$\node_1$};
        \draw node at (0, -1) [input] (in2) {$\node_2$};

        \draw node at (2, 0) [oper] (op) {$\node_3$};
        \draw node at (2, -1) [oper] (op2) {$\node_4$};

        \draw node at (4, -0.5) [output] (f) {$\node_5$};

        \draw[->,thick] (in) -- (op);
        \draw[->,thick] (in2) -- (op2);
        \draw[->,thick] (op) -- (f);
        \draw[->,thick] (op2) -- (f);
    \end{tikzpicture}
\end{center}
Viewing this graph as an imperative program, 
one way to represent it is as follows, where $\gets$ is the assignment symbol.
\begin{align*}
    \outs(\node_3) &\gets f_{\node_3}(\outs(\node_1))\\
    \outs({\color{RoyalBlue}{\node_4}}) &\gets f_{\node_4}(\outs(\node_2))\\
    \outs(\node_5) &\gets f_{\node_5}(\outs(\node_3), \outs(\node_4))
\end{align*}
This program dictates that the output value of node $\node_3$ is computed before that of node $\node_4$.
But this need not be, as the output of $\node_3$ does not depend on that of  $\node_4$.
Therefore, an equivalent implementation of the same graph can swap the first two operations:
\begin{align*}
    \outs({\color{RoyalBlue}{\node_4}}) &\gets f_{\node_4}(\outs(\node_2))\\
    \outs(\node_3) &\gets f_{\node_3}(\outs(\node_1))\\
    \outs(\node_5) &\gets f_{\node_5}(\outs(\node_3), \outs(\node_4))
\end{align*}

Formally, we can compute the values $\outs(\cdot)$ in any \emph{topological} ordering of graph nodes.
This ensures that all inputs of a node are computed before its own operation is performed. 

\subsection*{Properties of Functions}
So far, we have assumed that a node $\node$ can implement any function $f_\node$ it wants over real numbers.
In practice, to enable efficient training of  neural networks, these functions need be \emph{differentiable}
or differentiable \emph{almost everywhere}.
The sigmoid activation function, which we met earlier in \cref{fig:sigmoid},
is differentiable.
However, the ReLU activation function, \cref{fig:relu}, is differentiable almost everywhere, since at $x=0$,
there is a sharp turn in the function and the gradient is undefined.

Many of the functions we will be concerned with are \emph{linear} or \emph{piecewise linear}.
Formally, a function $f : \R^n \to \R$ is linear if it can be defined as follows:
$$f(\vec{x}) = \sum_{i=1}^n c_ix_i + b$$
where $c_i,b \in \R$.
A function is piecewise linear if it can be written in the form
\[ f(\vec{x}) = \begin{cases*}
    \sum_i c_i^1x_i + b^1, & $\vec{x} \in S_1$  \\
    \vdots &\\
    \sum_i c_i^mx_i + b^m, & $\vec{x} \in S_m$ 
 \end{cases*} \]
 where $S_i$ are mutually disjoint subsets of $\R^n$ and $\cup_i S_i = \R^n$.
ReLU, for instance, is a piecewise linear function, as it is of the form:
\[
\relu(x) = \begin{cases*}
    0, & $x < 0$  \\
    x, & $x \geq 0$ 
 \end{cases*}    
\]

Another important property that we will later exploit is \emph{monotonicity}.
A function $f: \R\to\R$ is monotonically increasing 
if for any $x \geq y$, we have $f(x) \geq f(y)$.
Both activation functions we saw earlier in the chapter, ReLUs and sigmoids, are monotonically increasing.
You can verify this in \cref{fig:relu,fig:sigmoid}: the functions never decrease with increasing values of $x$.













\section*{Looking Ahead}
Now that we have formally defined neural networks, we're ready to pose questions about their behavior.
In the next chapter, we will formally define a language for posing those questions.
Then, in the chapters that follow, we will look at algorithms for answering those questions.

Most discussions of neural networks in the literature 
use the language of linear algebra---see, for instance, the comprehensive book of \citet{goodfellow2016deep}.
Linear algebra is helpful because we can succinctly represent the operation
of many nodes in a single layer as a matrix $A$ that applies to the output of the previous layer.
Also, in practice, we use fast, parallel implementations of matrix multiplication to evaluate neural networks.
Here, we choose a lower-level presentation, where each node is a function in $\R^n \to \R$.
While this view is non-standard, it will help make our presentation of different verification techniques much cleaner, as we can decompose the problem 
into smaller ones that have to do with individual nodes.

The graphs of neural networks we presented are lower-level versions of
the computation graphs of deep-learning frameworks like Tensorflow~\citep{tf} and PyTorch~\citep{torch}

Neural networks are an instance of a general class of programs called \emph{differentiable programs}.  
As their name implies, differentiable programs are ones for which we can compute derivatives, a property that is needed for standard techniques 
for training neural networks.
Recently, there have been interesting studies of what it means for a program to be differentiable~\citep{abadi2019simple,sherman2020lambda_s}.
In the near future, it is likely that people will start using arbitrary differentiable programs to define and train neural networks.
Today, this is not the case, most neural networks have one of a few prevalent architectures and operations.

\chapter{Correctness Properties}\label{ch:correctness}

In this chapter, we will come up with a \emph{language} for specifying properties of neural networks. 
The specification language is a formulaic way of making statements about the behavior of a neural network (or sometimes multiple neural networks).
Our concerns in this chapter are solely about specifying properties, not about automatically verifying them. 
So we will take liberty in specifying complex properties, ridiculous ones, and useless ones. 
In later parts of the book, we will constrain the properties of interest to fit certain verification algorithms---%
for now, we have fun.

\section{Properties, Informally}

Remember that a neural network defines a function $f : \R^n \to \R^m$.
The properties we will consider here are of the form: 
\begin{displayquote}
    \textbf{for any input} $x$, the neural network \textbf{produces an output} that ...
\end{displayquote}
In other words, properties dictate the input--output behavior of the network, but not the internals of the network---how it comes up with the answer.

Sometimes, our properties will be more involved, talking about multiple inputs, and perhaps multiple networks:
\begin{displayquote}
    \textbf{for any inputs $x,y,$} ... \textbf{that} ... the neural networks \textbf{produce outputs} that ...
\end{displayquote}

The first part of these properties, the one talking about inputs, is called the \emph{precondition};
the second part, talking about outputs, is called the \emph{postcondition}.
In what follows, we will continue our informal introduction to properties using examples.

\begin{figure}[!tbp]
    \centering
    \begin{minipage}[b]{0.3\textwidth}
      \includegraphics[width=\textwidth]{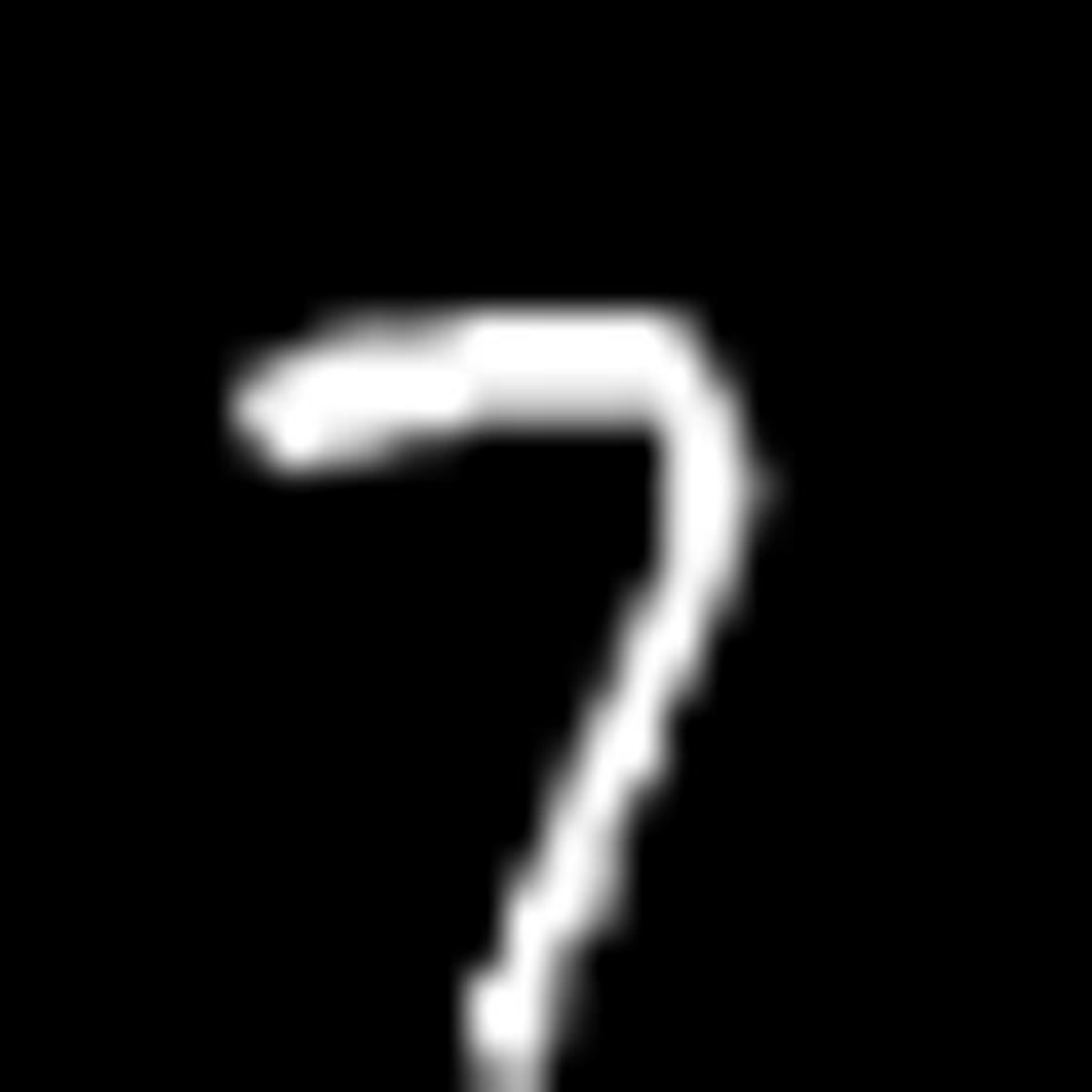}
    \end{minipage}
    \hfill
    \begin{minipage}[b]{0.3\textwidth}
      \includegraphics[width=\textwidth]{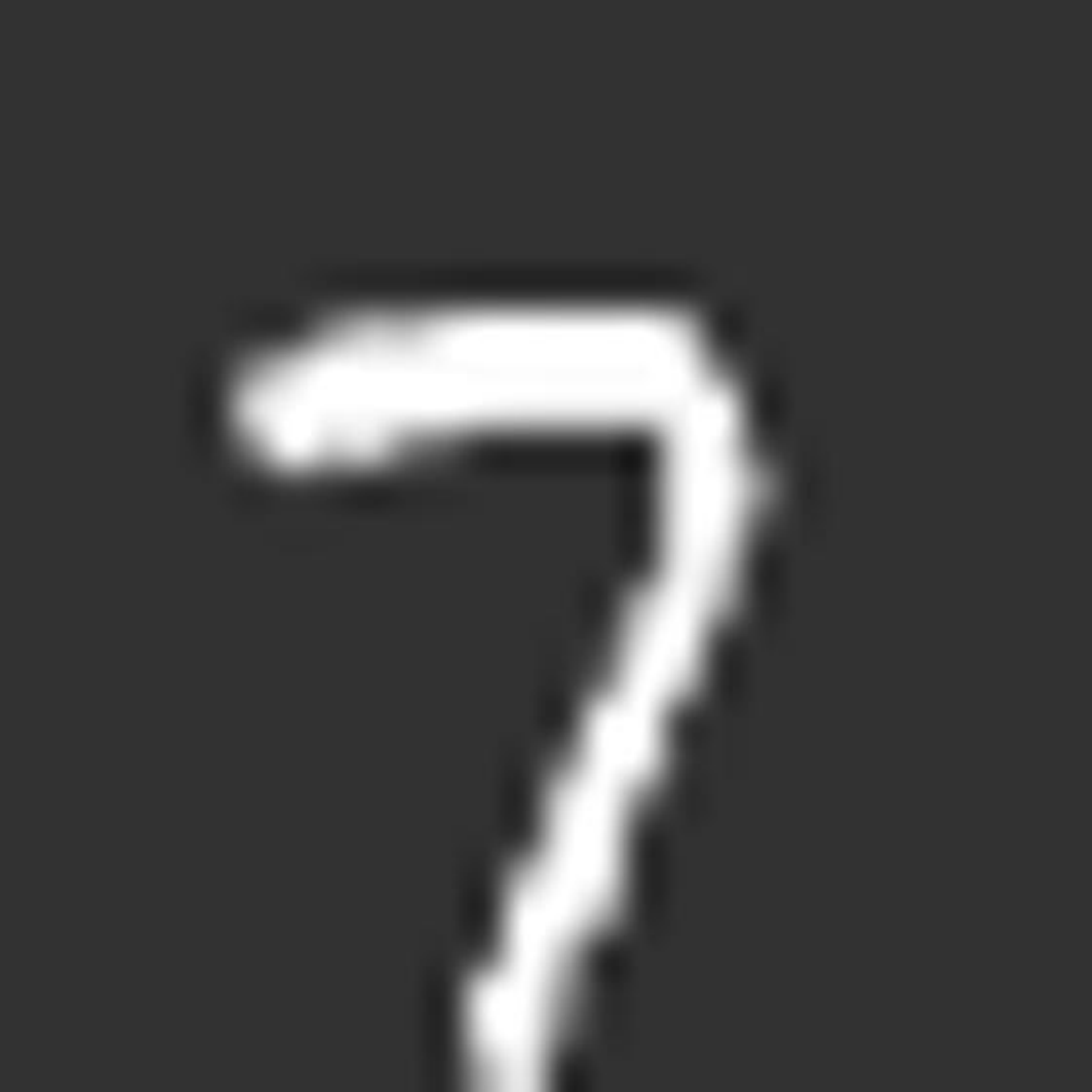}
    \end{minipage}
    \hfill
    \begin{minipage}[b]{0.3\textwidth}
      \includegraphics[width=\textwidth]{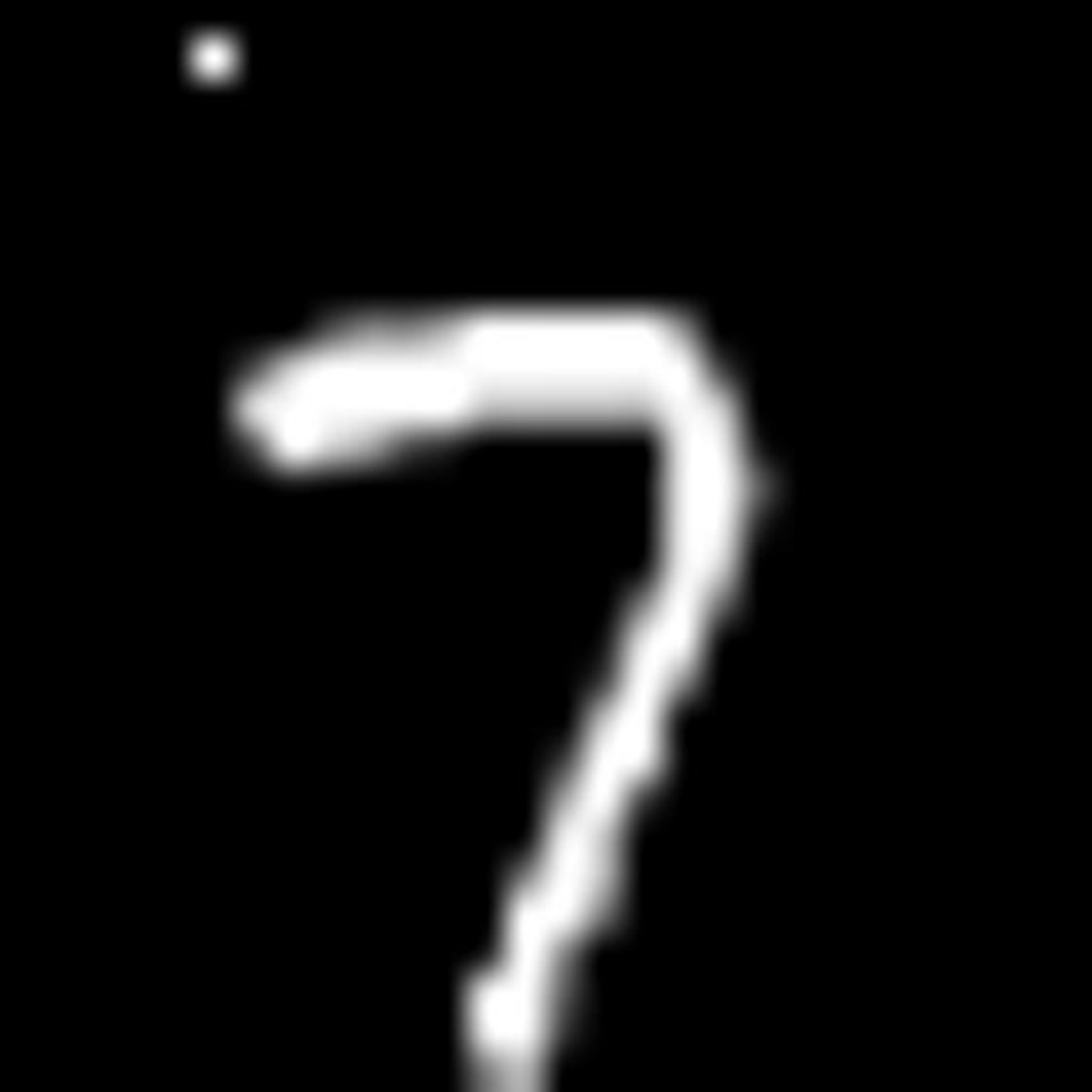}
    \end{minipage}
    \caption{Left: Handwritten 7 from \textsc{mnist} dataset.
    Middle: Same digit with increased brightness. 
    Right: Same digit but with a dot added in the top left.}
    \label{fig:rob7}
\end{figure}

\subsection*{Image Recognition}

Let's say we have a neural network $f$ that takes in an image and predicts a label from \emph{dog}, \emph{zebra}, etc.
An important property that we may be interested in ensuring is \emph{robustness} of such classifier. 
A classifier is robust if its prediction does not change with small variations (or perturbations) of the input. For example, changing the brightness slightly or damaging a few pixels should not change classification.

Let's fix some image $\vec{c}$ that is classified as \emph{dog} by $f$.
To make sure that $\vec{c}$ is not an \emph{adversarial image} of a dog that is designed to fool the neural network, we will check---\emph{prove} or \emph{verify}---the following property:
\begin{displayquote}
for any image $\vec{x}$ that is slightly brighter or darker than $\vec{c}$, $f(\vec{x})$ predicts \emph{dog}
\end{displayquote}
Notice here that the precondition specifies a set of images $\vec{x}$ that are brighter or darker than $\vec{c}$, and the postcondition specifies that the classification by $f$ remains unchanged.

Robustness is a desirable property: you don't want classification to change with a small movement in the brightness slider.
But there are many other  properties you desire---robustness to changes in contrast, rotations, Instagram filters, white balance, and the list goes on. 
This hits at the crux of the specification problem: we often cannot specify every possible thing that we desire, so we have to choose some.
(More on this later.)

For a concrete example, see \cref{fig:rob7}.
The \textsc{mnist} dataset~\citep{mnist} is a standard dataset for recognizing handwritten digits.
The figure shows a handwritten 7 along with two modified versions,
one where brightness is increased and one where a spurious dot is added---perhaps a drip of ink.
We would like our neural network to classify all three images as 7.

\subsection*{Natural-Language Processing}
Suppose now that $f$ takes an English sentence and decides whether it represents a  positive or negative sentiment.
This problem arises, for example, in automatically analyzing online reviews or tweets.
We're also interested in robustness in this setting.
For example, say we have fixed a sentence $\vec{c}$ with positive sentiment, then we might specify the following property:
\begin{displayquote}
    for any sentence $\vec{x}$ that is $\vec{c}$ with a few spelling mistakes added, $f(\vec{x})$ should predict positive sentiment
\end{displayquote}

For another example, instead of spelling mistakes, imagine replacing words with synonyms:
\begin{displayquote}
    for any sentence $\vec{x}$ that is $\vec{c}$ with some words replaced by synonyms, then $f(\vec{x})$ should predict positive sentiment
\end{displayquote}
For instance, a neural network should classify both of these movie
reviews as positive reviews:
\begin{displayquote}
    \emph{This movie is delightful}\\
    \emph{This movie is enjoyable}
\end{displayquote}

We could also combine the two properties above to get a stronger property specifying that prediction should not change in the presence of synonyms or spelling mistakes. 

\subsection*{Source Code}
Say that our neural network $f$ is a malware classifier, taking a piece of code and deciding whether it is malware or not.
A malicious entity may try to modify a malware to sneak it past the neural network by fooling it into thinking that it's a benign program.
One trick the attacker may use is adding a piece of code that does not change the malware's operation but that fools the neural network.
We can state this property as follows: 
Say we have piece of malware $\vec{c}$, then we can state the following property:
\begin{displayquote}
    for any program $\vec{x}$ that is equivalent to $\vec{c}$ and syntactically similar, then $f(\vec{x})$ predicts malware
\end{displayquote}

\subsection*{Controllers}
All of our examples so far have been robustness problems.
Let's now look at a slightly different property.
Say you have a controller deciding on the actions of a robot.
The controller looks at the state of the world and decides whether to move left, right, forward, or backward.
We, of course, do not want the robot to move into an obstacle, whether it is a wall, a human, or another robot.
As such, we might specify the following property:
\begin{displayquote}
    for any state $\vec{x}$, if there is an obstacle to the right of the robot, then $f(x)$ should \emph{not} predict right
\end{displayquote}
We can state one such property per direction.

\section{A Specification Language}
Our specifications are going to look like this:
\[
\begin{array}{c}
    \pre{\emph{precondition}}\\
    \vec{r} \gets f(\vec{x})\\
    \post{\emph{postcondition}}
\end{array}
\]
The \emph{precondition} is a Boolean function (\emph{predicate}) that evaluates to true or false. The precondition is defined over a set of variables which will be used as inputs to the neural networks we're reasoning about. We will use $\vec{x}_i$ to denote those variables.
The middle portion of the specification is a number of calls to functions defined by neural networks; in this example, we only see one call to $f$, and the return value is stored in a variable $r$. Generally, our specification language allows a sequence of such assignments, e.g.:
\[
\begin{array}{c}
    \pre{\emph{precondition}}\\
    \vec{r_1} \gets f(\vec{x_1})\\
    \vec{r_2} \gets g(\vec{x_2})\\
    \vdots\\
    \post{\emph{postcondition}}
\end{array}
\]
Finally, the postcondition is a Boolean predicate over the variables appearing in the precondition $\vec{x}_i$ and the assigned variables $\vec{r}_j$.

The way to read a specification, informally, is as follows:
\begin{displayquote}
    for any values of $\vec{x}_1,\ldots,\vec{x}_n$ that make the  precondition true,
    let $\vec{r}_1 = f(\vec{x_1}), \vec{r}_2 = g(\vec{x}_2),\ldots$. Then the postcondition is true.
\end{displayquote}
If a correctness property is not true, i.e., the postcondition yields false, we will also say that the property \emph{does not hold}.

\begin{example}\label{ex:robustness}
    Recall our image brightness example from the previous section, and say $\vec{c}$ is an actual grayscale image, where each element of $\vec{c}$ is the intensity of a pixel, from 0 to 1 (black to white). 
    For example, in our \textsc{mnist} example in \cref{fig:rob7},
    each digit is represented by $784$ pixels ($28\times28$),
    where each pixel is a number between 0 and 1.
Then, we can state the following specification, which informally says that  changing the brightness of $\vec{c}$ should not change the classification
(recall the definition of $\class$ from \cref{ssec:layers}):
\[
\begin{array}{c}
    \pre{|\vec{x} - \vec{c}| \leq \vec{0.1}}\\
    \vec{r}_1 \gets f(\vec{x})\\
    \vec{r}_2 \gets f(\vec{c})\\
    \post{\class(\vec{r}_1) = \class(\vec{r}_2)}
\end{array}
\]
Let's walk through this specification:
\begin{description}
\item[Precondition] Take any image $\vec{x}$ where each pixel is at most 0.1 away from its counterpart in $\vec{c}$. Here, both $\vec{x}$ and $\vec{c}$ are assumed to be the same size, and the $\leq$ is defined pointwise.\footnote{The pointwise operation $|\cdot|$ is known as the $\ell_\infty$ norm, which we formally discuss in \cref{ch:absver} and compare it to other norms.}
\item[Assignments] Let $\vec{r}_1$ be the result of computing $f(\vec{x})$ and $\vec{r}_2$ be the result of computing $f(\vec{c})$. 
\item[Postcondition] Then, the predicted labels in vectors $\vec{r}_1$ and $\vec{r}_2$ are the same. Recall that in a classification setting, each element of vector $\vec{r}_i$ refers to the probability of a specific label.
We use $\class$ as a shorthand to extract the index of the largest element of the vector.  
\end{description}
\end{example}

\subsection*{Counterexamples}
A \emph{counterexample} to a property
is a  valuation of the 
variables in the precondition (the $\vec{x}_i$s)
that falsifies the postcondition.
In \cref{ex:robustness},
a counterexample would be an image $\vec{x}$
whose classification by $f$ is different than 
that of image $\vec{c}$ and whose distance from $\vec{c}$, i.e., $|\vec{x} - \vec{c}|$, is less than $0.1$.

\begin{example}
    Here's a concrete example (not about image recognition, just a simple function that adds 1 to the input):
    \[
        \begin{array}{c}
            \pre{x \leq 0.1}\\
            r \gets x + 1\\
            \post{r \leq 1}
        \end{array}
    \]
This property does not hold.
Consider replacing $x$ with the value $0.1$.
Then, $r \gets 1 + 0.1 = 1.1$.
Therefore, the postcondition is falsified.
So, setting $x$ to $0.1$ is a counterexample.
\end{example}

\subsection*{A Note on Hoare Logic}
Our specification language looks like specifications written in \emph{Hoare logic}~\citep{hoare}.
Specifications in Hoare logic are called \emph{Hoare triples}, as they are composed of three parts, just like our specifications.
Hoare logic comes equipped with deduction rules that allows one to prove the validity of such specifications.
For our purposes in this book, we will not define the rules of Hoare logic, but many of them will crop up implicitly throughout the book.

\section{More Examples of Properties}

We will now go through a bunch of example properties and write them in our specification language.

\subsection*{Equivalence of Neural Networks}
Say you have a neural network $f$ for image recognition and you want to replace it with a new neural network $g$. Perhaps $g$ is smaller and faster, and since you're interested in running the network on a stream of incoming images, efficiency is very important.
One thing you might want to prove is that $f$ and $g$ are equivalent; here's how to write this property:
\[
\begin{array}{c}
    \pre{\true}\\
    \vec{r_1} \gets f(\vec{x})\\
    \vec{r_2} \gets g(\vec{x})\\
    \post{\class(\vec{r}_1) = \class(\vec{r}_2)}
\end{array}
\]
Notice that the precondition is $\true$, meaning that for any image $\vec{x}$, we want the predicted labels of $f$ and $g$ to be the same.
The $\true$ precondition indicates that the inputs to the neural networks ($\vec{x}$ in this case) are unconstrained.
This specification is very strong: the only way it can be true is if $f$ and $g$ agree on the classification on every possible input, which is highly unlikely in practice.

One possible alternative is to state that $f$ and $g$ return the same prediction on some subset of images, plus or minus some brightness, as in our above example.
Say $S$ is a finite set of images, then:
\[
\begin{array}{c}
    \pre{\vec{x}_1 \in S , \  |\vec{x}_1 - \vec{x}_3| \leq \vec{0.1} , \  |\vec{x}_1 - \vec{x}_2| \leq \vec{0.1} }\\
    \vec{r}_1 \gets f(\vec{x}_2)\\
    \vec{r}_2 \gets g(\vec{x}_3)\\
    \post{\class(\vec{r}_1) = \class(\vec{r}_2)}
\end{array}
\]
This says the following: Pick an image $\vec{x}_1$ and generate two variants, $\vec{x}_2$ and $\vec{x}_3$, whose brightness differs a little bit from $\vec{x}_1$. Then, $f$ and $g$ should agree on the classification of the two images.

This is a more practical notion of equivalence than our first attempt.
Our first attempt forced $f$ and $g$ to agree on all possible inputs, but keep in mind that most images (combinations of pixels) are meaningless noise, and therefore we don't care about their classification.
This specification, instead, constrains equivalence to an infinite set of images that look like those in the set $S$.

\subsection*{Collision Avoidance}
Our next example is one that has been a subject of study in the verification literature, beginning with the pioneering work of \citet{katz2017reluplex}.
Here we have a collision avoidance system that runs on an autonomous aircraft.
The system detects intruder aircrafts and decides what to do.
The reason the system is run on a neural network is due to its complexity: The trained neural network is much smaller than a very large table of rules.
In a sense, the neural network \emph{compresses} the rules into an efficiently executable program.

The inputs to the neural network are the following:
\begin{itemize}
\item $v_\emph{own}$: the aircraft's velocity
\item $v_\emph{int}$: the intruder aircraft's velocity
\item $a_\emph{int}$: the angle of the intruder with respect to the current flying direction
\item $a_\emph{own}$: the angle of the aircraft with respect to the intruder.
\item $d$: the distance between the two aircrafts
\item $\emph{prev}$: the previous action taken.
\end{itemize}
Given the above values, the neural network decides how to steer:
left/right, strong left/right, or nothing.
Specifically, the neural network assigns a score to every possible action, and the action with the lowest score is taken.

As you can imagine, many things can go wrong here, and if they do---disaster!
\citet{katz2017reluplex} identify a number of properties that they verify.
These properties do not account for all possible scenarios, but they are important to check.
Let's take a look at one that says if the intruder aircraft is far away, then the score for doing \emph{nothing} should be below some threshold.
\[
\begin{array}{c}
    \pre{d \geq 55947, \ v_\emph{own} \geq 1145 , \ v_\emph{int} \leq 60}\\
    \vec{r} \gets f(d,v_\emph{own},v_\emph{int}, \ldots )\\
    \post{\text{score of nothing in } \vec{r} \text{ is below 1500}}
\end{array}
\]
Notice that the precondition specifies that the distance between the two aircrafts is more than 55947 feet,
that the aircraft's velocity is high, and the intruder's velocity is low.
The postcondition specifies that doing nothing should have a low score, below some threshold.
Intuitively, we should not panic if the two aircrafts are quite far apart and have moving at very different velocities.

\citet{katz2017reluplex} explore a number of such properties, and also consider robustness properties in the collision-avoidance setting.
But how do we come up with such specific properties?
It's not straightforward. 
In this case, we really need a domain expert who knows about collision-avoidance systems, and even then, we might not cover all corner cases.
A number of people in the verification community, the author included, argue that specification is harder than verification---that is, the hard part is asking the right questions!

\subsection*{Physics Modeling}
Here is another example due to \citet{QinDOBSGUSK19}.
We want the neural network to internalize some physical laws, such as the movement of a pendulum.
At any point in time, the state of the pendulum is a triple $(v,h,w)$, its vertical position $v$, its horizontal position $h$, and its angular velocity $w$.
Given the state of the pendulum, the neural network is to predict the state in the next time instance, assuming that time is divided into discrete steps.

A natural property we may want to check is that the neural network's understanding of how the pendulum moves adheres to the law of conservation of energy. At any point in time, the energy of the pendulum is 
the sum of its potential energy and its kinetic energy. (Were you paying attention in high school physics?) As the pendulum goes up, its potential energy increases and kinetic energy decreases; as it goes down, the opposite happens. The sum of the kinetic and potential energies should only decrease over time.
We can state this property as follows:
\[
\begin{array}{c}
    \pre{\true}\\
    v',h',w'  \gets f(v,h,w)\\
    \post{E(h',w') \leq E(h,w)}
\end{array}
\]

The expression $E(h,w)$ is the energy of the pendulum, which is its potential energy $mgh$, where $m$ is the mass of the pendulum and $g$ is the gravitational constant, plus its kinetic energy  $0.5ml^2w^2$, where $l$ is the length of the pendulum.

\subsection*{Natural-Language Processing}
Let's recall the natural language example from earlier in the chapter, where we wanted to classify a sentence into whether it expresses a positive or negative sentiment. We decided that we want the classification not to change if we replaced a word by a synonym.
We can express this property in our language:
Let $\vec{c}$ be a fixed sentence of length $n$.
We assume that each element of vector $\vec{c}$ is a real number representing a word---called an \emph{embedding} of the word.
We also assume that we have a thesaurus $T$, which given a word gives us a set of equivalent words.
\[
\begin{array}{c}
    \pre{1\leq i \leq n , \  w \in T(c_i) , \  \vec{x} = \vec{c}[i \mapsto w] }\\
    \vec{r}_1  \gets f(\vec{x})\\
    \vec{r}_2  \gets f(\vec{c})\\
    \post{\class(\vec{r}_1) = \class(\vec{r}_2)}
\end{array}
\]
The precondition specifies that variable $\vec{x}$ is just like the sentence $\vec{c}$, except that some element $i$ is replaced by a word $w$ from the thesaurus. 
We use the notation $ \vec{c}[i \mapsto w] $ to denote $\vec{c}$ with the $i$th element replaced with $w$ and $c_i$ to denote the $i$th element of $\vec{c}$.

The above property allows a single word to be replaced by a synonym.
We can extend it to two words as follows (I know, it's very ugly, but it works):
\[
\begin{array}{c}
    \pre{1\leq i,j \leq n \; , \; i\neq j  \; , \; w_i \in T(c_i)  \; , \; w_j \in T(c_j) \; , \; \vec{x} = \vec{c}[i \mapsto w_i, j \mapsto w_j] }\\
    \vec{r}_1  \gets f(\vec{x})\\
    \vec{r}_2  \gets f(\vec{c})\\
    \post{\class(\vec{r}_1) = \class(\vec{r}_2)}
\end{array}
\]

\subsection*{Monotonicity}
A standard mathematical property that we may desire 
of neural networks is monotonicity~\citep{DBLP:conf/nips/SivaramanFMB20}, meaning that larger inputs should 
lead to larger outputs.
For example, imagine you're one of those websites that predict house prices
using machine learning. You'd expect the machine-learning model used 
is monotonic with respect to square footage---if you increase the square footage of a house, its price should not decrease, or perhaps increase.
Or imagine a model that estimates the risk of complications during surgery.
You'd expect that increasing the age of the patient should not decrease the risk. (I'm not a physician, but I like this example.)
Here's how you could encode monotonicity in our language:
\[
\begin{array}{c}
    \pre{\vec{x} > \vec{x}'}\\
    \vec{r}  \gets f(\vec{x})\\
    \vec{r}'  \gets f(\vec{x}')\\
    \post{\vec{r}' \geq \vec{r}'}
\end{array}
\]
In other words, pick any pair of inputs such that $\vec{x} > \vec{x}'$,
we want $f(\vec{x}) \geq f(\vec{x}')$.
Of course, we can \emph{strengthen} the property by making the postcondition
a strict inequality---that completely depends on the problem domain we're working with.

\section*{Looking Ahead}
We're done with the first part of the book.
We have defined neural networks and how to specify their properties.
In what follows, we will discuss different ways of verifying properties
automatically.

There has been an insane amount of work on robustness problems,
particularly for image recognition.
Lack of robustness was first observed by \citet{SzegedyZSBEGF13},
and since then many approaches to discover and defend against robustness violations (known as adversarial examples) have been proposed.
We will survey those later.
The robustness properties for natural-language processing we have defined follow those of~\citet{ebrahimi2017hotflip} and~\citet{huang2019achieving}.

\part{Constraint-Based Verification}

     \chapter{Logics and Satisfiability}\label{ch:fol}

In this part of the book, we will look into constraint-based techniques for verification. The idea is to take a correctness property and encode it as a set of constraints. By solving the constraints, we can decide whether the correctness property holds or not.

The constraints we will use are formulas in \emph{first-order logic} (\fol).
\fol is a very big and beautiful place, but neural networks only live in a small and cozy corner of it---the corner that we will explore in this chapter.

\section{Propositional Logic}
We begin with the purest of all, \emph{propositional logic}.
A formula $\varphi$ in propositional logic is over Boolean variables (traditionally given the names $p,q,r,\ldots$) and defined using the following grammar:
\[
\begin{array}{llr}
\varphi \coloneqq
     & \true & \\
     & \false & \\
     & \emph{var} & \text{Variable}\\
     & \mid \varphi \land \varphi  & \text{Conjunction (and)}\\
     & \mid \varphi \lor \varphi & \text{Disjunction (or)} \\
     & \mid \neg \varphi & \text{Negation (not)}
\end{array}
\]

Essentially, a formula in propositional logic defines a circuit with Boolean variables, \textsc{and} gates ($\land$), \textsc{or} gates ($\lor$), and \text{not} gates ($\neg$).
Negation has the highest operator precedence, followed by conjunction and then disjunction.
At the end of the day, all programs can be defined as circuits, because everything is a bit on a computer and there is a finite amount of memory, and therefore a finite number of variables.

We will use $\fv(\varphi)$ to denote the set of \emph{free} variables appearing in the formula. For our purposes, this is the set of all variables that are syntactically present in the formula; 

\begin{example}
As an example, here is a formula $$\varphi \triangleq (p \land q) \lor \neg r $$
Observe the use of $\triangleq$; this is to denote that we're syntactically defining $\varphi$ to be the formula on the right of $\triangleq$, as opposed to saying that the two formulas are semantically equivalent (more on this in a bit).
The set of free variables in $\varphi$ is $\fv(\varphi) = \{p,q,r\}$.
\end{example}

\subsection*{Interpretations}
Let $\varphi$ be a formula over a set of variables $\fv(\varphi)$.
An interpretation $I$ of $\varphi$ is a map from variables $\fv(\varphi)$ to $\true$ or $\false$.
Given an interpretation $I$ of a formula $\varphi$, we will use $I(\varphi)$
to denote the formula where we have replaced each variable in $\fv(\varphi)$ with its interpretation in $I$.

\begin{example}
Say we have the formula $$\varphi \triangleq (p \land q) \lor \neg r$$
and the interpretation $$I = \{p \mapsto \true,\ q \mapsto \true,\ r \mapsto \false\}$$
Note that  we represent $I$ as a set of pairs of variables and their interpretations.
Applying $I$ to $\varphi$,
we get $$I(\varphi) \triangleq (\true \land \true) \lor \neg \false$$
\end{example}

\subsection*{Evaluation Rules}
We will define the following evaluation, or simplification, rules for a formula.
The formula on the right of $\equiv$ is an equivalent, but syntactically simpler, variant of the one on the left: 
\[
\begin{array}{lclrr}
\true \land \varphi & \equiv & \varphi  && \text{Conjunction}\\
\varphi \land \true & \equiv &  \varphi &&\\
\false \land \varphi & \equiv & \false && \\
\varphi \land \false & \equiv &  \false && 
\end{array}
\]
\[
\begin{array}{lclrr}
     \false \lor \varphi & \equiv & \varphi && \text{Disjunction}\\
\varphi \lor \false & \equiv &  \varphi &&\\
\true \lor \varphi & \equiv & \true &&\\
\varphi \lor \true & \equiv &  \true &&
\end{array}
\]
\[
\begin{array}{lclrr}
\neg \true & \equiv & \false && \text{Negation}\\
\neg \false & \equiv &  \true &&
\end{array}
\]

If a given formula has no free variables, then by applying these rules repeatedly, you will get $\true$ or $\false$.
We will use $\eval(\varphi)$ to denote the simplest form of $\varphi$ we can get by repeatedly applying the above rules.

\subsection*{Satisfiability}
A formula $\varphi$ is \emph{satisfiable} (\sat) if  there exists an interpretation $I$ such that $$\eval(I(\varphi)) = \true$$
in which case we will say that $I$ is a \emph{model} of $\varphi$
and denote it $$I \models \varphi$$ 
We will also use $I \not\models \varphi$ to denote that $I$ is not a model of $\varphi$. 
It follows from our definitions that $I \not\models \varphi$ iff $I \models \neg \varphi$.

Equivalently, a formula $\varphi$ is \emph{unsatisfiable} (\unsat) if  for every interpretation $I$ we have $\eval(I(\varphi))
 = \false$.

\begin{example}
Consider the formula $\varphi \triangleq (p \lor q) \land (\neg p \lor r)$.
This formula is satisfiable; here is a model $I = \{p \mapsto \true, q \mapsto \false, r \mapsto \true\}$.
\end{example}

\begin{example}
Consider the formula $\varphi \triangleq (p \lor q) \land \neg p \land \neg q$.
This formula is unsatisfiable.
\end{example}

\subsection*{Validity and Equivalence}
To prove properties of neural networks, we will be asking \emph{validity} questions.
A formula $\varphi$ is valid if every possible interpretation $I$ is a model of $\varphi$.
It follows that a formula $\varphi$ is valid if and only if $\neg\varphi$ is unsatisfiable.

\begin{example}
Here is a valid formula $\varphi \triangleq (\neg p \lor q) \lor p$.
Pick any interpretation $I$ that you like; you will find that $I\models \varphi$. 
\end{example}

We will say that two formulas, $A$ and $B$,
are \emph{equivalent} if and only if every model $I$ of $A$ is a model of $B$, and vice versa. 
We will denote equivalence as $A \equiv B$.
There are many equivalences that are helpful when working with formulas.
For any formulas $A$, $B$, and $C$, we have commutativity of conjunction and disjunction,
\[
\begin{array}{lcl}
A \land B & \equiv & B \land A \\
A \lor B & \equiv & B \lor A   
\end{array}
\]
We can push negation inwards:
\[
\begin{array}{lcl}
\neg (A \land B) & \equiv & \neg A \lor \neg B \\
\neg (A \lor B) & \equiv & \neg A \land \neg B   
\end{array}
\]
Moreover, we have distributivity of conjunction over disjunction (\emph{DeMorgan's laws}), and vice versa:
\[
\begin{array}{lcl}
A \lor (B \land C) & \equiv & (A \lor B) \land (A \lor C) \\
A \land (B \lor C) & \equiv & (A \land B) \lor (A \land C) 
\end{array}
\]

\subsection*{Implication and Bi-implication}
We will often use an \emph{implication} $A \Rightarrow B$ to denote the formula $$\neg A \lor B$$
Similarly, we will use a \emph{bi-implication} $A \iff B$ to denote the formula $$(A \Rightarrow B) \land (B \Rightarrow A)$$

\section{Arithmetic Theories}
We can now extend propositional logic using \emph{theories}.
Each Boolean variable now becomes a more complex Boolean expression over variables of different types.
For example, we can use the theory of \emph{linear real arithmetic} (\lra), where a Boolean expression is, for instance,
$$ x + 3y + z \leq 10$$
Alternatively, we can use the theory of \emph{arrays}, and so an expression may look like:
$$a[10] = x$$
where $a$ is an array indexed by integers.
There are many other theories that people have studied, including \emph{bitvectors} (to model machine arithmetic) and \emph{strings} (to model string manipulation).
The satisfiabilty problem is now called \emph{satisfiability modulo theories} (\smt), as we check satisfiability with respect to  interpretations of the theory.

In this section, we will focus on the theory of linear real arithmetic (\lra), as it is (1) decidable and (2) can represent a large class of neural-network operations, as we will see in the next chapter. 

\subsection*{Linear Real Arithmetic}
In \lra, each propositional variable is replaced by a linear inequality of the form:
\[
\sum_{i=1}^n c_ix_i + b \leq 0 
\]
or 
\[
\sum_{i=1}^n c_ix_i + b < 0 
\]
where $c_i,b \in \R$ and $\{x_i\}_i$ is a fixed set of variables.  
For example, we can have a formula of the form:
\[
(x + y \leq 0  \ \land \  x - 2y < 10) \lor x > 100      
\]

Note that $>$ and $\geq$ can be rewritten into $<$ and $\leq$.
Also note that when a coefficient $c_i$ is 0, we simply drop the term $c_ix_i$, as in the inequality $x > 100$ above, which does not include the variable $y$.
An equality $x = 0$ can be written as the conjunction $x \geq 0 \land x \leq 0$.
Similarly, a disequality $x \neq 0$ can be written as $x < 0 \lor x > 0$.

\subsection*{Models in LRA}
As with propositional logic, the free variables $\fv(\varphi)$ of a formula $\varphi$ in \lra is the set of variables appearing in the formula.

An interpretation $I$ of a formula $\varphi$ is an assignment of every free variable to a real number.
An interpretation $I$ is a model of $\varphi$, i.e., $I \models \varphi$, iff $\eval(I(\varphi)) = \true$.
Here, the extension of the simplification rules to \lra formulas is straightforward:
all we need is to add standard rules for evaluating arithmetic inequalities, e.g.,
$2 \leq 0 \equiv \false$.

\begin{example}
As an example, consider the following formula:
\[
\varphi \triangleq  x - y > 0 \; \land \; x \geq 0     
\]
A model $I$ for $\varphi$ is \[\{x \mapsto 1, y \mapsto 0\}\]
Applying $I$ to $\varphi$, i.e., $I(\varphi)$, results in 
$$1-0 > 0 \; \land \; 1 \geq 0$$
Applying the evaluation  rules, we get $\true$.
\end{example}

\subsection*{Real vs. Rational}
In the literature, you might find \lra being referred to as \emph{linear rational arithmetic}.
There are two interrelated reasons for that:
First, whenever we write formulas in practice, the constants in those formulas are rational values---we can't really represent $\pi$, for instance, in computer memory.
Second, let's say that $\varphi$ contains only rational coefficients. 
Then, it follows that, if $\varphi$ is satisfiable, there is a model of $\varphi$ that assigns all free variables to rational values.

\begin{example}
Let's consider a simple formula like $x < 10$.
While $\{x \mapsto \pi\}$ is a model of $x < 10$,
it also has satisfying assignments that assign $x$ to a rational constant, like $\{x \mapsto 1/2\}$.
This will always be the case: we cannot construct formulas that only have irrational models, unless the formulas themselves contain irrational constants, e.g., $x = \pi$.
\end{example}

\subsection*{Non-Linear Arithmetic}
Deciding satisfiability of formulas in \lra is an \textsc{np}-complete problem.
If we extend our theory to allow for polynomial inequalities, then the best known algorithms are doubly exponential in the size of the formula in the worst case~\citep{caviness2012quantifier}.
If we allow for transcedental functions---like $\exp$, $\cos$, $\log$, etc.---then satisfiability becomes undecidable~\citep{tarski1998decision}.
Thus, for all practical purposes, we stick to \lra.
Even though it is \textsc{np}-complete (a term that sends shivers down the spines of theoreticians), we have very efficient algorithms that can scale to large formulas. 




\subsection*{Connections to MILP}
Formulas in \lra, and the \smt problem for \lra, is equivalent to the \emph{mixed integer linear programming} (\milp) problem.
Just as there are many \smt solvers, there are many \milp solvers out there, too. So the natural question to ask is why don't we use \milp solvers?
In short, we can, and maybe sometimes they will actually be faster than \smt solvers.
However, the \smt framework is quite general and flexible. So not only can we write formulas in \lra, but we can (1) write formulas in different theories, as well as (2)
formulas \emph{combining} theories.

First, in practice, neural networks do not operate over real or rational arithmetic. They run using floating point, fixed point, or machine-integer arithmetic. If we wish to be as precise as possible at analyzing neural networks, we can opt for a bit-level encoding of its operations and use bitvector theories employed by \smt solvers. (Machine arithmetic, surprisingly, is practically more expensive to solve than linear real arithmetic, so most of the time we opt for a real-arithmetic encoding of neural networks.)

Second, as we move forward and neural networks start showing up everywhere, we do not want to verify them in isolation, but in conjunction with other pieces of code that the neural network interacts with. For example, think of a piece of code that parses text and puts it in a form ready for the neural network to consume. Analyzing such piece of code might require using \emph{string} theories, which allow us to use string concatenation and other string operations in formulas.
\smt solvers employ theorem-proving techniques for \emph{combining} theories, and so we can write formulas, for example, over strings and linear arithmetic.

These are the reasons why in this book we use \smt solvers as the target of our constraint-based verification: they give us many first-order theories and allow us to combine them.
However, it is important to note that, at the time of writing this, most research on constraint-based verification focuses on linear real arithmetic encodings.

\section*{Looking Ahead}
In the next chapter, we will look at how to encode neural-network semantics, and correctness properties, as formulas in \lra, thus enabling automated verification using \smt solvers.
After that, we will spend some time studying the algorithms underlying \smt solvers.

In verification, we typically use fragments of first-order logic to encode programs. \fol has a long and storied history. \fol is a very general logic, and its satisfiability is undecidable, thanks to a proof by \citet{DBLP:journals/jsyml/Church36}.
\smt solvers, which have been heavily studied over the past twenty years or so 
aim at solving fragments of \fol, like \lra and other theories.
I encourage the interested reader to consult the \emph{Handbook of Satisfiability}  
for an in-depth exposition~\citep{biere2009handbook}.

     \chapter{Encodings of Neural Networks}\label{ch:encodings}

Our goal in this chapter is to translate a neural network into a formula in linear real arithmetic (\lra). 
The idea is to have the formula precisely (or soundly) capture the input--output relation of the neural network.
Once we have such a formula, we can use it to verify correctness properties using \smt solvers.

\section{Encoding Nodes}
 We begin by characterizing a relational view of a neural network.
 This will help us establish the correctness of our encoding.

\subsection*{Input-output Relations}
Recall that a neural network is represented as a graph $G$ that defines a function $f_G : \R^n \to \R^m$.
We define the \emph{input--output relation} of $f_G$ as the binary relation $\iorel_G$ containing every possible input and its corresponding output after executing $f_G$.
Formally, the input--output relation of $f_G$ is:
\[
\iorel_G = \{(\vec{a},\vec{b}) \mid \vec{a} \in \R^n, \; \vec{b} = f_G(\vec{a}) \}    
\]
We will similarly use $\iorel_\node$ to define the input--output relation of the function $f_\node$ of a single node $\node$ in $G$.
\begin{example}
    Consider the simple function $f_G(x) = x+1$.
    Its input--output relation is $$\iorel_G = \{(a,a+1) \mid a \in \R\}$$
\end{example}

\subsection*{Encoding a Single Node, Illustrated}
We begin by considering the case of a single node $\node$
and the associated function $f_\node : \R \to \R$.
A node with a single input is illustrated as follows
(recall that, by definition, a node in our neural network can only produce a single real-valued output):
\begin{figure}[h]
    \begin{center}
        \begin{tikzpicture}

            \draw node at (0, 0) [empty] (in) {};
            \draw node at (2, 0) [oper] (op) {$\node$};
            \draw node at (4, 0) [empty] (out) {};

            \draw[->,thick] (in) -- (op);
            \draw[->,thick] (op) -- (out);

        \end{tikzpicture}
    \end{center}
\end{figure}

Say $f_\node(x) = x + 1$.
Then, we can construct the following formula in \lra to model the relation $\iorel_\node = \{(a,a+1) \mid a\in \R\}$:
$$\varphi_\node \; \triangleq \; \outv{\node} = \inv{\node}{1} + 1$$
where $\outv{\node}$ and $\inv{\node}{1}$ are real-valued variables.
The symbol $\outv{\node}$ denotes the output of node $\node$
and $\inv{\node}{1}$ denotes its first input (it only has a single input).

Consider the models of $\varphi_\node$;
they are all of the form:
\[
\{\inv{\node}{1} \mapsto a,\; \outv{\node} \mapsto a + 1\}
\]
for any real number $a$.
We can see a clear one-to-one correspondence between elements of $\iorel_\node$ and models of $\varphi_\node$.

Let's now take a look at a node $\node$ with two inputs;
assume that $f_\node(\vec{x}) = x_1 + 1.5x_2$.
\begin{figure}[h]
    \begin{center}
        \begin{tikzpicture}

            \draw node at (0, 0) [empty] (in) {};
            \draw node at (0, 1) [empty] (in2) {};
            \draw node at (2, 0.5) [oper] (op) {$\node$};
            \draw node at (4, 0.5) [empty] (out) {};

            \draw[->,thick] (in) -- (op);
            \draw[->,thick] (in2) -- (op);
            \draw[->,thick] (op) -- (out);

        \end{tikzpicture}
    \end{center}
\end{figure}

\noindent
The encoding $\varphi_\node$ is as follows:
$$\varphi_\node \; \triangleq \; \outv{\node} = \inv{\node}{1} + 1.5\inv{\node}{2}$$
Observe how the elements of the input vector, $x_1$ and $x_2$, correspond to the two real-valued variables $\inv{\node}{1}$ and $\inv{\node}{2}$.

\subsection*{Encoding a Single Node, Formalized}
Now that we have seen a couple of examples, let's formalize the process of encoding the operation $f_\node$ of some node $\node$.
We will assume that $f_\node : \R^{n_\node} \to \R$ is piecewise-linear, i.e., of the form
\[ f(\vec{x}) = \begin{cases*}
    \sum_j c_j^1 \cdot x_j + b^1 & if $S_1$  \\
    ~~~~~~~~~~~~~~~\vdots &\\
    \sum_j c_j^l\cdot x_j + b^l & if $S_l$ 
 \end{cases*} 
\]
where $j$ ranges from $1$ to $n_\node$.
We will additionally assume that each condition $S_i$ is defined as a formula in \lra over the elements of the input $\vec{x}$.
Now, the encoding is as follows:
\[
\varphi_\node \; \triangleq \; \bigwedge_{i=1}^l \left[S_i \Rightarrow \left(\outv{\node} = \sum_{j=1}^{n_\node} c_j^i \cdot \inv{\node}{j} + b^i\right) \right]
\]
The way to think of this encoding is as a combination of \emph{if} statements: if $S_i$ is true, then $\outv{\node}$ is equal to the $i$th inequality.
The implication ($\Rightarrow$) gives us the ability to model a conditional, where the left side of the implication is the condition and the right side is the assignment.
The big conjunction on the outside of the formula, $\bigwedge_{i=1}^l$, essentially combines if statements: \emph{``if $S_1$ then ...} \textsc{and} \emph{if $S_2$ then ...} \textsc{and} \emph{if $S_3$ ...''}

\begin{example}
    The above encoding is way too general with too many superscripts and subscripts. Here's a simple and practical example, the ReLU function:
    \[ \relu(x) = \begin{cases*}
        x & if $x > 0$  \\
        0 & if $x \leq 0$ 
     \end{cases*} 
    \]
    A node $\node$ such that $f_\node$ is ReLU would be encoded as follows:
    \[
    \varphi_\node \; \triangleq \; (\underbrace{\inv{\node}{1} > 0}_{x > 0} \Rightarrow  \outv{\node} = \inv{\node}{1}) \; \land \; (\underbrace{\inv{\node}{1} \leq 0}_{x \leq 0} \Rightarrow  \outv{\node} = 0) 
    \]
    
\end{example}

\subsection*{Soundness and Completeness}

The above encoding precisely captures the semantics
of a piecewise-linear node. 
Let's formally capture this fact:
Fix some node $\node$ with a piecewise-linear function $f_\node$.
Let $\varphi_\node$ be its encoding, as defined above.

First, our encoding is \emph{sound}:
any execution of the node is captured by a model of the formula $\varphi_\node$.
Informally, soundness means that our encoding does not miss any behavior of $f_\node$.
Formally, let $(\vec{a},b) \in \iorel_\node$
    and let 
    \[
    I = \{\inv{\node}{1} \mapsto a_1,\ \ldots, \inv{\node}{n} \mapsto a_n, \  \outv{\node} \mapsto b\}    
    \]
    Then, $I\models \varphi_\node$.

Second, our encoding is \emph{complete}: 
any model of $\varphi_\node$ maps to  a behavior of $f_\node$.
Informally, completeness means that our encoding is tight, or does not introduce new behaviors not exhibited by $f_\node$.
Formally,
    let the following be a model of $\varphi_\node$:
    \[
    I = \{\inv{\node}{1} \mapsto a_1,\ \ldots, \inv{\node}{n} \mapsto a_n,\  \outv{\node} \mapsto b\}    
    \]
    Then, $(\vec{a},b) \in \iorel_\node$.

\section{Encoding a Neural Network}\label{sec:encoding}
We have shown how to encode a single node of a neural network.
We're now ready to encode the full-blown graph.
The encoding is broken up into two pieces:
(1) a formula encoding semantics of all nodes,
and (2) a formula encoding the connections between them, i.e., the edges.

\subsection*{Encoding the Nodes}
Recall that a neural network is a graph $G = (\nodes,\edges)$, where the set of nodes $\nodes$ contains input nodes $\inodes$, which do not perform any operations.
The following formula combines the encodings of all non-input nodes in $G$:
\[
\nodesf \triangleq \bigwedge_{\node \in \nodes \setminus \inodes} \varphi_\node   
\]
Again, the big conjunction essentially says, \emph{``the output of $v_1$ is ...} \textsc{and} \emph{the output of node $v_2$ is ...} \textsc{and} \emph{...''}
This formula, however, is meaningless on its own:
it simply encodes the input--output relation of every node, but not the connections between them!

\subsection*{Encoding the Edges}
Let's now encode the edges.
We will do this for every node individually, encoding all of its incoming edges.
Fix some node $\node \in \nodes \setminus \inodes$.
Let $(\node_1,\node), \ldots (\node_n,\node)$ be an ordered sequence of all edges whose target is $\node$.
Recall that in \cref{sec:dags}, we assumed that there is a total ordering on edges.
The reason for this ordering is to be able to know which incoming edges feed into which inputs of a node.
Informally, the edge relation $E$ gives us a bunch of wires to be plugged into node $\node$; the ordering  tells us where to plug those wires---the first wire in the first socket, the second wire in the second socket, and so on.

We can now define a formula for edges of $\node$:
\[
\edgesf{\node} \;\triangleq\; \bigwedge_{i=1}^n \inv{\node}{i} = \outv{\node}_i    
\]
Intuitively, for each edge $(\node_i,\node)$, we connect the output of node $\node_i$ with the $i$th input of $\node$.
We can now define $\edgesfall$ as the conjunction of all incoming edges of all non-input nodes:
\[
\edgesfall \;\triangleq\; \bigwedge_{\node \in \nodes \setminus \inodes} \edgesf{\node}
\]

\subsection*{Putting it all Together}
Now that we have shown how to encode nodes and edges, there is nothing left to encode! So let's put things together. Given a graph $G = (V,E)$, we will define its encoding as follows:
\[
\graphf{G} \; \triangleq \; \nodesf \land \edgesfall     
\]

Just as for the single-node encoding, we get soundness and completeness.
Let $\iorel_G$ be the input--output relation of $G$.
Soundness means that $\graphf{G}$ does not miss any of the behaviors in $\iorel_G$.
Completeness means that every model of $\graphf{G}$
maps to an input--output behavior of $G$.

Note that the size of the encoding is linear in the size of the neural network (number of nodes and edges). Simply, each node gets a formula and each edge gets a formula. The formula of node $\node$ is of size linear in the size of the (piecewise) linear function $f_\node$. 

\subsection*{Correctness of the Encoding}
Assume we that have 
the following ordered input nodes in $\inodes$
$$v_1,\ldots,v_n$$
and the following output nodes in $\onodes$
$$v_{n+1},\ldots,v_{n+m}$$

Our encoding is sound and complete.
First, let's state soundness:
    Let $(\vec{a},\vec{b}) \in \iorel_G$
    and let 
    \[
    I = \{\outv{\node}_1  \mapsto a_1, \ldots, \outv{\node}_n \mapsto a_n\} \cup \{\outv{\node}_{n+1} \mapsto b_1, \ldots, \outv{\node}_{n+m} \mapsto b_m\}    
    \]
    Then, there exists $I'$ such that $I\cup I' \models \graphf{G}$.

Notice that, unlike the single-node setting, the model of $\graphf{G}$ not only contains assignments to inputs and outputs of the network, but also the intermediate nodes.
This is taken care of using $I$,
which assigns values to the outputs of input and output nodes,
and $I'$, which assigns the inputs and outputs of all nodes and therefore its domain does not overlap with $I$.

Similarly, completeness is stated as follows:
    Let the following be a model of $\graphf{G}$:
    \[
        I = \{\outv{\node}_1 \mapsto a_1, \ldots, \outv{\node}_n \mapsto a_n\} \cup \{\outv{\node}_{n+1} \mapsto b_1, \ldots, \outv{\node}_{n+m} \mapsto b_m\} \cup I' 
    \]
    Then, $(\vec{a},\vec{b}) \in \iorel_G$.

\subsection*{An Example Network and its Encoding}
Enough abstract mathematics. Let's look at a concrete example neural network $G$.

\begin{center}
    \begin{tikzpicture}

        \draw node at (0, 0) [input] (in) {$\node_1$};
        \draw node at (0, -1) [input] (in2) {$\node_2$};

        \draw node at (2, -0.5) [oper] (op) {$\node_3$};
        \draw node at (4, -0.5) [output] (out) {$\node_4$};

        \draw[->,thick] (in) -- (op);
        \draw[->,thick] (in2) -- (op);
        \draw[->,thick] (op) -- (out);
    \end{tikzpicture}
\end{center}
\noindent
Assume that $f_{\node_3}(\vec{x}) = 2x_1 + x_2$
and $f_{\node_4}(x) = \relu(x)$.

We begin by constructing formulas for non-input nodes:
\begin{align*}
\varphi_{\node_3} & \; \triangleq \; 
\outv{\node}_3 = 2\inv{\node}{1}_3 + \inv{\node}{2}_3   \\
\varphi_{\node_4} & \; \triangleq \; (\inv{\node}{1}_4 > 0 \Longrightarrow  \outv{\node}_4 = \inv{\node}{1}_4) \; \land \; (\inv{\node}{1}_4 \leq 0 \Longrightarrow  \outv{\node}_4 = 0)
\end{align*}
Next, we construct edge formulas:
\begin{align*}
\edgesf{\node_3} & \; \triangleq \; (\inv{\node}{1}_3 = \outv{\node}_1) \land (\inv{\node}{2}_3 = \outv{\node}_2)\\
\edgesf{\node_4} & \; \triangleq \; \inv{\node}{1}_4 = \outv{\node}_3
\end{align*}
Finally, we conjoin all of the above formulas to arrive at the complete encoding of $G$:
\[
\graphf{G} \; \triangleq \; \underbrace{\varphi_{\node_3} \land \varphi_{\node_4}}_{\nodesf} \land   \underbrace{\edgesf{\node_3}  \land \edgesf{\node_4}}_{\edgesfall}
\]

\section{Handling Non-linear Activations}
In the above, we have assumed that all of our nodes are associated with piecewise-linear functions, allowing us to precisely capture their semantics in linear real arithmetic.
How can we handle non-piecewise-linear activations, like sigmoid and tanh?
One way to encode them is by \emph{overapproximating} their behavior, which gives us soundness but \emph{not} completeness.
As we will see, soundness means that our encoding can find proofs of correctness properties, and completeness means that our encoding can find counterexamples to correctness properties.
So, by overapproximating an activation function, we give up on counterexamples.

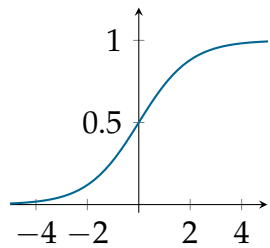
\begin{figure}[t]
    \centering
    \begin{tikzpicture}
        \begin{axis}[
            axis lines=middle,
            xmax=5,
            xmin=-5,
            ymin=-0.05,
            ymax=1.2,
            ytick={0,.5,1},
            width=5cm
        ]

        \addplot [domain=-10:10, samples=100,
                  thick, MidnightBlue] {1/(1+exp(-x)};
    
    \end{axis}
    \end{tikzpicture}
\caption{Sigmoid function}\label{fig:sigmoidnorect}
\end{figure}

\subsection*{Handling Sigmoid}
Let's begin with the concrete example of the sigmoid activation:
$$\sigma(x) = \frac{1}{1+\exp(-x)}$$
which is shown in \cref{fig:sigmoidnorect}.
The sigmoid function is (strictly) monotonically increasing,
so if we have two points $a_1 < a_2$,
we know that $\sigma(a_1) < \sigma(a_2)$.
We can as a result overapproximate the behavior of $\sigma$ by saying: 
for any input between $a_1$ and $a_2$, the output of the function can be \emph{any} value between $\sigma(a_1)$ and $\sigma(a_2)$.

Consider \cref{fig:sigmoidrect}.
Here we picked three points on the sigmoid curve, shown in red,
with $x$ coordinates $-1$, $0$, and $1$.
The red rectangles define the lower and upper bound on the output of the sigmoid function between two values of $x$.
For example, for inputs between $0$ and $1$, the output of the function is any value between $0.5$ and $0.73$.
For inputs more than $1$, we know that the output must be between $0.73$ to $1$ (the range of $\sigma$ is upper bounded by $1$).

Say for some node $\node$, $f_\node$ is a sigmoid activation.
Then, one possible encoding, following the approximation in \cref{fig:sigmoidrect}, is as follows:
\begin{align*}
\varphi_\node \; \triangleq \; & (\inv{\node}{1} \leq -1 \; \Longrightarrow \; 0 < \outv{\node} \leq 0.26)\\
&  \land (-1 < \inv{\node}{1} \leq 0 \; \Longrightarrow \;  0.26 < \outv{\node}  \leq 0.5)\\
&  \land (0 < \inv{\node}{1} \leq 1 \; \Longrightarrow \;  0.5 < \outv{\node}  \leq 0.73)\\
&  \land (\inv{\node}{1} > 1 \; \Longrightarrow \; 0.73 < \outv{\node}  < 1)
\end{align*}
Each conjunct specifies a range of inputs (left of implication) and the possible outputs in that range (right of implication).
For example, the first conjunct specifies that, for inputs $\leq -1$, the output can be any value between $0$ and $0.26$. 

\subsection*{Handling any Monotonic Function}
We can generalize the above process to any monotonically (increasing or decreasing) function $f_\node$.

Let's assume that $f_\node$  is monotonically increasing.
We can pick a sequence of real values
$c_1 < \cdots < c_n$.
Then, we can construct the following encoding:
\begin{align*}
\varphi_\node \; \triangleq \; &
    (\inv{\node}{1} \leq c_1 \Longrightarrow \emph{lb} < \outv{\node} \leq f_\node(c_1))\\
    & \land
    (c_1 < \inv{\node}{1} \leq c_2 \Longrightarrow f_\node(c_1) < \outv{\node} \leq f_\node(c_2))\\
    &~~ \vdots  \\
    & \land (c_n < \inv{\node}{1}  \Longrightarrow  f_{\node}(c_n) < \outv{\node} \leq \emph{ub})
\end{align*}
where $\emph{lb}$ and $\emph{ub}$ are the lower and upper bounds of the range of $f_\node$;
for example, for sigmoid, they are 0 and 1, respectively.
If a function is unbounded, then we can drop the constraints 
$\emph{lb} \leq \outv{\node}$ and $\outv{\node} \leq \emph{ub}$.

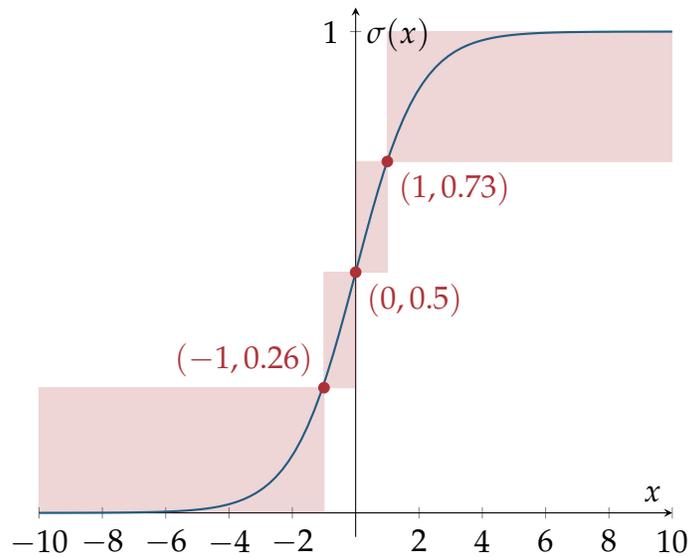
\begin{figure}[t]
    \centering
\begin{tikzpicture}
    \begin{axis}[
        axis lines=middle,
        xmax=10,
        xmin=-10,
        ymin=-0.05,
        ymax=1.05,
        xlabel={$x$},
        ylabel={$\sigma(x)$},
        ytick={0,1},
        width=10cm
    ]

    \addplot [domain=-10:10, samples=100,
              thick, MidnightBlue] {1/(1+exp(-x)};
    
    \addplot[Maroon,mark=*] coordinates {(0,0.5)}
    node [pos=0.3,  below right] {$(0,0.5)$};

    \addplot[Maroon,mark=*] coordinates {(1,0.73)}
    node [pos=0.3,  below right] {$(1,0.73)$};

    \addplot[Maroon,mark=*] coordinates {(-1,0.26)}
    node [pos=0.3, above left] {$(-1,0.26)$};
    
    \draw [Maroon, fill=Maroon, opacity=0.2] (axis cs:-1,0.26) rectangle (axis cs:0,0.5);

    \draw [Maroon, fill=Maroon, opacity=0.2] (axis cs:0,0.5) rectangle (axis cs:1,0.73);

    \draw [Maroon, fill=Maroon, opacity=0.2] (axis cs:1,0.73) rectangle (axis cs:10,1);

    \draw [Maroon, fill=Maroon, opacity=0.2] (axis cs:-10,0) rectangle (axis cs:-1,0.26);

    \end{axis}
\end{tikzpicture}
\caption{Sigmoid function with overapproximation}\label{fig:sigmoidrect}
\end{figure}

The more points $c_i$ we choose and the closer they are to each other, the better our approximation is.
This encoding is sound but \emph{incomplete},
because it captures more behaviors than conceivable from the activation function.
In our sigmoid example, for input $\geq 1$, the encoding says that the output of the sigmoid is \emph{any} value between $0.73$ and $1$, as indicated by the right-most shaded area in \cref{fig:sigmoidrect}.

\section{Encoding Correctness Properties}
Now that we have shown how to encode the semantics of neural networks as logical constraints, we're ready for the main dish: encoding entire correctness properties.

\subsection*{Checking Robustness Example}
We begin with a concrete example before seeing the general form.
Say we have a neural network $G$ defining a binary classifier $f_G : \R^n \to \R^2$. The neural network $f_G$ takes a grayscale image as a vector of reals, between $0$ and $1$, describing the intensity of each pixel (black to white), and predicts whether the image is of a \emph{cat} or a \emph{dog}.
Say we have an image $\vec{c}$ that is correctly classified as \emph{cat}. We want to prove that a small perturbation to the brightness of $\vec{c}$ does not change the prediction.
We formalize this as follows:
\[
\begin{array}{c}
    \pre{|\vec{x} - \vec{c}| \leq \vec{0.1}}\\
    \vec{r} \gets f_G(\vec{x})\\
    \post{r_1 > r_2}
\end{array}
\]
where the first output, $r_1$, is the probability of \emph{cat}, while $r_2$ is the probability of \emph{dog}.

The high-level intuition for the encoding of this correctness property follows how the property is written.
%
%
The formula that we generate to check this statement, called the \emph{verification condition} (VC), looks roughly like this:
\begin{displayquote}
    (precondition $\land$ neural network) 
    $\Longrightarrow$ postcondition 
\end{displayquote}
If this formula is \emph{valid}, 
then the correctness property holds.

Let's assume for our example that 
the input nodes of the neural network are $\{\node_1,\ldots,\node_n\}$ and the output nodes are $\{\node_{n+1}, \node_{n+2}\}$.
Assume also that the formula $\graphf{G}$ encodes the network, as described earlier in this chapter.
We encode the correctness property as follows:
\[
\underbrace{
\left(\bigwedge_{i=1}^n |x_i - c_i| \leq 0.1\right)
}_\text{precondition}
\land 
\underbrace{\graphf{G}}_\text{network}
\land
\underbrace{
\left(\bigwedge_{i=1}^n x_i = \outv{\node}_i\right)
}_\text{network input}
\land
\underbrace{
\left(r_1 = \outv{\node}_{n+1} \land r_2 = \outv{\node}_{n+2}\right)
}_\text{network output}
\]
\[
\Longrightarrow
\underbrace{
r_1 > r_2
}_\text{postcondition}
\]
Here's the breakdown:
\begin{itemize}
\item The precondition is directly translated to an \lra formula. Since \lra formulas don't natively support vector operations, we decompose the vector into its constituent scalars.
Note that the absolute-value operation $|\cdot|$ is not present natively in \lra, but, fear not, it is actually encodable:
A linear inequality with absolute value, like $|x| \leq 5$, can be written in \lra as $x \leq 5 \land -x \leq 5$.
\item The network is encoded as a formula $\graphf{G}$,
just as we saw earlier in \cref{sec:encoding}.
The trick is that we now also need to connect the variables of $\graphf{G}$ with the inputs $\vec{x}$ and output $\vec{r}$.
This is captured by the two subformulas labeled ``network input'' and ``network output''.
\item The postcondition is encoded as is.
\end{itemize}

\subsection*{Encoding Correctness, Formalized}
A correctness property is of the form
\[
\begin{array}{c}
    \pre{P}\\
    \vec{r_1} \gets f_{G_1}(\vec{x_1})\\
    \vec{r_2} \gets f_{G_2}(\vec{x_2})\\
    \vdots\\
    \vec{r_l} \gets f_{G_l}(\vec{x_l})\\
    \post{Q}
\end{array}
\]
While most examples we see in the book involve a single neural network $f_G$,
recall from \cref{ch:correctness} that our properties allow us to consider a collection of networks.

Assume that the precondition and postcondition 
are encodable in \lra.
We then encode the verification condition as follows:
\[
\left(P \land \bigwedge_{i=1}^l \varphi_i \right)\Longrightarrow Q     
\]
where $\varphi_i$ is the encoding of the $i$th assignment
$\vec{r_i} \gets f_{G_i}(\vec{x_i})$.
The assignment encoding $\varphi_i$ combines the encoding of the neural network $\graphf{G_i}$ along with \emph{connections} with inputs and outputs, $\vec{x_i}$ and $\vec{r_i}$, respectively:
\[
\varphi_i \triangleq \graphf{G_i} \land \left(\bigwedge_{j=1}^n x_{i,j} = \outv{\node}_i\right)
\land \left(
    \bigwedge_{j=1}^m
    r_{i,j} = \outv{\node}_{n+j} 
\right)
\]
Here we make two assumptions:

\begin{itemize}
 \item   The input and output variables of the encoding of $G_i$ are 
$v_1,\ldots,v_n$ and $v_{n+1},\ldots,v_{n+m}$, respectively.
\item Each graph $G_i$ has unique nodes and therefore input--output variables.
\end{itemize}

Informally, we can think of our encoding, $\left(P \land \bigwedge_{i=1}^l \varphi_i \right)\Longrightarrow Q     $, as saying the following:
``\emph{if the precondition is true} \textsc{and} \emph{we execute all $l$ networks, then the postcondition should be true}''

\subsection*{Soundness and Completeness}
Say we have a correctness property that we have encoded as a formula $\varphi$.
Then, we have the following soundness guarantee:
    If $\varphi$ is valid, then the correctness property is true.

Completeness depends on whether all functions are encodable in \lra.
Assuming all functions are encodable in \lra, then, if $\varphi$ is invalid, we know that there is a model $I \models \neg F$.
This model is a counterexample to the correctness property.
From this model, we can read values for the input variables that result in outputs that do not satisfy the postcondition.
This is best seen through an example:

\begin{example}
Take the following simple correctness property,
where $f(x) = x$:
\[
\begin{array}{c}
    \pre{|x - 1| \leq 0.1}\\
    r \gets f(\vec{x})\\
    \post{r \geq 1}
\end{array}
\]
This property is not true.
Let $x = 0.99$; this satisfies the precondition.
But, $f(0.99) = 0.99$, which is less than 1.
If we encode a formula $\varphi$ for this property,
then we will have a model $I \models \neg \varphi$
such that $x$ is assigned 0.99.
\end{example}

\section*{Looking Ahead}
Ahh, this chapter was tiring!
Thanks for sticking around.
We have taken neural networks, with all their glory, and translated them into formulas.
In the coming chapters, we will study algorithms for checking satisfiability of these formulas.

To my knowledge, the first encoding of neural networks 
as constraints for verification is due to \citet{DBLP:conf/cav/PulinaT10}, predating the current explosion in interest.
\citet{DBLP:conf/nips/BastaniILVNC16}
were the first to view robustness verification 
as a constraint-solving problem.

Our encoding of sigmoid follows that of \citet{ehlers2017formal}.
A number of papers have considered \milp encodings that are similar 
to the ones we give~\citep{tjeng2018evaluating}. In \milp, we don't have disjunction,
so we simulate disjunction with an integer that can take the values $\{0,1\}$.
The main issue with \lra and \milp encodings is disjunction;
with no disjunction, the problem is polynomial-time solvable.
Disjunctions mostly arise due to ReLUs.
We say that a ReLU is active if its output is $>0$ and inactive otherwise.
If a ReLU is active or inactive for all possible inputs to the neural network,
as prescribed by the precondition, then we can get rid of the disjunction,
and treat it as the function $f(x) = 0$ (inactive) or $f(x) = x$ (active).
With this idea in mind, there are two tricks to simplify verification:
(1) We can come up with lightweight techniques to discover which ReLUs are active or inactive. The abstraction-based verification techniques discussed in Part III of the book can be used.
(2) Typically, when we train neural networks, we aim to maximize accuracy on some training data; we can additionally bias the training towards neural networks where most of the ReLUs are always active/inactive~\citep{DBLP:conf/iclr/TjengXT19}.

In practice, neural networks are implemented using finite-precision arithmetic, where real numbers are approximated as floating-point numbers, fixed-point numbers, or even machine integers.
Some papers carefully ensure that verification results hold for a floating-point implementation of the network~\citep{katz2017reluplex}.
A recent paper has shown that verified neural networks in \lra may not really be robust when one considers the bit-level behavior~\citep{jia2020exploiting}.
A number of papers have also considered bit-level verification of neural networks, using propositional logic instead of \lra~\citep{jia2020efficient,narodytska2018verifying}.

     \chapter{DPLL Modulo Theories}\label{ch:dp}

In the previous chapter, we saw how to reduce the verification 
problem to that of checking satisfiability of a logical formula.
But how do we actually check satisfiability?
In this chapter, we will meet the \dpll (Davis--Putnam--Logemann--Loveland) algorithm,
which was developed decades ago for checking satisfiability
of Boolean formulas.
Then we will see an extension of \dpll that can handle first-order
formulas over theories.
These algorithms underlie modern \sat and \smt solvers.
I'll give a complete description of \dpll in the sense
that you can follow the chapter and implement a working algorithm.
But note that there are numerous data structures, implementation tricks, and heuristics
that make \dpll \emph{really} work in practice, and we will not cover those here. (At the end of the chapter, I point you to additional resources.)

\section{Conjunctive Normal Form (CNF)}
The goal of \dpll is to take a Boolean formula $\varphi$
and decide whether it is \sat or \unsat.
In case $\varphi$ is \sat, \dpll should also return a
model of $\varphi$.
We begin by talking about the \emph{shape} of the formula \dpll
expects as input.

\dpll expects formulas to be in \emph{conjunctive normal form} (\cnf).
Luckily, all Boolean formulas can be rewritten into equivalent formulas in \cnf.
(We will see how later in this chapter.)
A formula $\varphi$ in \cnf is of the following form:
\begin{align*}
    C_1 \land \cdots \land C_n
\end{align*}
where each subformula $C_i$ is called a \emph{clause}
and is of the form
\begin{align*}
    \ell_1 \lor \cdots \lor \ell_{m_i}
\end{align*}
where each $\ell_i$ is called a \emph{literal}
and is either a variable, e.g., $p$, or its negation, e.g., $\neg p$.

\begin{example}
    The following is a \cnf formula with two clauses, each of which contains two literals:
    \begin{align*}
        (p \lor \neg r) \land (\neg p \lor q)
    \end{align*}
    The following formula is \emph{not} in \cnf:
    \begin{align*}
        (p \land q) \lor (\neg r)
    \end{align*}
\end{example}

\section{The DPLL Algorithm}
The completely na\"ive way to decide satisfiability 
is by trying every possible interpretation and checking if it is a model of the formula.
Of course, there are exponentially many interpretations in the number of variables.
\dpll tries to avoid doing a completely blind search,
but, on a bad day, it will just devolve into an exponential enumeration of 
all possible interpretations---after all, satisfiability is the canonical NP-complete problem.

\dpll alternates between two phases: \emph{deduction} and \emph{search}.
Deduction tries to simplify the formula using the laws of logic.
Search simply searches for an interpretation.

\subsection*{Deduction}
The deduction part of \dpll is called \emph{Boolean constant propagation} (\bcp).
Imagine that you have the following formula in \cnf:
$$(\ell) \land C_2 \land \cdots C_n$$
Notice that the first clause consists of a single literal---we call it a \emph{unit clause}.
Clearly, any model of this formula must assign $\ell$ the value $\true$:
Specifically, if $\ell$ is a variable $p$, then $p$ must be assigned $\true$;
if $\ell$ is a negation $\neg p$, then $p$ must be assigned $\false$.

The \bcp phase of \dpll will simply look for all unit clauses
and replace their literals with $\true$.
\bcp is enabled by the fact that formulas are in \cnf,
and it can be quite effective at proving \sat or \unsat.

\begin{example}
    Consider the following formula 
    $$\varphi \triangleq (p) \land (\neg p \lor r) \land (\neg r \lor q)$$
    \bcp first finds the unit clause $(p)$
    and assigns $p$ the value $\true$.
    This results in the following formula:
    \begin{align*}
        & (\true) \land (\neg \true \lor r) \land (\neg r \lor q) \\
        \equiv~ & (r) \land (\neg r \lor q)
    \end{align*}
    Clearly \bcp's job is not done: the simplification has produced
    another unit clause, $(r)$.
    \bcp sets $r$ to $\true$, resulting in the following formula:
    \begin{align*}
        & (\true) \land (\neg \true \lor q) \\
        \equiv~ & (q)
    \end{align*}
    Finally, we're left with a single clause, the unit clause $(q)$.
    Therefore \bcp assigns $q$ the value $\true$,
    resulting in the final formula $\true$.
    This means that $F$ is \sat,
    and $\{p \mapsto \true, q \mapsto \true, r \mapsto \true\}$
    is a model.
\end{example}

In the above example, \bcp managed to show satisfiability of $\varphi$.
Note that \bcp can also prove unsatisfiability, by simplifying the formula to $\false$.
But \bcp might get stuck if it cannot find unit clauses.
This is where search is employed by \dpll.

\subsection*{Deduction + Search}
\cref{alg:dpll} shows the entire \dpll algorithm.
The algorithm takes a formula $\varphi$ in \cnf.

The first part performs \bcp:
the algorithm keeps simplifying the formula
until no more unit clauses exist.
We use the notation $\varphi[\ell \mapsto \true]$
to mean replace all occurrences of $\ell$ in $\varphi$ with $\true$
and simplify the resulting formula.
Specifically, if $\ell$ is a variable $p$, then all occurrences of $p$
are replaced with $\true$; if $\ell$ is a negation $\neg p$, then all occurrences of $p$ are replaced with $\false$.

After \bcp is done, the algorithm 
checks if the formula is $\true$,
which means that \bcp has proven that the formula is \sat.

If \bcp is unsuccessful in proving \sat, 
then \dpll moves to the search phase:
it iteratively chooses variables and tries to replace them 
with $\true$ or $\false$, calling \dpll
recursively on the resulting formula.
The order in which variables are chosen in search
is critical to \dpll's performance.
There is a lot of research on variable selection.
One of the popular heuristics maintains a continuously updated 
score sheet,
where variables with higher scores are chosen first.

\SetEndCharOfAlgoLine{}
\begin{algorithm}[t]
  \KwData{A formula $\varphi$ in \cnf form}
  \KwResult{$I \models \varphi$ or \unsat}
  ~\\
  \acomment{Boolean constant propagation (\bcp)} \\
  \While {there is a unit clause $(\ell)$ in $\varphi$}{
      Let $\varphi$ be $\varphi[\ell \mapsto \true]$\\
  }
  ~\\
  \lIf{$\varphi$ is $\true$}{\Return \sat}
  ~\\
  \acomment{Search}\\
  \For{every variable $p$ in $\varphi$}{
    \textbf{If} $\dpll(\varphi[p \mapsto \true])$ is \sat  \textbf{then} \Return \sat\\
    \textbf{If} $\dpll(\varphi[p \mapsto \false])$ is \textrm{\sat} \textbf{then} \Return \sat
  }
  \Return \unsat
  ~\\
  ~\\
  \acomment{The model $I$ that is returned by \dpll when the result is \sat is maintained implicitly in the sequence of assignments to variables (of the form $[l \mapsto \cdot]$ and $[p \mapsto \cdot]$)}
  \caption{\dpll}\label{alg:dpll}
 \end{algorithm}

\cref{alg:dpll}, as presented, returns \sat when the input formula is satisfiable, but does not return a model.
The model $I$ that is returned by \dpll when the result is \sat is maintained implicitly in the sequence of assignments to variables (of the form $[l \mapsto \cdot]$ and $[p \mapsto \cdot]$) made by \bcp
and search that led to \sat being returned.
The algorithm returns \unsat when it has exhausted all possible satisfying 
assignments.

\begin{example}
    Consider the following formula given to \dpll
    $$\varphi \triangleq (p \lor r) \land (\neg p \lor q) \land (\neg q \lor \neg r)$$
    \begin{description}
    \item[First level of recursion]
    \dpll begins by attempting \bcp, which cannot find any unit clauses.
    Then, it proceeds to search.
    Suppose that search chooses variable $p$,
    setting it to $\true$ by invoking \dpll recursively on 
    \begin{align*}
        \varphi_1 = \varphi[p \mapsto \true] = q \land (\neg q \lor \neg r)
    \end{align*}
        
    \item[Second level of recursion]
    Next, \dpll attempts \bcp on $\varphi_1$.
    First, it sets $q$ to $\true$,
    resulting in the formula
    $$\varphi_2 = \varphi_1[q \mapsto true] = (\neg r)$$
    Then, it sets $r$ to $\false$,
    resulting in $$\varphi_3 = \varphi_2[r \mapsto \false]
    = \true$$
    Since $F_3$ is $\true$, \dpll returns \sat.
    Implicitly, \dpll has built up a model of $\varphi$:
    $$\{p \mapsto \true, q \mapsto \true, r \mapsto \false\}$$
\end{description}
\end{example}

\subsection*{Partial Models}
Note that \dpll may terminate with \sat 
but without assigning every variable in the formula.
We call the resulting model a \emph{partial model}.
You can take a partial model and \emph{extend} it by assigning
the remaining variables in any way you like, 
and you'll still have a model of the formula.

\begin{example}
    Consider this simple formula:
    $$\varphi \triangleq p \land (q \lor p \lor \neg r) \land (p \lor \neg q)$$
    The first thing \dpll will do is apply \bcp, 
    which will set the unit clause $p$ to $\true$.
    The rest of the formula then simplifies to $\true$.
    This means that $q$ and $r$ are useless variables---give them any interpretation and you'll end up with a model, as long as $p$ is assigned $\true$.
    Therefore, we call $I = \{p \to \true\}$
    a partial model of  $\varphi$.
    Formally, $\eval(I(\varphi)) = \true$.
\end{example}

\section{DPLL Modulo Theories}
We have seen how \dpll can decide satisfiability of
Boolean formulas.
We now present \dpll \emph{modulo theories}, or \dpllt,
an extension of \dpll
that can handle formulas over, for example,
arithmetic theories like \lra.
The key idea of \dpllt is to start by treating a formula as if it is completely 
Boolean, and then incrementally add more and more
\emph{theory} information until we can conclusively 
say that a formula is \sat or \unsat.
We begin by defining the notion of Boolean abstraction of a formula.

\subsection*{Boolean Abstraction}
For illustration, we assume that we're dealing with formulas in \lra,
as with the previous chapters.
Say we have the following formula in \lra:
$$\varphi \triangleq (x \leq 0 \lor x \leq 10) \land (\neg x \leq 0)$$
The Boolean abstraction of $\varphi$, denoted $\varphi_B$,
is the formula where every unique linear inequality
in the formula is replaced with a special Boolean variable, as follows:
$$\varphi^B \triangleq (p \lor q) \land (\neg p)$$
The inequality $x \leq 0$ is \emph{abstracted} as $p$ and
$x \leq 10$ is abstracted as $q$.
Note that both occurrences of $x \leq 0$ are replaced by the same Boolean variable, though this need not be the case for the correctness of our exposition.
We will also use the superscript $T$ to map Boolean formulas back
to theory formulas, e.g., $(\varphi^B)^T$ is $\varphi$.

We call this process \emph{abstraction} because constraints are lost in the process; namely, the relation between different inequalities is obliterated.
Formally speaking, if $\varphi^B$ is \unsat, then $\varphi$ is \unsat.
But the converse does not hold: if $\varphi^B$ is \sat, it does not mean that
$\varphi$ is \sat.

\begin{example}
    Consider the formula $\varphi \triangleq x \leq 0 \land x \geq 10$.
    This formula is clearly \unsat.
    However, its abstraction, $p \land q$, is \sat.
\end{example}

\SetEndCharOfAlgoLine{}
\begin{algorithm}[t]
  \KwData{A formula $\varphi$ in \cnf form over theory $T$}
  \KwResult{$I \models \varphi$ or \unsat}
  ~\\
  Let $\varphi^B$ be the abstraction of $\varphi$\\
  \While{\emph{true}}{
  \textbf{If} $\dpll(\varphi^B)$ is \unsat \textbf{then} \Return \unsat
  
    Let $I$ be the  model returned by $\dpll(\varphi^B)$\\
    Assume $I$ is represented as a formula\\
    \eIf{$I^T$ is satisfiable (using a theory solver)}{\Return \sat\ {and the model returned by theory solver}}
    {
        Let $\varphi^B$ be $\varphi^B \land \neg I$
    }
  
  }

  \caption{\dpllt}\label{alg:dpllt}
 \end{algorithm}

\subsection*{Lazy DPLL Modulo Theories}
The \dpllt algorithm  takes a formula $\varphi$,
over some theory like \lra, and decides satisfiability.
\dpllt assumes access to a \emph{theory solver}.
The theory solver takes a conjunction of, for example, linear inequalities, and checks their satisfiability.
In a sense, the theory solver takes care of conjunctions and the \dpll algorithm
takes care of disjunctions.
In the case of \lra, the theory solver can be the Simplex algorithm,
which we will see in the next chapter.

\dpllt, shown in \cref{alg:dpllt}, works as follows:
First, using vanilla \dpll, it checks if the abstraction $\varphi^B$
is \unsat, in which case it can declare that $\varphi$ is \unsat,
following the properties of abstraction discussed above.
The tricky part comes when dealing with the case where $\varphi^B$
is \sat, because that does not necessarily mean that  $\varphi$ is \sat.
This is where the theory solver comes into play.
We take the model $I$ returned by $\dpll(\varphi^B)$ and map it to a formula $I^T$
in the theory; for example, if the theory we're working with is \lra, $I^T$ is a conjunction of linear inequalities.
If the theory solver deems $I^T$ satisfiable, then we know that $\varphi$
is satisfiable and we're done. 
Otherwise, \dpllt \emph{learns} the fact that $I$ is not a model.
So it negates $I$ and conjoins it to $\varphi^B$.
In a sense, the \dpllt \emph{lazily} learns more and more facts,
refining the abstraction, until it can decide \sat or \unsat. 

\begin{example}
Consider the following \lra formula $\varphi$:
$$x \geq 10 \land (x < 0 \lor y \geq 0)$$
and its abstraction $\varphi^B$:
$$p \land (q \lor r)$$
where $p$ denotes $x \geq 10$, $q$ denotes $x < 0$, and $r$ denotes $y \geq 0$.

\begin{description}
    \item[First iteration]
    \dpllt begins by invoking \dpll on $\varphi^B$.
    Suppose $\dpll$ returns the partial model $$I_1 = \{p \mapsto \true, q \mapsto \true\}$$
    We will represent $I_1$ as a formula $$p \land  q $$
    Next, we check if $I_1$ is indeed a model of $\varphi$.
    We do so by invoking the theory solver on $I^T_1$, which is
    $$\underbrace{x \geq 10}_p \land \underbrace{x < 0}_q$$
    The theory solver will say that $I^T_1$ is \unsat, because $x$ cannot be $\geq 10$ and $< 0$.
    Therefore, \dpllt blocks this model by conjoining $\neg I_1$ to $\varphi^B$.
    This makes $\varphi^B$ the following formula, which is still in \cnf,
    because $\neg I_1$ is a clause:
    $$p \land (q \lor r) \land \underbrace{(\neg p \lor \neg q)}_{\neg I_1}$$
    In other words, we're saying that we cannot have a model that sets
    both $x \geq 10$ and $x < 0$ to $\true$.
    \item[Second iteration] 
    In the second iteration, \dpllt invokes \dpll on the updated $\varphi^B$.
    $\dpll$ cannot give us the same model $I_1$.
    So it gives us another one, say $I_2 = p \land \neg q \land r$.
    The theory solver checks $I_2^T$, which is satisfiable,
    and returns its own theory-specific model, e.g., $\{x \mapsto 10,\ y \mapsto 0\}$. We're now done, and we return the model.
\end{description}
\end{example}

\section{Tseitin's Transformation}
We've so far assumed that formulas are in \cnf.
Indeed, we can take any formula and turn it into an equivalent formula in \cnf.
We can do this by applying DeMorgan's laws (see \cref{ch:fol}), by distributing disjunction over conjunction.
For example, $r \lor (p\land q)$ can be rewritten into the equivalent $(r \lor p) \land (r \lor q)$.
This transformation, unfortunately, can lead to an exponential explosion in the size of the formula.
It turns out that there's a simple technique, known as \emph{Tseitin's transformation}, that produces a formula of size linear in the size of the non-\cnf formula.

Tseitin's transformation takes a formula $\varphi$ and produces a \cnf formula $\varphi'$. The set of variables of $\varphi$ is a subset of the variables in $\varphi'$; i.e., Tseitin's transformation creates new variables.
Tseitin's transformation guarantees the following properties:
\begin{enumerate}
    \item Any model of $\varphi'$ is also a model of $\varphi$, if we disregard the interpretations of newly added variables.
    \item If $\varphi'$ is \unsat, then $\varphi$ is \unsat.
\end{enumerate}
Therefore, given a non-\cnf formula $\varphi$,
to check its satisfiability, we can simply invoke \dpll on $\varphi'$.

\subsection*{Intuition}
Tseitin's transformation is pretty much the same 
as rewriting a complex arithmetic expression in a program into
a sequence of instructions
where every instruction is an application of a single unary or binary operator.
(This is roughly what a compiler does when compiling a high-level program to assembly or an intermediate representation.)
For example, consider the following function (in Python syntax):
\begin{lstlisting}
    def f(x,y,z):
        return x + (2*y + 3)
\end{lstlisting}
The \emph{return} expression can be rewritten into a sequence 
of operations, each operating on one or two variables, as follows,
where \lstinline{t1}, \lstinline{t2}, and \lstinline{t3} are temporary
variables:
\begin{lstlisting}
    def f(x,y,z):
        t1 = 2 * y
        t2 = t1 + 3
        t3 = x + t2
        return t3
\end{lstlisting}
Intuitively, each subexpression is computed and stored in a temporary variable:
\lstinline{t1} contains the expression \lstinline{2*y},
\lstinline{t2} contains the expression \lstinline{2*y + 3},
and \lstinline{t3} contains the entire expression \lstinline{x + (2*y + 3)}.

\subsection*{Tseitin Step 1: NNF}
The first thing  that Tseitin's transformation does is to push negation inwards so that $\neg$ only appears next to variables.
E.g., $\neg(p\land r)$ is rewritten into $\neg p \lor \neg r$.
This is known as \emph{negation normal form} (\textsc{nnf}).
Any formula can be easily translated to \textsc{nnf}
by repeatedly applying the following  rewrite rules until we can't rewrite the formula any further:
\begin{align*}
    \neg (F_1 \land F_2)  & \rightarrow \neg F_1 \lor \neg F_2 \\
    \neg (F_1 \lor F_2)  & \rightarrow \neg F_1 \land \neg F_2 \\
    \neg \neg F & \rightarrow F
\end{align*} 
In other words, (sub)formulas that match the patterns on the left of $\rightarrow$ are transformed into the patterns on the right.
In what follows we assume formulas are in \textsc{nnf}.

\subsection*{Tseitin Step 2: Subformula Rewriting}
We define a subformula of $\varphi$ to be any subformula that  
 contains a disjunction or conjunction---i.e., we don't consider subformulas at the level of literals.

\begin{example}
The following formula $\varphi$ is decomposed into 4 subformulas:
$$\varphi \triangleq \underbrace{\underbrace{(p \land q)}_{\varphi_1} \lor \underbrace{(\underbrace{q \land \neg r}_{\varphi_2} \land s)}
_{\varphi_3}}_{\varphi_4}$$
$\varphi_1$ and $\varphi_2$ are in the deepest \emph{level of nesting},
while $\varphi_4$ is in the shallowest level.
Notice that $\varphi_2$ is a subformula of $\varphi_3$,
and all of $\varphi_i$ are subformulas of $\varphi_4$.
\end{example}

Now for the transformation steps, we assume that $\varphi$ has $n$ subformulas:
\begin{enumerate}
\item For every subformula $\varphi_i$ of $\varphi$, create a fresh variable $t_i$. \emph{These variables are analogous to the temporary variables \lstinline{t}$_i$ we introduced to our Python program above.}
\item Next,
starting with those most-deeply nested subformula,
for every subformula $\varphi_i$, 
create the following formula:
Let $\varphi_i$ be of the form $\ell_i \circ \ell_i'$, where $\circ$ is $\land$ or $\lor$ and $\ell_i,\ell_i'$ are literals.
Note that one or both of $\ell_i$ and $\ell_i'$ may be the new variable $t_j$ denoting a subformula $\varphi_j$ of $\varphi_i$.
Create the formula $$\varphi_i' \triangleq t_i \iff (\ell_i \circ \ell_i')$$
\emph{These formulas are analogous to the assignments to the temporary variables
in our Python program above, where $\iff$ is the logical analogue of variable assignment (\lstinline{=}).}
\end{enumerate}

\begin{example}\label{ex:tseitin}
Continuing our running example, for subformula $\varphi_1$,
we create the formula 
$$\varphi_1' \triangleq t_1 \iff (p\land q)$$
For subformula $\varphi_2$, we create
$$\varphi_2' \triangleq t_2 \iff (q\land \neg r)$$
For subformula $\varphi_3$, we create
$$\varphi_3' \triangleq t_3 \iff (t_2 \land s)$$
(Notice that $q \land \neg r$ is replaced by the variable $t_2$.)
Finally, for subformula $\varphi_4$, we create
$$\varphi_4' \triangleq t_4 \iff (t_1 \lor t_3)$$
\end{example}

Notice that each $\varphi'_i$ can be written in \cnf.
This is because, following the definition of $\iff$ and DeMorgan's laws, we have:
$$\ell_1 \iff (\ell_2 \lor \ell_3) \equiv (\neg \ell_1 \lor \ell_2 \lor \ell_3) \land (\ell_1 \lor \neg \ell_2) \land (\ell_1  \lor \neg \ell_3)$$
and 
$$\ell_1 \iff (\ell_2 \land \ell_3) \equiv (\neg \ell_1 \lor \ell_2) \land (\neg \ell_1 \lor \ell_3) \land (\ell_1 \lor \neg \ell_2 \lor \neg \ell_3)$$

Finally, we construct the following \cnf formula:
$$\varphi' \triangleq t_n \land \bigwedge_i  \varphi'_i$$
By construction, in any model of $\varphi'$, each $t_i$ is assigned $\true$
if and only if the subformula $\varphi_i$ evaluates to $\true$.
Therefore, the  constraint $t_n$ in $F$' says that $\varphi$ must be true.
\emph{You can think of $t_n$ as the return statement in our transformed Python program.}

\begin{example}
    Continuing our running example,
    we finally construct the \cnf formula
    $$\varphi' \triangleq t_4 \land \varphi_1' \land \varphi_2' \land \varphi_3' \land \varphi_4'$$
    Since all of the formulas $\varphi'_i$ can be written in \cnf, as described above,  
    $\varphi'$ is in \cnf.
\end{example}

At this point, we're done. 
Given a formula in some theory and a theory solver,
we first rewrite the formula in \cnf, using Tseitin's tranformation,
and invoke \dpllt.

\section*{Looking Ahead}
I gave a trimmed down version of \dpllt.
A key idea in modern \sat and \smt solvers is \emph{conflict-driven clause learning}, a graph data structure that helps us cut the search space by identifying sets of interpretations that do not make satisfying assignments.
I encourage the interested reader to consult \citet{biere2009handbook}
for a detailed exposition of clause learning and other ideas.

I also encourage you to play with popular \sat and \smt solvers.
For example, MiniSAT~\citep{sorensson2005minisat}, as the name suggests, has a small and readable codebase.
For \smt solvers, I recommend checking out Z3~\citep{de2008z3}
and CVC4~\citep{barrett2011cvc4}.
One of the interesting ideas underlying \smt solvers is 
\emph{theory combination}~\citep{nelson1979simplification}, where you can solve formulas combining different theories.
This is useful when doing general program verification, where programs may manipulate strings, arrays, integers, etc.
In the future, I strongly suspect that theory combination will be needed for neural-network verification, because we will start looking at neural networks as components of bigger pieces of software.











     \chapter{Neural Theory Solvers}\label{ch:specialized}

In the previous chapter, we discussed \dpllt
for solving formulas in \fol.
In this chapter, we study the \emph{Simplex algorithm} for solving conjunctions of literals in linear real arithmetic.
Then, we extend the solver to natively handle rectified linear units (ReLUs),
which would normally be encoded as disjunctions and thus dealt with using the \sat solver.

\section{Theory Solving and Normal Forms}

\subsection*{The Problem}
The theory solver for \lra receives a formula $\varphi$
as a conjunction of linear inequalities:
\[
\bigwedge_{i=1}^n \left(\sum_{j=1}^m c_{ij} \cdot x_j \geq b_i \right)
\]
where $c_{ij}, b_i \in \R$.
The goal is to check satisfiability of $\varphi$,
and, if satisfiable, discover a model $I \models \varphi$.

Notice that our formulas do not have strict inequalities ($>$).
The approach we present here can be easily generalized to handle strict inequalities, but for simplicity, we stick with inequalities.\footnote{
  See \cite{dutertre2006integrating} for how to handle strict inequalities. 
    In most instances of verifying properties of neural networks,
    we do not need strict inequalities to encode network semantics
    or properties.
} 

\subsection*{Simplex Form}
The Simplex algorithm, discussed in the next section,
expects formulas to be in a certain form (just like how \dpll expected propositional formulas to be in \cnf).
Specifically, Simplex expects formulas to be 
conjunctions of \emph{equalities} of the form
\[
\sum_{i} c_{i} \cdot x_i = 0
\]
and \emph{bounds} of the form
\[
\lb_i \leq x_i  \leq \ub_i    
\]
where $u_i, l_i \in \R \cup \{\infty, -\infty\}$.
We use $\infty$ (resp. $-\infty$) to indicate that a variable has no upper (resp. lower) bound.

Therefore, given a formula $\varphi$, we need to translate it into an equivalent formula of the above form, which we will call the \emph{Simplex form} (also known as \emph{slack} form).
It turns out that translating a formula into Simplex form is pretty simple.
Suppose that
\[
\varphi \triangleq \bigwedge_{i=1}^n \left(\sum_{j=1}^m c_{ij} \cdot x_j \geq b_i \right)
\]
Then, we take every inequality and translate it into two conjuncts, an equality and a bound. From the $i$th inequality,
$$\sum_{j=1}^m c_{ij} \cdot x_j \geq b_i$$
we generate the equality
$$s_i = \sum_{j=1}^m c_{ij} \cdot x_j$$
and the bound
$$s_i \geq b_i$$
where $s_i$ is a new variable, called a \emph{slack variable}.
Slack variables are analogous to the temporary variables
introduced by Tseitin's transformation (\cref{ch:dp}).

\begin{example}
    Consider the formula $\varphi$, which we will use as our running example:
    \begin{align*}
      x + y  &\geq 0\\
      - 2x + y  &\geq 2\\
      - 10x + y  &\geq -5
  \end{align*}
    For clarity, we will drop the conjunction operator and simply list the inequalities.
    We convert $\varphi$ into a formula $\varphi_s$ in Simplex form:
    \begin{align*}
      s_1 &= x + y  \\
      s_2 &= -2x + y \\
      s_3 &= -10x + y\\
        s_1 &\geq 0\\
        s_2 &\geq 2\\
        s_3 &\geq -5
    \end{align*}
\end{example}

This transformation is a simple rewriting of the original formula
that maintains satisfiability.
    Let $\varphi_s$ be the Simplex form of some formula $\varphi$.
    Then, we have the following guarantees (again, the analogue of Tseitin's transformation for non-\cnf formulas):
    \begin{enumerate}
    \item Any model of $\varphi_s$ is a model of $\varphi$,
    disregarding assignments to slack variables.
    \item If $\varphi_s$ is \unsat, then $\varphi$ is \unsat.
    \end{enumerate}


\section{The Simplex Algorithm}
We're now ready to present the Simplex algorithm.
This is a very old idea due to George Dantizg, who developed it in 1947 (see Dantzig's recollection of the origins of Simplex~\citep{dantzig1990origins}).
The goal of the algorithm is to find a satisfying assignment that maximizes some objective function.
Our interest in verification is typically to find \emph{any} satisfying assignment, and so the algorithm we will present is a subset of Simplex.

\subsection*{Intuition}
One can think of the Simplex algorithm as a procedure that simultaneously looks for a model and a proof of unsatisfiability.
It starts with some interpretation
and continues to update it in every iteration, until it finds a model 
or discovers a proof of unsatisfiability.
We start from the interpretation $I$ that sets
all variables to 0.
This assignment satisfies all the equalities,
but may not satisfy the bounds.
In every iteration of Simplex, we pick a bound that is not satisfied
and we modify $I$ to satisfy it, or we discover that the formula is unsatisfiable.
Let's see this process pictorially on a satisfiable example before we dig into the math.
\begin{figure}[htpb]
    \centering
    \begin{tikzpicture}
      \begin{axis}[only axis on top,
        axis line style= thick,
        axis x line=middle,
        axis y line=middle,
         ymin=-3,ymax=5,xmin=-4,xmax=4,
         xtick={-4,-3,-2,-1,0,1,2,3,4},
         ytick={-3,-2,-1,0,1,2,3,4,5},
         xticklabels={,-3,-2,-1,1,0,2,3},
         yticklabels={,-2,-1,0,2,3,4},
      ]        
        \draw[draw=none, fill=Maroon!30]
        (axis cs: 1,5)--(axis cs:-5,5)--(axis cs: -0.66,0.66)--(axis cs: 7/8,14/8 + 2)--cycle;
        \addplot[very thick,domain=-4:4] plot (\x,-\x);
        \addplot[very thick,domain=-5:5] plot (\x,2*\x+2);
        \addplot[very thick,domain=-5:5] plot (\x,10*\x - 5);
        \node[font=\small] at (axis cs: 3.8,-0.5) {$x$};
        \node[font=\small, right] at (axis cs: 0,4.75) {$y$};
        \node[font=\small] at (axis cs: -3,4.1) {$x + y \geq 0$};
        \node[font=\small,right] at (axis cs: 0.5,1) {$-10x + y \geq -5$};
        \node[font=\small,right] at (axis cs: 1.3,4.4) {$-2x + y \geq 2$};


    \addplot[Maroon,mark=*,font=\small] coordinates {(0,0)}
    node [pos=0.1, above right] {$I_0$};
    \addplot[Maroon,mark=*,font=\small] coordinates {(-1,0)}
    node [pos=0.1, above left] {$I_1$};
    \addplot[Maroon,mark=*,font=\small] coordinates {(-2/3,2/3)}
    node [pos=.1, right ] {$I_2$};
      \end{axis}
    \end{tikzpicture}
    \caption{Simplex example}\label{fig:Simplex}
  \end{figure}
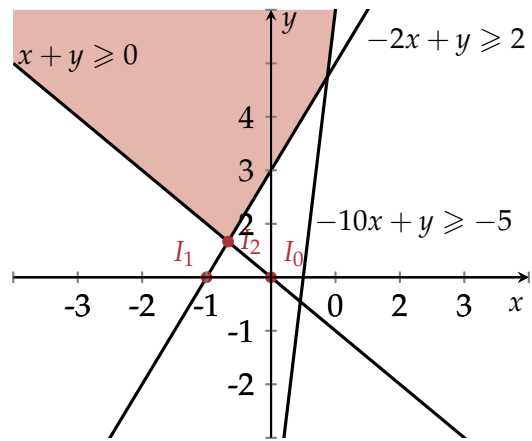

\begin{example}
  Recall the formula $\varphi$ from our running example, illustrated in \cref{fig:Simplex},
  where the satisfying assignments are the shaded region.
  Each inequality defines \emph{halfspace}, i.e., 
  splits $\R^2$ in half.
  Taking all of the inequalities, we get the shaded region---the intersection of the all halfspaces.
    \begin{align*}
        x + y  &\geq 0\\
        - 2x + y  &\geq 2\\
        - 10x + y  &\geq -5
    \end{align*}
    Simplex begins with the initial interpretation
    \[
    I_0 = \{x \mapsto 0, y \mapsto 0\}  
    \]
    shown in \cref{fig:Simplex},
    which is not a model of the formula.
    
    Simplex notices that 
    $$I_0 \not\models -2x + y \geq 2$$
    and so it decreases the interpretation of $x$ to $-1$,
    resulting in $I_1$.
    Then, Simplex notices that $$I_1 \not\models x+y \geq 0$$
    and so it increases the interpretation of $y$ from $0$ to $2/3$,
    resulting in the satisfying assignment $I_2$.
    (Notice that $x$ also changes in $I_2$; we will see why shortly.)
    In a sense, Simplex plays Whac-A-Mole,
    trying to satisfy one inequality only to break another,
    until it arrives at an assignment that satisfies all inequalities.
    Luckily, the algorithm actually terminates.
\end{example}

\subsection*{Basic and Non-basic Variables}
Recall that Simplex expects an input formula to be in Simplex form.
The set of variables in the formula are broken into two subsets:
\begin{description}
\item[Basic variables] are those that appear on the left hand side of an equality; initially, basic variables are the slack variables.
\item[Non-basic variables] are all other variables. 
\end{description}
As Simplex progresses, it will rewrite the formula, thus some basic variables will become non-basic and vice versa.

\begin{example}
  In our running example, initially the set of basic variables is 
  $\{s_1,s_2,s_3\}$ and non-basic variables is $\{x,y\}$.
\end{example}

To ensure termination of Simplex,
we will fix a total ordering on the set of all (basic and non-basic) variables.
So, when we say \emph{``the first variable that...''}, we're referring to the first variable per our ordering.
To easily refer to variables, we will assume they are of the form
$x_1, \ldots, x_n$.
Given a basic variable $x_i$ and a non-basic variable $x_j$,
we will use $c_{ij}$ to denote the coefficient of $x_j$ in the equality
\[
x_i = \ldots + c_{ij} \cdot x_j + \ldots  
\]
For a variable $x_i$, we will use $\lb_i$ and $\ub_i$
to denote its lower bound and upper bound, respectively.
If a variable does not have an upper bound (resp. lower bound), its upper bound is  $\infty$ (resp. $-\infty$).
Note that non-slack variables have no bounds.

\subsection*{Simplex in Detail}
We're now equipped to present the Simplex algorithm,
shown in \cref{alg:Simplex}.
The algorithm maintains the following two invariants:
\begin{enumerate}
\item The interpretation $I$ always satisfies the equalities, so only the bounds may be violated.
This is initially true, as $I$ assigns all variables to $0$.
\item The bounds of non-basic variables are all satisfied.
This is initially true, as non-basic variables have no bounds.
\end{enumerate}

In every iteration of the while loop,
Simplex looks for a basic variable whose bounds are not satisfied by the current interpretation,
and attempts to fix the interpretation.
There are two symmetric cases, encoded as two branches of the \emph{if} statement,
$x_i < \lb_i$ or $x_i > \ub_i$.

Let's consider the first case, $x_i < \lb_i$.
Since $x_i$ is less than $\lb_i$, we need to increase its assignment in $I$.
We do this indirectly by modifying the assignment of a non-basic variable $x_j$.
But which $x_j$ should we pick?
In principle, we can pick any $x_j$ such that the coefficient $c_{ij} \neq 0$,
and adjust the interpretation of $x_j$ accordingly. 
If you look at the algorithm, there are a few extra conditions.
If we cannot find an $x_j$ that satisfies these conditions, then 
the problem is \unsat.
We will discuss the unsatisfiability conditions shortly. 
For now, assume we have found an $x_j$.
We can increase its current interpretation by $\frac{\lb_i - I(x_i)}{c_{ij}}$;
this makes the interpretation of $x_i$ increase by $\lb_i - I(x_i)$, thus 
barely satisfying the lower bound, i.e., $I(x_i) = \lb_i$.
Note that the interpretations of basic variables are assumed to change automatically when we change the interpretation of non-basic variables.
This maintains the first invariant of the algorithm.\footnote{
  Basic variables are sometimes called \emph{dependent} variables
  and non-basic variables \emph{independent} variables,
  indicating that the assignments of basic variables depend
  on those of non-basic variables.
}

After we have updated the interpretation of $x_j$, there is a chance that we have violated one of the bounds of $x_j$. Therefore, we rewrite the formulas such that $x_j$ becomes a 
basic variable and $x_i$ a non-basic variable.
This is known as the \emph{pivot} operation,
and it is mechanically done as follows:
Take the following equality, where $N$ is the set of indices of non-basic variables:
$$x_i = \sum_{k \in N} c_{ik} x_k$$
and rewrite it by moving $x_j$ to the left-hand side:
$$x_j = \underbrace{-\frac{x_i}{c_{ij}} + \sum_{k \in N\setminus\{j\}} \frac{c_{ik}}{c_{ij}} x_k}_{\text{replace } x_j \text{ with this}}$$
Now, replace $x_j$ in all other equalities with the expression above.
This operation results in a set of equalities where $x_j$ only appears once, 
on the left-hand side.
And so, after pivoting, $x_j$ becomes a basic variable and $x_i$ a non-basic one.

\SetEndCharOfAlgoLine{}
\begin{algorithm}[t]
  \KwData{A formula $\varphi$ in Simplex form}
  \KwResult{$I \models \varphi$ or \unsat}
  ~\\
  Let $I$ be the interpretation that sets all variables $\fv(\varphi)$ to $0$\\
  \While{true}{
    \lIf{$I \models \varphi$}{\Return $I$}
    Let $x_i$ be the first basic variable s.t. $I(x_i) < \lb_i$ or $I(x_i) > \ub_i$ \\
    
    \eIf{$I(x_i) < l_i$}{
      Let $x_j$ be the first non-basic variable s.t. 
      $$(I(x_j) < \ub_j \text{ and } c_{ij} > 0) \text{ or } (I(x_j) > \lb_j \text{ and } c_{ij} < 0)$$ \\
      \lIf{If no such $x_j$ exists}{\Return \unsat}
      $I(x_j) \gets I(x_j) + \frac{l_i - I(x_i)}{c_{ij}}$
    }
    {
      Let $x_j$ be the first non-basic variable s.t. 
      $$(I(x_j) > \lb_j \text{ and } c_{ij} > 0) \text{ or } (I(x_j) < \ub_j \text{ and } c_{ij} < 0)$$ \\
      \lIf{If no such $x_j$ exists}{\Return \unsat}
      $I(x_j) \gets I(x_j) + \frac{u_i - I(x_i)}{c_{ij}}$
    }
    Pivot $x_i$ and $x_j$
  }
  \caption{Simplex}\label{alg:Simplex}
 \end{algorithm}

\begin{example}
Let's now work through our running example in detail.
Recall that our formula is:
\begin{align*}
  s_1 &= x + y  \\
  s_2 &= -2x + y \\
  s_3 &= -10x + y\\
    s_1 &\geq 0\\
    s_2 &\geq 2\\
    s_3 &\geq -5
\end{align*}
Say the variables are ordered as follows:
$$x,y,s_1,s_2,s_3$$
Initially, the bounds of $s_1$ and $s_3$ are satisfied,
but $s_2$ is violated, because $s_2 \geq 2$
but $I_0(s_2) = 0$, as all variables are assigned 0 initially.

\begin{description}
\item[First iteration]  
In the first iteration, we pick the variable $x$
to fix the bounds of $s_2$, as it is the first one in our ordering.
Note that $x$ is unbounded (i.e., its bounds are $-\infty$ and $\infty$),
so it easily satisfies the conditions.
To increase the interpretation of $s_2$ to 2, and satisfy its lower bound,
we can decrease $I_0(x)$ to $-1$,
resulting in the following satisfying assignment:
\[
I_1 = \{x \mapsto -1,\ y \mapsto 0,\ s_1 \mapsto -1,\ s_2 \mapsto 2,\ s_3 \mapsto 10\}  
\]
(Revisit \cref{fig:Simplex} for an illustration.)
We now pivot $s_2$ and $x$, producing the following set of equalities (the bounds always remain the same):
\begin{align*}
x &= 0.5 y - 0.5 s_2\\
s_1 &= 1.5y - 0.5s_2\\
s_3 & = -4y + 5s_2
\end{align*}

\item[Second iteration]
The only basic variable not satisfying its bounds is now $s_1$,
since $I_1(s_1) = -1 < 0$.
The first non-basic variable that we can tweak is $y$.
We can increase the value of $I(y)$ by 1/1.5 = 2/3, 
resulting in the following interpretation:
\[
I_2 = \{x \mapsto -2/3,\ y \mapsto 2/3,\ s_1 \mapsto 0,\ s_2 \mapsto 2,\ s_3 \mapsto 7/{3}\}  
\]
At this point, we pivot $y$ with $s_1$.

\item[Third iteration]
Simplex terminates since $I_2 \models \varphi$.

\end{description}

\end{example}

\subsection*{Why is Simplex Correct?}
First, you may wonder, why does Simplex terminate?
The answer is due to the fact that we order variables
and always look for the \emph{first} variable violating bounds.
This is known as \emph{Bland's rule} \citep{bland1977new}.
Bland's rule ensures that we never revisit the same set of basic and non-basic variables.

Second, you may wonder, is Simplex actually correct?
If Simplex returns an interpretation $I$, it is easy to see that $I \models \varphi$, since Simplex checks that condition before it terminates.
But what about the case when it says \unsat?
To illustrate correctness in this setting, we will look at an example.

\begin{example}
  Consider the following formula in Simplex form:
  \begin{align*}
    s_1 &= x + y\\
    s_2 &= -x -2y\\
    s_3 &= -x + y\\
    s_1 &\geq 0\\
    s_2 &\geq 2\\
    s_3 &\geq 1
  \end{align*}
  This formula is \unsat---use your favorite \smt solver to check this.
  Imagine an execution of  Simplex that performs the following two pivot operations:
  (1) $s_1$ with $x$ and (2) $s_2$ with $y$.
  
  The first pivot results in the following formula:
  \begin{align*}
    x &= s_1 - y\\
    s_2 &= -s_1 -y\\
    s_3 &= -s_1 + 2y
  \end{align*}
  The second pivot results in the following formula:
  \begin{align*}
    x &= 2s_1 + s_2\\
    y &= -s_2 - s_1\\
    s_3 &= - 3s_1 -2s_s 
  \end{align*}
  The algorithm maintains the invariant that 
  all non-basic variables satisfy their bounds.
  So we have $s_1 \geq 0$ and $s_2 \geq 2$.
  Say $s_3$ violates its bound,
  i.e., $$-3s_1 -2s_2 < 1$$
  The only way to fix this is by decreasing the interpretations of $s_1$ and $s_2$.
  But even if we assign $s_1$ and $s_2$ the values $0$ and $2$ (their lower bounds), respectively,
  we cannot make $s_3 \geq 1$.
  \emph{Contradiction!}
  So Simplex figures out the  formula is \unsat.
  The conditions for choosing variable $x_j$ in \cref{alg:Simplex} encode this argument. 
\end{example}

As the above example illustrates, we can think of Simplex as constructing
a proof by contradiction to prove that a set of linear inequalities is unsatisfiable.

\section{The Reluplex Algorithm}
Using the Simplex algorithm as the theory solver within \dpllt
allows us to solve formulas in \lra. 
So, at this point in our development, we know how to algorithmically
reason about neural networks with piecewise-linear activations, like ReLUs.
Unfortunately, this approach has been shown to not scale to large networks.
One of the reasons is that ReLUs are encoded as disjunctions,
as we saw in \cref{ch:encodings}.
This means that the \sat-solving part of \dpllt will handle the disjunctions, and may end up considering every possible case of the disjunction---ReLU being active (output = input) or inactive (output = 0)---leading to many calls to Simplex,  exponential in the number of ReLUs.

To fix those issues, the work of \citet{katz2017reluplex} developed an extension of Simplex, called \emph{Reluplex}, that natively handles ReLU constraints in addition to linear inequalities.
The key idea is to try to \emph{delay} case splitting on ReLUs.
In the worst case, Reluplex may end up with an exponential explosion, just like \dpllt with Simplex,
but empirically it has been shown to be a promising approach for scaling \smt solving to larger neural networks.
In what follows, we present the Reluplex algorithm.

\subsection*{Reluplex Form}
Just like with Simplex, Reluplex expects formulas to be in a certain form.
We will call this form \emph{Reluplex form},
where formulas contain (1) equalities (same as Simplex), (2) bounds (same as Simplex), and (3) \emph{ReLU constraints} of the form
\[
x_i = \relu(x_j)  
\] 
Given a conjunction of inequalities and ReLU constraints,
we can translate them into Reluplex form by translating the 
inequalities into Simplex form.
Additionally, for each ReLU constraint $x_i = \relu(x_j)$,
we can add the bound $x_i \geq 0$, which is implied by the definition  of a ReLU.

\begin{example}
  Consider the following formula:
  \begin{align*}
    x& + y \geq 2\\
    y& = \relu(x)
  \end{align*}
  We translate it into the following Reluplex form:
  \begin{align*}
    s_1 &= x + y\\
    y& = \relu(x)\\
    s_1& \geq 2\\
    y & \geq 0
  \end{align*}
\end{example}

\SetEndCharOfAlgoLine{}
\begin{algorithm}[h!]
  \KwData{A formula $\varphi$ in Reluplex form}
  \KwResult{$I \models \varphi$ or \unsat}
  ~\\
  Let $I$ be the interpretation that sets all variables $\fv(\varphi)$ to $0$\\
  Let $\varphi'$ be the non-ReLU constraints of $\varphi$\\
  \While{true}{
    \acomment{Calling Simplex (note that we supply Simplex with a reference to the initial interpretation and it can modify it)}\\
    $r \gets \text{Simplex}(\varphi', I)$\\
    \textbf{If} {$r$ is \unsat} \textbf{then} \Return \unsat\\
    {
      $r$ is an interpretation $I$\\
      \lIf{$I \models \varphi$}{
        \Return $I$
      }
    }
    ~\\
    \acomment{Handle violated ReLU constraint}\\
    Let ReLU constraint $x_i = \relu(x_j)$ be s.t. $I(x_i) \neq \relu(I(x_j))$\\
    \If{$x_i$ is basic}{pivot $x_i$ with non-basic variable $x_k$, where $k\neq j $ and $c_{ik}\neq 0$}
    \If{$x_j$ is basic}{pivot $x_j$ with non-basic variable $x_k$, where $k\neq i $ and $c_{jk}\neq 0$}
    Perform one of the following operations:
      $$I(x_i) \gets \relu(I(x_j)) \text{~~~\textbf{or}~~~} I(x_j) \gets I(x_i)$$
    
    \acomment{Case splitting (ensures termination)} \\
    \If{$u_j > 0$, $l_j < 0$, and $x_i = \relu(x_j)$ considered more than $\tau$ times}
    {
      $r_1 \gets \text{Reluplex}(\varphi \land x_j \geq 0 \land x_i=x_j)$\\
      $r_2 \gets \text{Reluplex}(\varphi \land x_j \leq 0 \land x_i = 0)$\\
      \lIf{$r_1 = r_2 = $ \unsat}{\Return \unsat}
      \lIf{$r_1 \neq $ \unsat}{\Return $r_1$}
      \Return $r_2$ 
    }
  }
  \caption{Reluplex}\label{alg:Reluplex}
 \end{algorithm}

\subsection*{Reluplex in Detail}
We now present the Reluplex algorithm.
The original presentation by \citet{katz2017reluplex}
is a set of rules that can be applied non-deterministically
to arrive at an answer.
Here, we present a specific schedule of the Reluplex algorithm.
 
The key idea of Reluplex is to call Simplex on equalities and bounds, and then try to massage the interpretation returned by Simplex to satisfy all ReLU constraints.
Reluplex is shown in \cref{alg:Reluplex}.

Initially, Simplex is invoked on the formula $\varphi'$,
which is the original formula $\varphi$ but 
without the ReLU constraints.
If Simplex returns \unsat, then we know that $\varphi$ is \unsat---%
this is because $\varphi \Rightarrow \varphi'$ is valid.
Otherwise, if Simplex returns a model $I \models \varphi'$,
it may not be the case that $I \models \varphi$,
since $\varphi'$ is a weaker (less constrained) formula.

If $I \not\models \varphi$, then we know that one of the ReLU 
constraints is not satisfied.
We pick one of the violated ReLU constraints $x_i = \relu(x_j)$ 
and modify $I$ to make sure it is not violated.
Note that if any of  $x_i$ and $x_j$ is a  basic variable,
we pivot it with a non-basic variable.
This is because  we want to modify the interpretation of one of $x_i$ or $x_j$, which may affect
the interpretation of the other variable if it is a basic variable and $c_{ij} \neq 0$.\footnote{
  These conditions are not explicit in \cite{katz2017reluplex},
  but their absence may lead to wasted iterations (or Update rules in \cite{katz2017reluplex}) that do not fix violations of ReLU constraints.
}
Finally, we modify the interpretation of $x_i$ or $x_j$,
ensuring that $I \models x_i = \relu(x_j)$.
Note that the choice of $x_i$ or $x_j$ is up to the implementation.

The problem is that fixing a ReLU constraint may end up violating a bound, and so Simplex need be invoked again.
We assume that the interpretation $I$ in Reluplex is the same one that is modified by invocations of Simplex.

\subsection*{Case Splitting}
If we simply apply Reluplex without the last piece of the algorithm---case splitting---it may not terminate.
Specifically, it may get into a loop where Simplex satisfies all bounds but violates a ReLU, and then satisfying the ReLU causes a bound to be violated, and on and on.

The last piece of Reluplex checks if we are getting into an infinite loop,
by ensuring that we do not attempt to fix a ReLU constraint more than $\tau$ times, some fixed threshold.
If this threshold is exceeded, then the ReLU constraint $x_i = \relu(x_j)$ is split into its two cases:
$$\varphi_1 \triangleq x_j \geq 0 \land x_i = x_j$$
and 
$$\varphi_2 \triangleq x_j \leq 0 \land x_i = 0$$
Reluplex is invoked recursively on two instances of the problem,
$\varphi\land\varphi_1$ and $\varphi\land\varphi_2$.
If both instances are \unsat, then the formula  $\varphi$ is \unsat.
If any of the instances is SAT, then $\varphi$ is SAT.
This is due to the fact that 
\[
  \varphi \equiv (\varphi \land \varphi_1) \lor (\varphi \land \varphi_2)
\]

\section*{Looking Ahead}
We're done with constraint-based verification.
In the next part of the book, we will look at different approaches that are more efficient at the expense of failing to provide proofs in some cases.

There are a number of interesting problems that we didn't cover.
A critical one is soundness with respect to machine arithmetic.
Our discussion has assumed real-valued variables, but, of course,
that's not the case in the real world---we use machine arithmetic.
A recent paper has shown that verified neural networks in \lra may not really be robust when one considers the bit-level behavior~\citep{jia2020exploiting}.

Another issue is scalability of the analysis. 
Using arbitrary-precision rational numbers can be very expensive,
as the size of the numerators and denominators can blow up due to pivot operations.
Reluplex~\citep{katz2017reluplex} ends up using floating-point approximations, and carefully ensures
the results are sound by keeping track of round-off errors.

At the time of writing, constraint-based verification approaches have only managed to scale to neural networks with around a hundred thousand ReLUs~\citep{tjeng2018evaluating},
which is small compared to state-of-the-art neural networks.
This is still a triumph, as the verification problem is NP-hard~\citep{katz2017reluplex}.
It is, however, unclear how much further we can push this technology.
Scaling constraint-based verification further requires
progress along two fronts: (1) developing and training neural networks  
that are friendlier to verification (less ReLUs is better),
and (2) advancing the algorithms underlying \smt solvers and \milp solvers.


\part{Abstraction-Based Verification}

    \chapter{Neural Interval Abstraction}\label{ch:absint}  

In the previous part of the book, we described how to precisely capture the semantics of a neural network by encoding it, along with a correctness property, as a formula in first-order logic. 
Typically, this means that we're solving an NP-complete problem, like satisfiability modulo linear real arithmetic (equivalently, mixed integer linear programming). 
While we have fantastic algorithms and tools that surprisingly work well for such hard problems, scaling to large neural networks remains an issue.

In this part of the book, we will look at approximate techniques for neural-network verification. By approximate, we mean that they overapproximate---or abstract---the semantics of a neural-network, and therefore can produce proofs of correctness, but when they fail, we do not know whether a correctness property holds or not.
The approach we use is based on \emph{abstract interpretation}~\citep{cousot1977abstract}, a well-studied framework for defining program analyses.
Abstract interpretation is a very rich theory, and the math can easily make you want to quit computer science and live a monastic life in the woods, away from anything that can be considered technology. But fear not, it is a very simple idea, and we will take a pragmatic approach here in defining it and using it for neural-network verification.

\section{Set Semantics and Verification}
Let's focus on the following correctness property, defining robustness of a neural network $f: \R^n \to \R^m$ on an input grayscale image $\vec{c}$
whose classification label is 1.
\[
\begin{array}{c}
    \pre{|\vec{x} - \vec{c}| \leq \vec{0.1}}\\
    \vec{r} \gets f(\vec{x})\\
    \post{\class(\vec{r}) = 1}
\end{array}
\]
Concretely, this property makes the following statement: 
Pick any image $\vec{x}$ that is like $\vec{c}$
but is slightly brighter or darker by at most $0.1$ per pixel,
assuming each pixel is some real number encoding its brightness.
Now, execute the network on $\vec{x}$. The network must predict that $\vec{x}$ is of class $1$.

The issue in checking such statement is that there are infinitely many possible images $\vec{x}$.
Even if there are finitely many images---because, at the end of the day, we're using bits---the number is still enormous, and we cannot conceivably run all those images through the network and ensure that each and every one of them is assigned class $1$.
But let's just, for the sake of argument, imagine that we can lift the function $f$ to work over \emph{sets of images}. That is, we will define a version of $f$ of the form:
$$\setf{f}: \ps{\R^n} \to \ps{\R^m}$$
where $\ps{S}$ is the powerset of set $S$.
Specifically, 
$$\setf{f}(\xset) = \{\vec{y} \mid \vec{x} \in \xset,\; \vec{y} = f(\vec{x})\}$$

Armed with $\setf{f}$, we can run it with the following input set:
$$\xset = \{\vec{x} \mid |\vec{x} - \vec{c}| \leq \vec{0.1} \}$$
which is the set of all images $\vec{x}$ defined above in the precondition of our correctness property.
By computing $\setf{f}(\xset)$, we get the predictions of the neural network $f$ for all images
in $\xset$.
To verify our property, we simply check that
$$\setf{f}(\xset) \subseteq \{\vec{y} \mid \class(\vec{y}) = 1\}$$
In other words, all runs of $f$ on every image $\vec{x}\in \xset$ 
result in the network predicting class $1$.

The above discussion may sound like crazy talk:
we cannot simply take a neural network $f$ and generate
a version $\setf{f}$ that takes an infinite set of images.
In this chapter, we will see that we actually \emph{can},
but we will often have to lose precision:
we will define an abstract version of our theoretical $\setf{f}$
that may return  more answers.
The trick is to define infinite sets of inputs using data structures
that we can manipulate, called \emph{abstract domains}.

In this chapter, we will meet the \emph{interval}
abstract domain.
We will focus our attention on the problem of executing the neural network on an infinite set.
Later, in \cref{ch:absver}, we come back to the verification problem. 

\section{The Interval Domain}

Let's begin by considering a very simple function
$$f(x) = x + 1$$
I would like to evaluate this function 
on a set of inputs $X$; that is, I would like
to somehow evaluate $$f^s(X) = \{x + 1 \mid x \in X\}$$
We call $f^s$ the \emph{concrete transformer} of $f$.

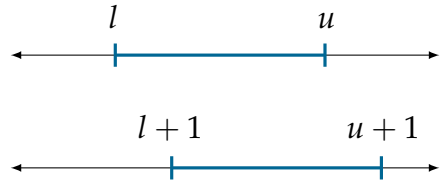
\begin{figure}
\centering
\begin{tikzpicture}[decoration={brace,mirror,amplitude=7}]

    \draw[latex-latex] (-2.2,0) -- (3.5,0) ; 
    \draw[very thick, |-|,MidnightBlue]  (-0.83,0) node[above=2mm,black] {$l$} --   (2,0) node[above=2mm,black] {$u$};

    \draw[latex-latex] (-2.2,-1.5) -- (3.5,-1.5) ; 
    \draw[very thick, |-|,MidnightBlue]  (-0.83+.75,-1.5) node[above=2mm,black] {$l+1$} --   (2+.75,-1.5) node[above=2mm,black] {$u+1$};





\end{tikzpicture}
\caption{Illustration of an abstract transformer of $f(x) = x + 1$.}\label{fig:interval}
\end{figure}

Abstract interpretation simplifies 
this problem by only considering sets $X$ that have a nice
form. Specifically, the \emph{interval abstract domain}
considers an interval of real numbers
written as $[l,u]$, where $l,u \in \R$ and $l\leq u$.
An interval $[l,u]$ denotes the potentially infinite set $$\{x \mid l \leq x \leq u\}$$
So we can now define a version of our function $f^s$
that operates over an interval, as follows:
$$f^a([l,u]) = [l+1, u+1]$$
We call $f^a$ an \emph{abstract transformer} of $f$.
In other words, $f^a$ takes a set of real numbers
and returns a set of real numbers, but the sets are restricted to those that can be defined as intervals.
Observe how we can mechanically evaluate this abstract transformer on an arbitrary interval $[l,u]$:
add 1 to $l$ and add 1 to $u$, arriving at the interval $[l+1,u+1]$.
Geometrically, if we have an interval on the number line from $l$ to $u$, and we add 1 to every point in this interval,
then the whole interval shifts to the right by 1. This is illustrated in \cref{fig:interval}.
Note that the interval $[l,u]$ is an infinite set (assuming $l< u$), and so $f^a$ adds 1 to an infinite set of real numbers!

\begin{example}
Continuing our example, $$f^a([0,10]) = [1,11]$$
If we pass a singleton interval, e.g., $[1,1]$,
we get $f^a([1,1]) = [2,2]$---exactly the behavior of $f$.
\end{example}

Generally, we will use the notation $([l_1,u_1],\ldots,[l_n,u_n])$
to denote an $n$-dimensional interval,
or a hyperrectangular region in $\R^n$, i.e.,
the set of all $n$-ary vectors 
$$\{\vec{x} \in \R^n \mid l_i \leq x_i \leq u_i\}$$

\subsection*{Soundness}
Whenever we design an abstract transformer $f^a$,
we need to ensure that it is a \emph{sound} approximation of $f^s$.
This means that its output is a superset of
that of the concrete transformer, $f^s$.
The reason is that we will be using $f^a$ for verification,
so to declare that a property holds, we cannot afford
to miss any behavior of the neural network.

Formally, we define soundness as follows:
For any interval $[l,u]$, we have 
$$f^s([l,u]) \subseteq f^a([l,u])$$
Equivalently, 
we can say that for any $x \in [l,u]$,
we have $$f(x) \in f^a([l,u])$$

In practice, we will often find
that $$f^s([l,u]) \subset f^a([l,u])$$
for many functions and intervals of interest.
This is expected, as our goal is to design abstract transformers
that are easy to evaluate, and so we will often \emph{lose precision}, meaning overapproximate the results of $f^s$.
We will see some simple examples shortly.

\subsection*{The Interval Domain is Non-relational}
The interval domain is \emph{non-relational},
meaning that it cannot capture the relations between different dimensions.
We illustrate this fact with an example.

\begin{example}
    Consider the set 
    $$\xset = \{(x,x) \mid 0\leq x\leq 1\}$$
    We cannot represent this set precisely in the interval domain.
    The best we can do is the square between $(0,0)$ and $(1,1)$,
    denoted as the 2-dimensional  interval
    $$([0,1],[0,1])$$
    and illustrated as the gray region below:
    \begin{center}
    \begin{tikzpicture}
        \begin{axis}[
            axis equal image, 
            axis lines=middle,
            xmax=1.5,
            xmin=-.5,
            ymin=-.5,
            ymax=1.5,
            xlabel={},
            ylabel={},
            ytick={0},
            xtick ={0},width=5cm,
        ]
        \draw [fill=black, opacity=0.2] (axis cs:0,0) -- (axis cs:0,1)
        --(axis cs:1,1) -- (axis cs:1,0) -- (axis cs:0,0);
        \addplot [domain=0:1, samples=500,
                  thick, Maroon] {x};
        
        \end{axis}
    \end{tikzpicture}
\end{center}
The set $\xset$ defines points where higher values of
the $x$ coordinate associate with higher values of the $y$ coordinate.
But our abstract domain can only represent rectangles
whose faces are parallel to the axes.
This means that we can't capture
the relation between the two dimensions:
we simply
say that any value of $x$ in $[0,1]$ can 
associate with any value of $y$ in $[0,1]$.
\end{example}

\section{Basic Abstract Transformers}
We now look at examples of abstract transformers
for basic arithmetic operations.

\subsection*{Addition}
Consider the  binary function:
$f(x,y) = x+ y$.
The concrete transformer $f^s: \ps{\R^2} \to \ps{\R}$
is defined as follows:
$$f^s(X) = \{ x+ y \mid (x,y)\in X\}$$
We define $f^a$ as a function that takes
 two intervals, i.e., a rectangle, one representing the range
of values of $x_1$ and the other of $x_2$:
$$f^a([l,u], [l',u']) = [l+l', u+u']$$
The definition looks very much like $f$,
except that we perform addition on the lower bounds and the upper bounds of the two input intervals.

\begin{example}
    Consider
    $$f^a([1,5],[100,200]) = [101,205]$$
    The lower bound, 101, results from adding the lower
    bounds of $x$ and $y$ (1 + 100);
    the upper bound, 205, results from adding the upper bounds
    of $x$ and $y$ (5 + 200).
\end{example}

It is simple to prove soundness of our abstract transformer $f^a$.
Take any $$(x,y) \in ([l,u],[l',u'])$$
By definition, $l \leq x \leq u$ and $l' \leq y \leq u'$.
So we have  $$l + l' \leq x + y \leq u + u'$$
By definition of an interval, we have $$x + y \in [l+l',u+u']$$

\subsection*{Multiplication}
Multiplication is a bit trickier.
The reason is that the signs might flip,
making the lower bound an upper bound.
So we have to be a bit more careful.

Let $f(x,y) = x*y$.
If we only consider positive inputs,
then we can define $f^a$ just like we did for addition:
$$f^a([l,u],[l',u']) = [l*l', u*u']$$
But consider 
$$f^a([-1,1],[-3,-2]) = [3,-2]$$
We're in trouble: $[3,-2]$ is not even an interval
as per our definition---the upper bound is less than the lower bound!

To fix this issue, we need to consider  every possible combination of lower and upper bounds as follows:
$$f^a([l,u],[l',u']) = [\min(B),\max(B)]$$
where 
$$B = \{l*l',\ l*u',\ u*l',\ u*u'\}$$

\begin{example} Consider the following abstract multiplication of two intervals:
        \begin{align*}
        f^a([-1,1],[-3,-2]) & = [\min(B),\max(B)] \\
        & = [-3,3]
        \end{align*}
        where $B = \{3,2,-3,-2\}$.
\end{example}

\section{General Abstract Transformers}
We will now define general abstract transformers
for classes of operations that commonly appear in neural networks.

\subsection*{Affine Functions}
For an affine function $$f(x_1,\ldots,x_n) = \sum_i c_ix_i$$
where $c_i \in \R$,
we can define the abstract transformer as follows:
$$f^a([l_1,u_1],\ldots,[l_n,u_n]) = 
\left[\sum_i l_i',\ \sum_i u_i'\right]$$
where $l_i' = \min(c_i  l_i, c_i u_i)$ and $u_i' = \max(c_il_i,c_iu_i)$.

Notice that the definition looks pretty much like addition: sum up the lower bounds and the upper bounds.
The difference is that we also have to consider the coefficients, $c_i$, which may result in flipping an interval's bounds when $c_i < 0$.

\begin{example}
Consider $f(x,y) = 3x + 2y$.
Then, 
\begin{align*}
f([5,10], [20,30]) & = [3\cdot 5 + 2 \cdot 20,\ 3\cdot 10 + 2 \cdot 30]\\
 & = [55,90]
\end{align*}

\end{example}

\paragraph{Monotonic Functions}
Most activation functions used in neural networks
are monotonically increasing, e.g., ReLU and sigmoid.
It turns out that it's easy to define an abstract
transformer for any monotonically increasing 
function $f: \R \to \R$, as follows:
$$f^a([l,u]) = [f(l),f(u)]$$
Simply, we apply $f$ to the lower and upper bounds.

\begin{figure}[t]
    \centering
\begin{tikzpicture}
    \begin{axis}[
        axis equal image, 
        axis lines=middle,
        xmax=10,
        xmin=-10,
        ymin=-1.5,
        ymax=10,
        xlabel={$x$},
        ylabel={$\relu(x)$},
        ytick={0},
        xtick ={0}
    ]

    \addplot [domain=-10:8, samples=500,
              thick, MidnightBlue] {max(0,x)};
    
    \addplot[Maroon,mark=*] coordinates {(3,0)}
    node [pos=-0.1,  below ] {$l=3$};

    \addplot[Maroon,mark=*] coordinates {(5,0)}
    node [pos=-0.1,  below right] {$u=5$};
    
    \draw [Maroon, fill=Maroon, opacity=0.2] (axis cs:3,0) -- (axis cs:5,0)
    --(axis cs:5,5) -- (axis cs:3,3) -- (axis cs:3,0);
    
    \draw [Maroon, fill=Maroon, opacity=0.2] (axis cs:3,3) -- (axis cs:5,5)
    --(axis cs:0,5) -- (axis cs:0,3) -- (axis cs:3,3);

    \addplot[Maroon,mark=*] coordinates {(0,3)}
    node [pos=0,  left] {$\relu(l)=3$};

    \addplot[Maroon,mark=*] coordinates {(0,5)}
    node [pos=0,  left] {$\relu(u)=5$};

    \end{axis}
\end{tikzpicture}
\caption{ReLU function over an interval of inputs $[l,u]$}\label{fig:reluint}
\end{figure}
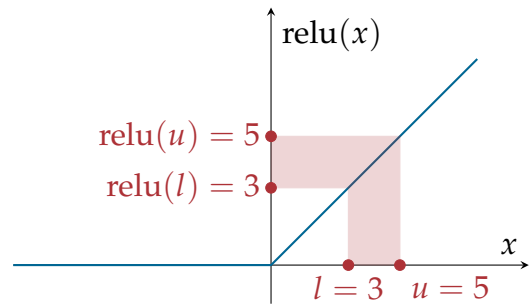

\begin{example}
    \cref{fig:reluint} illustrates
    how to apply ReLU to an interval $[3,5]$.
    The shaded region shows that any value $y$ in the interval
    $[3,5]$ results in a value $$ \relu(3) \leq \relu(y) \leq \relu(5)$$
    that is, a value in the interval $[\relu(3),\relu(5)]$.
\end{example}

\subsection*{Composing Abstract Transformers}
Say we have a function composition $f \circ g$---this notation means $(f \circ g)(x) = f(g(x))$.
We don't have to define an abstract transformer
for the composition:
we can simply compose the two abstract transformers
of $f$ and $g$, as $f^a \circ g^a$, and this will be a sound
abstract transformer of $f \circ g$.

Composition is very important, 
as neural networks are a composition of many operations.
\begin{example}
    Let 
    \begin{align*}
        g(x) & = 3x\\
        f(x) & = \relu(x)\\
        h(x) & = f(g(x))
    \end{align*}

    The function $h$ represents a very simple neural network,
    one that applies an affine function followed by a ReLU
    on an input in $\R$.

    We define $$h^a([l,u]) = f^a(g^a([l,u]))$$
    where $f^a$ and $g^a$ are as defined earlier for monotonic functions and affine functions, respectively.
    For example, on the input interval $[2,3]$,
    we have 
    \begin{align*}
        h^a([2,3]) & = f^a(g^a([2,3])) \\ &= f^a([6,9])\\ & = [6,9]
    \end{align*}
\end{example}

\section{Abstractly Interpreting Neural Networks}
We have seen how to construct abstract transformers
for a range of functions and how to compose abstract transformers.
We now direct our attention to constructing 
an abstract transformer for a neural network.

Recall that a neural network is defined as a graph 
$G = (\nodes, \edges)$, giving rise to a function $f_G : \R^n \to \R^m$, where $n = |\inodes|$ and $m = |\onodes|$.
Recall that $\inodes$ are input nodes and $\onodes$ are output nodes of $G$.
We would like to construct an abstract transformer
$f^a_G$ that takes $n$ intervals and outputs $m$ intervals.

We define $f^a_G([l_1,u_1],\ldots,[l_n,u_n])$ as follows:
\begin{itemize}
\item
First, for every input node $v_i$, we define 
$$\outs^a(v_i) = [l_i,u_i]$$
Recall that we assume a fixed ordering of 
nodes.
\item
Second, for every non-input node $v$,
we define $$\outs^a(v) = f_v^a(\outs^a(v_1),\ldots,\outs^a(v_k))$$
where $f^a_v$ is the abstract transformer of $f_v$,
and $v$ has the incoming edges $(v_1,v),\ldots,(v_k,v)$,
\item Finally, the output of $f^a_G$
is the set of intervals $\outs^a(v_1),\ldots,\outs^a(v_m)$,
where $v_1,\ldots,v_m$ are the output nodes.

\end{itemize}

\begin{example}
    Consider the following simple neural network $G$:
    \begin{center}
        \begin{tikzpicture}
    
            \draw node at (0, 0) [input] (in) {$\node_1$};
            \draw node at (0, -1) [input] (in2) {$\node_2$};
    
            \draw node at (2, -0.5) [oper] (op) {$\node_3$};
            \draw node at (4, -0.5) [output] (out) {$\node_4$};
    
            \draw[->,thick] (in) -- (op);
            \draw[->,thick] (in2) -- (op);
            \draw[->,thick] (op) -- (out);
        \end{tikzpicture}
    \end{center}
    \noindent
    Assume that $f_{\node_3}(\vec{x}) = 2x_1 + x_2$
    and $f_{\node_4}(x) = \relu(x)$.

    Say we want to evaluate $f_G^a([0,1],[2,3])$.
    We can do this as follows,
    where $f_{v_3}^a$ and $f_{v_4}^a$ follow the definitions
    we discussed above for affine  and monotonically increasing functions, respectively.
    \begin{align*}
        \outs^a(v_1) &= [0,1]\\
        \outs^a(v_2) &= [2,3]\\
        \outs^a(v_3) &= [2*0 + 2,\ 2*1 + 3] = [2,5]\\
        \outs^a(v_4) &= [\relu(2),\relu(5)] = [2,5]
    \end{align*}

    It's nice to see the outputs of every node written
    on the edges of the graph as follows:
    \begin{center}
        \begin{tikzpicture}
    
            \draw node at (0, 0) [input] (in) {$\node_1$};
            \draw node at (0, -1) [input] (in2) {$\node_2$};
    
            \draw node at (2, -0.5) [oper] (op) {$\node_3$};
            \draw node at (4, -0.5) [output] (out) {$\node_4$};
    
            \draw[->,thick] (in) -- (op) node[midway, above] {$[0,1]$};
            \draw[->,thick] (in2) -- (op) node[midway, below] {$[2,3]$};
            \draw[->,thick] (op) -- (out) node[midway, above] {$[2,5]$};
            \draw node at (5, -0.5) [empty] (out) {$[2,5]$};
        \end{tikzpicture}
    \end{center}
    
\end{example}

\subsection*{Limitations of the Interval Domain}
The interval domain, as described, seems infallible.
We will now see how it can, and often does, overshoot:
compute wildly overapproximating solutions.
The primary reason for this is that the interval domain
is non-relational, meaning it cannot keep track 
of relations between different values, e.g., the inputs and outputs of a function.

\begin{example}
Consider the following, admittedly bizarre, neural network:
\begin{center}
    \begin{tikzpicture}
        \draw node at (2, -.5) [input] (in) {$\node_1$};

        \draw node at (4, 0) [oper] (op) {$\node_2$};
        \draw node at (6, -.5) [output] (out) {$\node_3$};

        \draw[->,thick] (in) -- (op) node[midway, above] {$[0,1]$};
        \draw[->,thick] (in) -- (out) node[midway, below] {$[0,1]$};
        \draw[->,thick] (op) -- (out) node[midway, above] {$[-1,0]$};
        \draw node at (7, -0.5) [empty] (out) {$[-1,1]$};
    \end{tikzpicture}
\end{center}
where 
\begin{align*}
    f_{v_2}(x) & = -x \\
    f_{v_3}(\vec{x}) & = x_1+x_2
\end{align*}
Clearly, for any input $x$, $f_G(x) = 0$.
Therefore, ideally, we can define our abstract transformer
simply as $f^a_G([l,u]) = [0,0]$ for any interval $[l,u]$.

Unfortunately, if we follow the recipe above,
we get a much bigger interval than $[0,0]$.
For example, on the input $[0,1]$, $f^a_G$
returns $[-1,1]$,
as illustrated on the graph above.
The reason this happens is because 
the output node, $v_3$, receives two intervals 
as input, not knowing that one is the negation 
of the other.
In other words, it doesn't know the \emph{relation} between, or provenance of, the two intervals.
\end{example}

\begin{example}
Here's another simple network, $G$,
where $f_{v_2}$ and $f_{v_3}$
are ReLUs.
Therefore, $f_G(x) = (x,x)$
for any positive input $x$.

\begin{center}
    \begin{tikzpicture}
        \draw node at (2, -.5) [input] (in) {$\node_1$};

        \draw node at (4, 0) [output] (out) {$\node_2$};
        \draw node at (4, -1) [output] (out1) {$\node_3$};

        \draw[->,thick] (in) -- (out) node[midway, above] {$[0,1]$};
        \draw[->,thick] (in) -- (out1) node[midway, below] {$[0,1]$};
        \draw node at (5, -0) [empty] (out) {$[0,1]$};
        \draw node at (5, -1) [empty] (out) {$[0,1]$};
    \end{tikzpicture}
\end{center}

Following our recipe, we have $f_G^a([0,1]) = ([0,1],[0,1])$.
In other words, the abstract transformer tells us that, for inputs between 0 and 1,
the neural network can output any pair $(x,y)$ where $0 \leq x,y \leq 1$.
But that's too loose an approximation: we should expect to see only outputs $(x,x)$
where $0 \leq x \leq 1$.
Again, we have lost the \emph{relation}
between the two output nodes.
They both should return the same number,
but the interval domain,
and our abstract transformers,
are not strong enough to capture that fact.
\end{example}

\section*{Looking Ahead}
We've seen how interval arithmetic can be used to efficiently evaluate a neural network on a set of inputs, paying the price of efficiency with precision.
Next, we will see more precise abstract domains.

The abstract interpretation framework was introduced by \citet{cousot1977abstract} in their seminal paper.
Abstract interpretation is a general framework, based on lattice theory,
for defining and reasoning about program analyses.
In our exposition, we avoided the use of lattices, because we do not aim for generality---we just want to analyze neural networks.
Nonetheless, the lattice-based formalization allows us to easily construct 
the most-precise abstract transformers for any operation.

Interval arithmetic is an old idea that predates program analysis,  even computer science: it is a standard tool in the natural sciences for measuring accumulated measurement errors. 
For neural-network verification, interval arithmetic first appeared
in a number of papers starting in 2018~\citep{gehr2018ai2,gowal2018effectiveness,wang2018formal}.
To implement interval arithmetic for real neural networks efficiently,
one needs to employ parallel matrix operations (e.g., using a \gpu).
Intuitively, an operation like matrix addition
can be implemented with two matrix additions for interval arithmetic,
one for upper bounds and one for lower bounds.

There are also powerful techniques
that employ the interval domain (or any means to bound the output of various nodes of the network) with search.
We did not cover this combination here but I would encourage you to check out
FastLin approach~\citep{DBLP:conf/icml/WengZCSHDBD18} and its successor, \textsc{crown}~\citep{DBLP:conf/nips/ZhangWCHD18}.
(Both are nicely summarized by \cite{DBLP:conf/nips/LiHALGJ19})

One interesting application of the interval domain is 
as a quick-and-dirty way for speeding up constraint-based verification.
\cite{DBLP:conf/iclr/TjengXT19} propose using something like the interval domain
to bound the interval of values taken by a ReLU for a range of inputs to the neural network.
If the interval of inputs of a ReLU is above or below 0, then we can replace
the ReLU with a linear function, $f(x) = x$ or $f(x) = 0$, respectively.
This simplifies the constraints for constraint-based verification,
as there's no longer a disjunction.

    \chapter{Neural Zonotope Abstraction}\label{ch:numerical}
In the previous chapter,
we defined the interval abstract domain,
which allows us to succinctly capture infinite sets
in $\R^n$ by defining lower and upper bounds per dimension.
In $\R^2$, an interval defines a rectangle;
in $\R^3$, an interval defines a box;
in higher dimensions, it defines hyperrectangles.

The issue with the interval domain
is that it does not \emph{relate} the values 
of various dimensions---it just bounds each dimension.
For example, in $\R^2$, we cannot capture
the set of points where $x=y$ and $0\leq x \leq 1$.
The best we can do is the square region $([0,1],[0,1])$.
Syntactically speaking, an abstract element in the interval domain
is captured by constraints of the form:
\[\bigwedge_i l_i\leq x_i \leq u_i\]
where every inequality involves a single variable,
and therefore no relationships between variables are captured.
So the interval domain is called \emph{non-relational}.
In this chapter, we will look at a \emph{relational} 
abstract domain, the \emph{zonotope domain}, and discuss its application to neural networks.


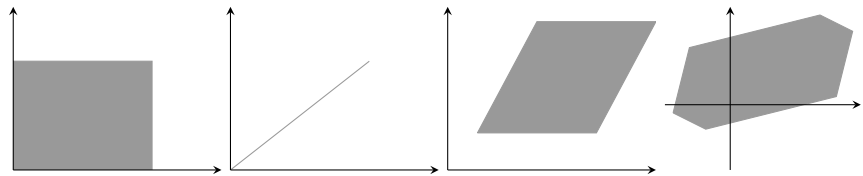
\begin{figure}[h!]
    \centering
        \begin{tikzpicture}
            \begin{axis}[
                axis lines=middle,
                xmax=3,
                xmin=0,
                ymin=0,
                ymax=3,
                xlabel={},
                ylabel={},
                ytick={0},
                xtick ={0},width=4.35cm,
            ]
        
            \draw [fill=black, opacity=0.4] (axis cs:0.0,0) -- (axis cs:0,2)
            --(axis cs:2,2) -- (axis cs:2,0) -- (axis cs:0,0);
            
            \end{axis}
        \end{tikzpicture}
        \begin{tikzpicture}
            \begin{axis}[
                axis lines=middle,
                xmax=3,
                xmin=0,
                ymin=0,
                ymax=3,
                xlabel={},
                ylabel={},
                ytick={0},
                xtick ={0},width=4.35cm,
            ]
        
            \draw [fill=black, opacity=0.4] (axis cs:0.0,0) -- (axis cs:2,2);
            
            \end{axis}
        \end{tikzpicture}
        \begin{tikzpicture}
            \begin{axis}[
                axis lines=middle,
                xmax=3.5,
                xmin=0,
                ymin=0,
                ymax=2.2,
                xlabel={},
                ylabel={},
                ytick={0},
                xtick ={0},width=4.35cm
            ]
        
            \draw [fill=black, opacity=0.4] (axis cs:0.5,0.5) -- (axis cs:1.5,2)
            --(axis cs:3.5,2) -- (axis cs:2.5,.5) -- (axis cs:0.5,.5);
            
            \end{axis}
        \end{tikzpicture}
        \begin{tikzpicture}
            \begin{axis}[
                axis equal image, 
                axis lines=middle,
                xmax=4.0,
                xmin=-2,
                ymin=-2,
                ymax=3.0,
                xlabel={},
                ylabel={},
                ytick={0},
                xtick ={0},
                width=4.35cm
            ]

            \draw [fill=black, opacity=0.4] (axis cs:-1.750000,-0.250000) -- 
            (axis cs:-0.750000,-0.750000) -- (axis cs:3.250000,0.250000) --
            (axis cs:3.750000,2.250000) -- (axis cs:2.750000,2.750000) --
            (axis cs:-1.250000,1.750000) -- (axis cs:-1.750000,-0.250000) ;
            \end{axis}
        \end{tikzpicture}
        \caption{Examples of zonotopes in $\R^2$}\label{fig:zonotope}
    \end{figure}

\section{What the Heck is a Zonotope?}
Let's begin with defining a 1-dimensional \emph{zonotope}.
We assume we have a set of $m$ real-valued \emph{generator} variables,
denoted $\epsilon_1,\ldots,\epsilon_m$.
A 1-dimensional zonotope is the set of all points 
$$\left\{c_0 + \sum_{i=1}^m c_i \cdot \epsilon_i \middle\vert\ \epsilon_i \in [-1,1] \right\}$$
where $c_i \in \R$.

If you work out a few examples of the above definition,
you'll notice that a 1-dimensional zonotope is just a convoluted way of defining an interval.
For example, if we have one generator variable, $\epsilon$, then a zonotope 
is the set 
$$\{c_0 + c_1 \epsilon \mid \epsilon \in [-1,1]\}$$
which is the interval $[c_0 - c_1 ,c_0 + c_1]$, assuming $c_1 \geq 0$.
Note that $c_0$ is the \emph{center} of the interval. 

Zonotopes start being more expressive than intervals in $\R^2$ and beyond.
In $n$-dimensions, a zonotope with $m$ generators is the set of all points 
$$\left\{\left(\underbrace{c_{10} + \sum_{i=1}^m c_{1i}\cdot \epsilon_i}_{\text{first dimension}}\ \ldots,\ \underbrace{c_{n0} + \sum_{i=1}^m c_{ni} \cdot \epsilon_i}_{n\text{th dimension}}\right) \middle\vert\ \epsilon_i \in [-1,1] \right\}$$

This is best illustrated through a series of examples in $\R^2$.\footnote{In the VR edition of the book, I take the reader on a guided 3D journey of zonotopes; since you cheaped out and just downloaded the free pdf, we'll have to do with $\R^2$.}

\begin{example}
    Consider the following two-dimensional zonotope with two generators.
    $$(1 + \epsilon_1,\ 2+\epsilon_2)$$
    where we drop the set notation for clarity.
    Notice that in the first dimension the coefficient of $\epsilon_2$ is 0,
    and in the second dimension the coefficient of $\epsilon_1$ is 0.
    Since the two dimensions do not share generators, 
    we get the following box shape whose center is $(1,2)$.
    \begin{figure}[h!]
        \centering
    \begin{tikzpicture}
        \begin{axis}[
            axis equal image, 
            axis lines=middle,
            xmax=3.5,
            xmin=0,
            ymin=0,
            ymax=3.5,
            xlabel={},
            ylabel={},
            ytick={0,1,2},
            xtick ={0,1,2},width=4.35cm,
        ]
        \draw [fill=black, opacity=0.2] (axis cs:0.0,1) -- (axis cs:0,3)
        --(axis cs:2,3) -- (axis cs:2,1) -- (axis cs:0,1);
        \addplot[Maroon,mark=*] coordinates {(1,2)} node [pos=0,left] {}     ;  
        \end{axis}
    \end{tikzpicture}
    \end{figure}

    Observe that the center of the zonotope is the
    vector of constant coefficients of the two 
    dimensions, $(1,2)$, as illustrated below:
    $$(\underbrace{1} +\ \epsilon_1, \underbrace{2} +\ \epsilon_2)$$

\end{example}

\begin{example}
    Now consider the following zonotope with 1 generator:
    $$(2 + \epsilon_1,\ 2+\epsilon_1)$$
    Since the two dimensions  share the same expression,
    this means that two dimensions are equal,
    and so we get  we get a line shape
    centered at $(2,2)$:
    \begin{center}
        \begin{tikzpicture}
             \begin{axis}[
                 axis equal image, 
                 axis lines=middle,
                 xmax=3.5,
                 xmin=0,
                 ymin=0,
                 ymax=3.5,
                 xlabel={},
                 ylabel={},
                 ytick={1,2,3},
                 xtick ={1,2,3},width=4.35cm]
             \draw [fill=black] (axis cs:1,1) -- (axis cs:3,3);
            \end{axis}
        \end{tikzpicture}
    \end{center}
    
    The reason $\epsilon_1$ is called a generator is because we can think 
    of it as a constructor of a zonotope.
    In this example, starting from the center point (2,2),
    the generator $\epsilon_1$ \emph{stretches} the point (2,2) to (3,3),
    by adding (1,1) (the two coefficients of $\epsilon_1$)
    and stretches the center to (1,1) by subtracting (1,1).
    See the following illustration:
    \begin{center}
        \centering
        \begin{tikzpicture}
             \begin{axis}[
                 axis equal image, 
                 axis lines=middle,
                 xmax=3.5,
                 xmin=0,
                 ymin=0,
                 ymax=3.5,
                 xlabel={},
                 ylabel={},
                 ytick={1,2,3},
                 xtick ={1,2,3},width=4.35cm]
             \draw [->,fill=Maroon,Maroon] (axis cs:2,2) -- (axis cs:3,3);
             \draw [->,fill=Maroon,Maroon] (axis cs:2,2) -- (axis cs:1,1);
            \addplot[Maroon,mark=*] coordinates {(2,2)} node [pos=0,left] {}     ;       
        \end{axis}
        \end{tikzpicture}
    \end{center}
\end{example}

\begin{example}\label{ex:parallelogram}
    Now consider the following zonotope with 2 generators,
    \[(2 + \epsilon_1,\ 3 + \epsilon_1 + \epsilon_2)\]
    which is visualized as follows, with the center point (2,3) in red.

    \begin{center}
        \begin{tikzpicture}
             \begin{axis}[
                 axis lines=middle,
                 xmax=4,
                 xmin=0,
                 ymin=0,
                 ymax=5.2,
                 xlabel={},
                 ylabel={},
                 ytick={1,2,3,4,5},
                 xtick ={1,2,3,4,5},width=5cm]
                 \draw [fill=black, opacity=0.2] (axis cs:1,1) -- (axis cs:1,3) -- (axis cs:3,5) -- (axis cs:3,3) -- (axis cs:1,1);  
                 \addplot[Maroon,mark=*] coordinates {(2,3)} node [pos=0,left] {}     ;    
        \end{axis}
        \end{tikzpicture}
    \end{center}

    Let's see how this zonotope is generated in two steps,
    by considering one generator at a time.
    The coefficients of $\epsilon_1$ are (1,1),
    so it stretches the center point (2,3) along the (1,1) vector,
    generating a line:
    
    \begin{center}
        \begin{tikzpicture}
             \begin{axis}[
                 axis lines=middle,
                 xmax=4,
                 xmin=0,
                 ymin=0,
                 ymax=5.2,
                 xlabel={},
                 ylabel={},
                 ytick={1,2,3,4,5},
                 xtick ={1,2,3,4,5},width=5cm]
                 \draw [->, Maroon] (axis cs:2,3) -- (axis cs:3,4);
                 \draw [->, Maroon] (axis cs:2,3) -- (axis cs:1,2);  
                 \addplot[Maroon,mark=*] coordinates {(2,3)} node [pos=0,left] {}     ;    
        \end{axis}
        \end{tikzpicture}
    \end{center}

    Next, the coefficients of $\epsilon_2$ are (0,1),
    so it stretches \emph{all} points along the $(0,1)$ vector, resulting in the zonotope we plotted earlier:
    \begin{center}
        \centering
        \begin{tikzpicture}
             \begin{axis}[
                 axis lines=middle,
                 xmax=4,
                 xmin=0,
                 ymin=0,
                 ymax=5.2,
                 xlabel={},
                 ylabel={},
                 ytick={1,2,3,4,5},
                 xtick ={1,2,3,4,5},width=5cm]
                 \draw [->, Maroon] (axis cs:2,3) -- (axis cs:3,4);
                 \draw [->, Maroon] (axis cs:2,3) -- (axis cs:1,2); 

                 \draw [->, Maroon] (axis cs:2,3) -- (axis cs:2,4);
                 \draw [->, Maroon] (axis cs:2,3) -- (axis cs:2,2); 
                 
                 \draw [->, Maroon] (axis cs:3,4) -- (axis cs:3,5);
                 \draw [->, Maroon] (axis cs:3,4) -- (axis cs:3,3);

                 \draw [->, Maroon] (axis cs:1,2) -- (axis cs:1,1);
                 \draw [->, Maroon] (axis cs:1,2) -- (axis cs:1,3);

                 \addplot[Maroon,mark=*] coordinates {(2,3)} node [pos=0,left] {}     ;    
                 \draw [fill=black, opacity=0.1] (axis cs:1,1) -- (axis cs:1,3) -- (axis cs:3,5) -- (axis cs:3,3) -- (axis cs:1,1);  
        \end{axis}
        \end{tikzpicture}
    \end{center}
\end{example}

You may have deduced by now that adding more generators adds more faces to the zonotope. For example, the right-most zonotope in \cref{fig:zonotope} uses three generators to produce the three pairs of parallel faces.

\subsection*{A Compact Notation}
Going forward, we will use a compact notation to describe an $n$-dimensional zonotope with $m$ generator variables:
$$\left\{\left(c_{10} + \sum_{i=1}^m c_{1i}\cdot \epsilon_i,\ \ldots,\ c_{n0} + \sum_{i=1}^m c_{ni} \cdot \epsilon_i\right) \middle\vert\ \epsilon_i \in [-1,1] \right\}$$
Specifically, we will define it as a tuple of vectors of coefficients:
$$(\langle c_{10}, \ldots, c_{1m}\rangle, \ldots, \langle c_{n0}, \ldots, c_{nm}\rangle)$$
For an even more compact presentation, will also use 
$$(\langle c_{1i}\rangle_i, \ldots, \langle c_{ni}\rangle_i)$$
where $i$ ranges from $0$ to $m$, the number of generators; we drop the index $i$ when it's clear from context.

We can compute the upper bound of the zonotope (the largest possible value)
in the $j$ dimension by solving the following optimization problem:
\begin{align*}
&\max\ c_{j0} + \sum_{i=1}^m c_{ji}\epsilon_{i}\\
&\text{ s.t. }\ \epsilon_i \in [-1,1]
\end{align*}
This can be easily solved by setting $\epsilon_i$ to $1$ if $c_{ji} > 0$
and $-1$ otherwise.

Similarly, we can compute the lower bound of the zonotope in the $j$th dimension by minimizing instead of maximizing, and solving the optimization problem by setting $\epsilon_i$ to $-1$ if $c_{ji} > 0$ and $1$ otherwise.

\begin{example}
    Recall our parallelogram from \cref{ex:parallelogram}:
    $$(2 + \epsilon_1,\ 3 + \epsilon_1 + \epsilon_2)$$
    In our compact notation, we write this as
    $$(\langle2,1,0\rangle, \langle 3,1,1\rangle)$$
    The upper bound in the vertical dimension, $3+\epsilon_1+\epsilon_2$, is 
    $$3 + 1 + 1 = 5$$
    where $\epsilon_1$ and $\epsilon_2$ are set to $1$.
\end{example}

\section{Basic Abstract Transformers}
Now that we have seen zonotopes, let's define some abstract transformers over zonotopes.

\subsection*{Addition}
For addition, $f(x,y) = x+y$,
we will define the abstract transformer $f^a$ that takes a two-dimensional 
zonotope defining a set of values of $(x,y)$.
We will assume a fixed number of generators $m$.
So, for addition,
its abstract transformer is of the form
$$f^a(\langle c_{10}, \ldots, c_{1m}\rangle, \langle c_{20}, \ldots, c_{2m}\rangle)$$
Compare this to the interval domain, where $f^a([l_1,u_1], [l_2,u_2])$

It turns out that addition over zonotopes is straightforward: we just sum up the coefficients:
$$f^a(\langle c_{10}, \ldots, c_{1m}\rangle, \langle c_{20}, \ldots, c_{2m}\rangle) = \langle c_{10} + c_{20}, \ldots, c_{1m} + c_{2m}\rangle$$

\begin{example}
Consider the simple zonotope $(0 + \epsilon_1, 1 + \epsilon_2)$.
This represents the following box:

\begin{center}
    \centering
\begin{tikzpicture}
    \begin{axis}[
        axis equal image, 
        axis lines=middle,
        xmax=1.5,
        xmin=-1.5,
        ymin=-1.2,
        ymax=2.5,
        xlabel={},
        ylabel={},
        ytick={0,1,2},
        xtick ={-1,0,1},width=4.35cm,
    ]
    \draw [fill=black, opacity=0.2] (axis cs:-1,0) -- (axis cs:-1,2)
    --(axis cs:1,2) -- (axis cs:1,0) -- (axis cs:-1,0);
    \end{axis}
\end{tikzpicture}
\end{center}

The set of possible values we can get by adding the $x$ and $y$ dimensions in this box 
is the interval between $-1$ and $3$.
Following the definition of the abstract transformer for addition:
$$f^a(\langle 0,1,0\rangle, \langle 1,0,1\rangle) = \langle 1,1,1 \rangle $$
That is the output zonotope is the set 
$$\{1+\epsilon_1+\epsilon_2 \mid \epsilon_1,\epsilon_2 \in [-1,1]\}$$
which is the interval $[-1,3]$.
\end{example}

\subsection*{Affine Functions}
For an affine function $$f(x_1,\ldots,x_n) = \sum_j a_jx_j$$
where $a_j \in \R$,
we can define the abstract transformer as follows:
$$f^a(\langle c_{1i}\rangle,\ldots\langle c_{ni} \rangle) = 
\left\langle \sum_j a_jc_{j0}, \ldots, \sum_j a_jc_{jm} \right\rangle
$$
Intuitively, we apply $f$ to the center point and coefficients of $\epsilon_1$, $\epsilon_2$, etc.

\begin{example}
Consider $f(x,y) = 3x + 2y$.
Then, 
\begin{align*}
f^a(\langle 1,2,3 \rangle, \langle 0,1,1 \rangle) & =
\langle f(1,0), f(2,1), f(3,1) \rangle \\
&=  \langle 3,8,11 \rangle
\end{align*}
\end{example}

\section{Abstract Transformers of Activation Functions}
We now discuss how to construct an abstract transformer for the ReLU activation function.

\subsection*{Limitations of the Interval Domain}
Let's first recall the interval abstract transformer of ReLU:
$$\relu^a([l,u]) = [\relu(l),\relu(u)]$$
The issue with the interval domain is we don't know how points
in the output interval $\relu^a([l,u])$ relate to the input interval $[l,u]$---i.e., which inputs are responsible for which outputs.

Geometrically, we think of the interval domain as approximating the ReLU function with a box as follows:
\begin{center}
\begin{tikzpicture}
    \begin{axis}[
        axis lines=middle,
        xmax=2,
        xmin=-2,
        ymin=-.7,
        ymax=3,
        xlabel={},
        ylabel={},
        ytick={0},
        xtick ={0},width=5cm,
    ]
    \draw [fill=black, opacity=0.2] (axis cs:-1.5,0) -- (axis cs:-1.5,1.5)
    --(axis cs:1.5,1.5) -- (axis cs:1.5,0) -- (axis cs:-1.5,0);
    \addplot [domain=-10:8, samples=500,
              thick, Maroon] {max(0,x)};
     \addplot[Maroon,mark=*] coordinates {(1.5,0)} node [pos=0,below] {$u$}     ;    
     \addplot[Maroon,mark=*] coordinates {(-1.5,0)} node [pos=0,below] {$l$}     ;    
     \addplot[Maroon,mark=*] coordinates {(0,1.5)} node [pos=0,above right] {$\relu(u)$}     ;

    \end{axis}
\end{tikzpicture}
~~~~~~~~~~~~
\begin{tikzpicture}
    \begin{axis}[
        axis lines=middle,
        xmax=2,
        xmin=-1.5,
        ymin=-.7,
        ymax=3,
        xlabel={},
        ylabel={},
        ytick={0},
        xtick ={0},width=5cm,
    ]
    \draw [fill=black, opacity=0.2] (axis cs:0.5,0.5) -- (axis cs:0.5,1.5)
    --(axis cs:1.5,1.5) -- (axis cs:1.5,0.5) -- (axis cs:0.5,0.5);
    \addplot [domain=-3:8, samples=500,
              thick, Maroon] {max(0,x)};
     \addplot[Maroon,mark=*] coordinates {(1.5,0)} node [pos=0,below] {$u$}     ;    
     \addplot[Maroon,mark=*] coordinates {(0.5,0)} node [pos=0,below ] {$l$}     ;    
     \addplot[Maroon,mark=*] coordinates {(0,1.5)} node [pos=0,left] {$\relu(u)$}     ;    
     \addplot[Maroon,mark=*] coordinates {(0,.5)} node [pos=0, left] {$\relu(l)$}     ;  
    \end{axis}
\end{tikzpicture}
\end{center}
The figure on the left shows the case where the lower bound is negative and the upper bound is positive; the right figure shows the case where the lower bound is positive.

\subsection*{A Zonotope Transformer for ReLU}
Let's slowly build the ReLU abstract transformer for zonotopes.
We're given a 1-dimensional zonotope $\langle c_i \rangle_i$ as input.
We will use $u$ to denote the upper bound of the zonotope
and $l$ the lower bound. 

\[
    \relu^a(\langle c_i \rangle_i) = 
     \begin{cases*}
    \langle c_i \rangle_i & for $l \geq 0$\\
    \langle 0 \rangle_i & \text{for } $u \leq 0$\\
    ? & \text{otherwise}
    \end{cases*}
\]

If $l \geq 0$, then we simply return the input zonotope back;
if $ u \leq 0$, then the answer is 0;
when the zonotope has both negative and positive values,
there are many ways to define the output, and so I've left it as a question mark.
The easy approach is to simply return the interval $[l,u]$ encoded as a zonotope.
But it turns out that we can do better:
since zonotopes allow us to relate inputs and outputs,
we can \emph{shear} a box into a parallelogram that fits the shape of ReLU more tightly, as follows:
\begin{figure}[h!]
    \centering
    \begin{tikzpicture}
        \begin{axis}[
            axis lines=middle,
            xmax=2,
            xmin=-2,
            ymin=-.7,
            ymax=3,
            xlabel={},
            ylabel={},
            ytick={0},
            xtick ={0},width=6cm,
        ]
        \draw [fill=black, opacity=0.2] (axis cs:-1,0) -- (axis cs:-1,1)
        --(axis cs:1,1) -- (axis cs:1,0) -- (axis cs:-1,0);
        \addplot [domain=-10:8, samples=500,
                  thick, Maroon] {max(0,x)};
         \addplot[Maroon,mark=*] coordinates {(1,0)} node [pos=0,below] {$u$}     ;    
         \addplot[Maroon,mark=*] coordinates {(-1,0)} node [pos=0,below] {$l$}     ;    
         \addplot[Maroon,mark=*] coordinates {(0,1)} node [pos=0,above right] {$\relu(u)$}     ;

        \end{axis}
    \end{tikzpicture}
    \quad\quad\quad
\begin{tikzpicture}
    \begin{axis}[
        axis lines=middle,
        xmax=2,
            xmin=-2,
            ymin=-.7,
            ymax=3,
        xlabel={},
        ylabel={},
        ytick={0},
        xtick ={0},width=6cm,
    ]
    \draw [fill=black, opacity=0.2] (axis cs:-1,-0.5) -- (axis cs:-1,0)
    --(axis cs:1,1) -- (axis cs:1,0.5) -- (axis cs:-1,-.5);
    \addplot [domain=-10:8, samples=500,
              thick, Maroon] {max(0,x)};
     \addplot[Maroon,mark=*] coordinates {(1,0)} node [pos=0,below] {$u$}     ;    
     \addplot[Maroon,mark=*] coordinates {(-1,0)} node [pos=0,below left] {$l$}     ;  
     \addplot[Maroon,mark=*] coordinates {(0,1)} node [pos=0,above right] {$\relu(u)$}     ;    
     \draw [fill=black, dotted] (axis cs:0,1) -- (axis cs:1,1);
     \draw [fill=black, dotted] (axis cs:1,0) -- (axis cs:1,1);
    \end{axis}
\end{tikzpicture}
\end{figure}

The approximation on the right has a smaller area than the approximation afforded by the interval domain on the left.
The idea is that a smaller area results in a better approximation,
albeit an incomparable one, as the parallelogram returns negative values,
while the box doesn't.
Let's try to describe this parallelogram as a zonotope.

The bottom face of the zonotope is the line $$y = \lambda x$$ for some slope $\lambda$.
It follows that the top face must be $$y = \lambda x + u (1-\lambda)$$
If we set $\lambda = 0$, we get two horizontal faces, i.e.,  the interval approximation shown above.
The higher we crank up $\lambda$, the tighter the parallelogram gets.
But, we can't increase $\lambda$ past $u/(u-l)$; this ensures that the parallelogram covers the ReLU along the input range $[l,u]$. So, we will set $$\lambda = \frac{u}{u-l}$$

It follows that the distance between the top and bottom faces of the parallelogram
is $u(1-\lambda)$,.
Therefore, the center of the zonotope (in the vertical axis) must be the point
$$\eta = \frac{u(1-\lambda)}{2}$$

With this information, we can complete the definition of $\relu^a$ as follows:
\[
    \relu^a(\langle c_1,\ldots,c_m \rangle) = 
     \begin{cases*}
    \langle c_i \rangle_i, & for $l \geq 0$\\
    \langle 0 \rangle_i, & \text{for} $u \leq 0$\\
    \langle \lambda c_1,\ldots,\lambda c_m,0 \rangle + \langle \eta,0,0,\ldots,\eta \rangle  & \text{otherwise}
    \end{cases*}
\]
There are two non-trivial things we do here:
\begin{itemize}
\item First, we add a new generator, $\epsilon_{m+1}$,
in order to stretch the parallelogram in the vertical axis;
its coefficient is $\eta$, which is half the hight of the parallelogram.
\item Second, we add the input zonotope scaled by $\lambda$
with coefficient 0 for the new generator;
this ensures that we capture the relation between the input and output.
\end{itemize}
Let's look at an example for clarity:

\begin{example}
    Say we invoke $\relu^a$ with the interval between $l = -1$ and $ u = 1$, i.e.,
    $$\relu^a(\langle 0, 1 \rangle)$$
    Here, $\lambda = 0.5$ and $\eta = 0.25$.
    So the result of $\relu^a$ is the following zonotope:
    \begin{align*}
        \langle 0,0.5,0 \rangle + \langle 0.25,0,0.25\rangle =\langle 0.25, 0.5, 0.25\rangle
    \end{align*}
    The 2-dimensional zonotope composed of the input and output zonotopes of $\relu^a$ is
    $$(\langle 0, 1, 0 \rangle, \ \langle 0.25, 0.5, 0.25\rangle)$$
    or, explicitly,
    $$(0 + \epsilon_1, \ 0.25 + 0.5\epsilon_1 + 0.25\epsilon_2)$$
    This zonotope, centered at $(0,0.25)$,
    is illustrated below:
    \begin{center}
        \centering
    \begin{tikzpicture}
        \begin{axis}[
            axis lines=middle,
            xmax=1.5,
                xmin=-1.5,
                ymin=-.7,
                ymax=1.5,
            xlabel={},
            ylabel={},
            ytick={0},
            xtick ={0},width=6cm,
        ]
        \draw [fill=black, opacity=0.2] (axis cs:-1,-0.5) -- (axis cs:-1,0)
        --(axis cs:1,1) -- (axis cs:1,0.5) -- (axis cs:-1,-.5);
        \addplot [domain=-10:8, samples=500,
                  thick, Maroon] {max(0,x)};
         \addplot[Maroon,mark=*] coordinates {(1,0)} node [pos=0,below] {$1$}     ;    
         \addplot[Maroon,mark=*] coordinates {(-1,0)} node [pos=0,below left] {$-1$}     ;  
         \addplot[Maroon,mark=*] coordinates {(0,1)} node [pos=0,above right] {$1$}     ;    
         \addplot[Maroon,mark=*] coordinates {(0,.25)} node [pos=0,above left] {$\eta = 0.25$}     ;  
         \draw [fill=black, dotted] (axis cs:0,1) -- (axis cs:1,1);
         \draw [fill=black, dotted] (axis cs:1,0) -- (axis cs:1,1);
        \end{axis}
    \end{tikzpicture}
    \end{center}

\end{example}

\subsection*{Other Abstract Transformers}
We saw how to design an abstract transformer for ReLU.
We can follow a similar approach to design abstract transformers for sigmoid.
It is indeed a good exercise to spend some time designing a zonotope transformer for sigmoid or tanh---and don't look at the literature \citep{singh2018fast}!

It is interesting to note that as the abstract domain gets richer---allowing crazier and crazier shapes---the more incomparable abstract transformers you can derive~\citep{DBLP:conf/popl/0001NA14}. With the interval abstract domain, which is the simplest you can go
without being trivial, the best you can do is a box to approximate a ReLU or a sigmoid.
But with zonotopes, there are infinitely many shapes that you can come up with.
So designing abstract transformers becomes an art, and it's hard to predict which transformers will do well in practice.

\section{Abstractly Interpreting Neural Networks with Zonotopes}
We can now use our zonotope abstract transformers 
to abstractly interpret an entire neural network
in precisely the same way we did intervals.
We review the process here for completeness.

Recall that a neural network is defined as a graph 
$G = (\nodes, \edges)$, giving rise to a function $f_G : \R^n \to \R^m$, where $n = |\inodes|$ and $m = |\onodes|$.
We would like to construct an abstract transformer
$f^a_G$ that takes an $n$-dimensional zonotope and outputs an $m$-dimensional zonotope.

We define $f^a_G(\langle c_{1j}\rangle,\ldots\langle c_{nj} \rangle)$ as follows:
\begin{itemize}
\item
First, for every input node $v_i$, we define 
$$\outs^a(v_i) = \langle c_{ij} \rangle_j$$
Recall that we assume a fixed ordering of 
nodes.
\item
Second, for every non-input node $v$,
we define $$\outs^a(v) = f_v^a(\outs^a(v_1),\ldots,\outs^a(v_k))$$
where $f^a_v$ is the abstract transformer of $f_v$,
and $v$ has the incoming edges $(v_1,v),\ldots,(v_k,v)$,
\item Finally, the output of $f^a_G$
is the $m$-dimensional zonotope $$(\outs^a(v_1),\ldots,\outs^a(v_m))$$
where $v_1,\ldots,v_m$ are the output nodes.

\end{itemize}

One thing to note is that some abstract transformers (for activation functions)
add new generators. 
We can assume that all of these generators are already in the input zonotope but with coefficients set to 0, and they only get non-zero coefficients 
in the outputs of activation function nodes.

\begin{example}
    Consider the following neural network, which we saw in the last chapter,
    \begin{center}
        \begin{tikzpicture}
            \draw node at (2, -.5) [input] (in) {$\node_1$};
    
            \draw node at (4, 0) [oper] (op) {$\node_2$};
            \draw node at (6, -.5) [output] (out) {$\node_3$};
    
            \draw[->,thick] (in) -- (op) node[midway, above] {$ $};
            \draw[->,thick] (in) -- (out) node[midway, below] {$ $};
            \draw[->,thick] (op) -- (out) node[midway, above] {$ $};
            \draw node at (7.1, -0.5) [empty] (out) {};
        \end{tikzpicture}
    \end{center}
    where 
    \begin{align*}
        f_{v_2}(x) & = -x \\
        f_{v_3}(\vec{x}) & = x_1+x_2
    \end{align*}
    Clearly, for any input $x$, $f_G(x) = 0$.
    Consider any input zonotope $\langle c_i \rangle$.
    The output node, $v_3$, receives the two-dimensional zonotope
    $$(\langle -c_i \rangle, \langle c_i \rangle)$$
    The two dimensions cancel each other out, resulting in the zonotope $\langle 0 \rangle$,
    which is the singleton set  $\{0\}$.
    
    In contrast, with the interval domain, given input interval $[0,1]$,
    you get the output interval $[-1,1]$.
\end{example}

\section*{Looking Ahead}
We've seen the zonotope domain, an efficient extension beyond simple interval arithmetic.
Next, we will look at full-blown polyhedra.

To my knowledge, the zonotope domain was first introduced by \citet{girard2005reachability} in the context of hybrid-system model checking.
In the context of neural-network verification, \citet{gehr2018ai2}
were the first to use zonotopes,
and introduced precise abstract transformers~\citep{singh2018fast},
one of which we covered here.
In practice, we try to limit the number of generators
to keep verification fast. This can be done 
by occasionally \emph{projecting out} some of the generators
heuristically as we're abstractly interpreting the neural network.

A standard optimization in program analysis is to combine
program operations and construct more precise abstract transformers
for the combination.
This allows us to extract more relational information.
In the context of neural networks, this amounts to combining activation functions in a layer of the network.
\citet{singh2019beyond} showed how to elegantly do this for zonotopes.

    \chapter{Neural Polyhedron Abstraction}\label{ch:polyhedra}

In the previous chapter, we saw the zonotope abstract domain,
which is more expressive than the interval domain.
Specifically, instead of approximating functions using a hyperrectangle,
the zonotope domain allows us to approximate functions using a zonotope, e.g., a parallelogram,
capturing relations between different dimensions.

In this section, we look at an even more expressive abstract domain,
the \emph{polyhedron domain}.
Unlike the zonotope domain, the polyhedron domain allows us to approximate
functions using arbitrary \emph{convex polyhedra}.
A polyhedron in $\R^n$ is a region made of straight (as opposed to curved) faces; a convex shape is one where the line between any two points in the shape 
is completely contained in the shape.
Convex polyhedra can be specified as a set of linear inequalities.
Using convex polyhedra, we approximate a ReLU as follows:

\begin{center}
\begin{tikzpicture}
    \begin{axis}[
        axis lines=middle,
        xmax=2,
            xmin=-2,
            ymin=-.7,
            ymax=3,
        xlabel={},
        ylabel={},
        ytick={0},
        xtick ={0},width=6cm,
    ]
    \draw [fill=black, opacity=0.2] (axis cs:-1,0) -- (axis cs:0,0)
    --(axis cs:1,1) -- (axis cs:-1,0);
    \addplot [domain=-10:8, samples=500,
              thick, Maroon] {max(0,x)};
     \addplot[Maroon,mark=*] coordinates {(1,0)} node [pos=0,below] {$u$}     ;    
     \addplot[Maroon,mark=*] coordinates {(-1,0)} node [pos=0,below left] {$l$}     ;  
     \addplot[Maroon,mark=*] coordinates {(0,1)} node [pos=0,above right] {$\relu(u)$}     ;    
     \draw [fill=black, dotted] (axis cs:0,1) -- (axis cs:1,1);
     \draw [fill=black, dotted] (axis cs:1,0) -- (axis cs:1,1);
    \end{axis}
\end{tikzpicture}
\end{center}

This is the smallest convex polyhedron that approximates ReLU.
You can visually check that it is convex.
This approximation is clearly more precise than that afforded by the interval and zonotope domains, as it is fully contained in the approximations of ReLU in
 those domains: 

\begin{figure}[h!]
    \centering
    \begin{tikzpicture}
        \begin{axis}[
            axis lines=middle,
            xmax=2,
            xmin=-2,
            ymin=-.7,
            ymax=3,
            xlabel={},
            ylabel={},
            ytick={0},
            xtick ={0},width=6cm,
        ]
        \draw [fill=black, opacity=0.2] (axis cs:-1,0) -- (axis cs:-1,1)
        --(axis cs:1,1) -- (axis cs:1,0) -- (axis cs:-1,0);
        \addplot [domain=-10:8, samples=500,
                  thick, Maroon] {max(0,x)};
         \addplot[Maroon,mark=*] coordinates {(1,0)} node [pos=0,below] {$u$}     ;    
         \addplot[Maroon,mark=*] coordinates {(-1,0)} node [pos=0,below] {$l$}     ;    
         \addplot[Maroon,mark=*] coordinates {(0,1)} node [pos=0,above right] {$\relu(u)$}     ;

        \end{axis}
    \end{tikzpicture}
    \quad\quad\quad
\begin{tikzpicture}
    \begin{axis}[
        axis lines=middle,
        xmax=2,
            xmin=-2,
            ymin=-.7,
            ymax=3,
        xlabel={},
        ylabel={},
        ytick={0},
        xtick ={0},width=6cm,
    ]
    \draw [fill=black, opacity=0.2] (axis cs:-1,-0.5) -- (axis cs:-1,0)
    --(axis cs:1,1) -- (axis cs:1,0.5) -- (axis cs:-1,-.5);
    \addplot [domain=-10:8, samples=500,
              thick, Maroon] {max(0,x)};
     \addplot[Maroon,mark=*] coordinates {(1,0)} node [pos=0,below] {$u$}     ;    
     \addplot[Maroon,mark=*] coordinates {(-1,0)} node [pos=0,below left] {$l$}     ;  
     \addplot[Maroon,mark=*] coordinates {(0,1)} node [pos=0,above right] {$\relu(u)$}     ;    
     \draw [fill=black, dotted] (axis cs:0,1) -- (axis cs:1,1);
     \draw [fill=black, dotted] (axis cs:1,0) -- (axis cs:1,1);
    \end{axis}
\end{tikzpicture}
\end{figure}

\section{Convex Polyhedra}
We will define a polyhedron in a manner analogous to a zonotope,
using a set of $m$ generator variables, $\epsilon_1,\ldots,\epsilon_m$.
With zonotopes the generators are bounded in the interval $[-1,1]$;
with polyhedra, generators are bounded by a set of linear inequalities.

Let's first revisit and generalize the definition of a zonotope.
A zonotope in $\R^n$ is a set of points defined as follows:
$$\left\{\left(c_{10} + \sum_{i=1}^m c_{1i}\cdot \epsilon_i,\ \ldots,\ c_{n0} + \sum_{i=1}^m c_{ni} \cdot \epsilon_i\right) \middle\vert\ \varphi(\epsilon_1,\ldots,\epsilon_m) \right\}$$
where $\varphi$ is a Boolean function that evaluates to true iff
all of its arguments are between $-1$ and $1$.

With polyhedra, we will define $\varphi$ as a set (conjunction) of linear inequalities 
over the generator variables, e.g.,
$$0\leq \epsilon_1 \leq 5 \ \land\ \epsilon_1 = \epsilon_2$$
(equalities are defined as two inequalities). 
We will always assume that $\varphi$ defines a bounded polyhedron, i.e., gives a lower and upper bound for each generator; e.g., $\epsilon_1 \leq 0$ is not allowed, because it does not enforce a lower bound on $\epsilon_1$.

In the 1-dimensional case, a polyhedron is simply an interval.
Let's look at higher dimensional examples:

\begin{example}\label{ex:triangle}
    Consider the following 2-dimensional polyhedron:
    $$\left\{\left(\epsilon_1, \epsilon_2\right) \mid \varphi(\epsilon_1,\epsilon_2)  \right\}$$
    where $$\varphi \equiv 0 \leq \epsilon_1 \leq 1\ \land
\ \epsilon_2 \leq \epsilon_1\ \land\ \epsilon_2 \geq 0$$
    This polyhedron is illustrated as follows:

    \begin{center}
        \centering
        \begin{tikzpicture}
            \begin{axis}[
                axis lines=middle,
                xmax=1.5,
                xmin=-0.5,
                ymin=-.5,
                ymax=1.5,
                xlabel={},
                ylabel={},
                ytick={0},
                xtick ={0},width=6cm,
            ]
            \draw [fill=black, opacity=0.2] (axis cs:0,0) -- (axis cs:1,0)
            --(axis cs:1,1) -- (axis cs:0,0);
            \end{axis}
        \end{tikzpicture}
    \end{center}
    Clearly, this shape is not a zonotope, because its faces are not parallel.
\end{example}

\begin{example}
    In 3 dimensions, a polyhedron may look something like this\footnote{Adapted from \citet{westburg}.}
    \begin{center}
    \begin{tikzpicture}[scale=2]

        \coordinate (A1) at (-1,0);
        \coordinate (A2) at (0.6,0.2);
        \coordinate (A3) at (1,0);
        \coordinate (A4) at (0.4,-0.2);
        \coordinate (B1) at (0.5,0.5);
        \coordinate (B2) at (0.5,-0.5);
        
        \begin{scope}[dashed,Maroon,opacity=0.6]
        \draw (A1) -- (A2) -- (A3);
        \draw (B1) -- (A2) -- (B2);
        \end{scope}
        \draw[solid][line width=0.5pt] (A1) -- (A4) -- (B1);
        \draw[solid][line width=0.5pt] (A1) -- (A4) -- (B2);
        \draw[solid][line width=0.5pt] (A3) -- (A4) -- (B1);
        \draw[solid][line width=0.5pt] (A3) -- (A4) -- (B2);
        \draw[solid][line width=0.5pt] (B1) -- (A1) -- (B2) -- (A3) --cycle;
        \end{tikzpicture}
    \end{center}
    One can add more faces by adding more linear inequalities to $\varphi$.
\end{example}

From now on, given a polyhedron 
$$\left\{\left(c_{10} + \sum_{i=1}^m c_{1i}\cdot \epsilon_i,\ \ldots,\ c_{n0} + \sum_{i=1}^m c_{ni} \cdot \epsilon_i\right) \middle\vert\ \varphi(\epsilon_1,\ldots,\epsilon_m) \right\}$$
we will abbreviate it as the tuple:
$$(\langle c_{1i}\rangle_i,\ldots\langle c_{ni} \rangle_i, \varphi)$$

\section{Computing Upper and Lower Bounds}
Given a polyhedron $(\langle c_{1i}\rangle_i,\ldots\langle c_{ni} \rangle_i, \varphi)$,
we will often want to compute the lower and upper bounds of one of the dimensions.
Unlike with the interval and zonotope domains, this process is not straightforward. Specifically, it involves solving a linear program,
which takes polynomial time in the number of variables and constraints.

To compute the lower bound of the $j$th dimension,
we solve the following linear programming problem:
\begin{align*}
    & \min c_{j0} + \sum_{i=1}^m c_{ji} \epsilon_i\\
     & \ \text{s.t. } \varphi
\end{align*}
Similarly, we compute the upper bound of the $j$th dimension by maximizing instead of minimizing. 

\begin{example}
    Take our triangle shape from \cref{ex:triangle},
    defined using two generators:
    $$(\langle 0,1,0 \rangle, \langle 0,0,1 \rangle, \varphi)$$
    where $$\varphi \equiv 0 \leq \epsilon_1 \leq 1\ \land\ \epsilon_2 \leq \epsilon_1\ \land\ \epsilon_2 \geq 0$$
    To compute the upper bound of first dimension, we solve
    \begin{align*}
        & \max  \epsilon_1 \\
    & \ \text{s.t. } \varphi
    \end{align*}
    The answer here is 1, which is obvious from the constraints.
\end{example}

\section{Abstract Transformers for Polyhedra}
We're now ready to go over some abstract transformers for polyhedra.

\subsection*{Affine Functions}
For affine functions, it is really the same transformer as the one for the zonotope domain, except that we carry around the set of linear inequalities $\varphi$---for the zonotope domain, $\varphi$ is fixed throughout.

Specifically, for an affine function $$f(x_1,\ldots,x_n) = \sum_j a_jx_j$$
where $a_j \in \R$,
we can define the abstract transformer as follows:
$$f^a(\langle c_{1i}\rangle,\ldots\langle c_{ni} \rangle, \varphi) = 
\left(\left\langle \sum_j a_jc_{j0}, \ldots, \sum_j a_jc_{jm} \right\rangle, \varphi\right)
$$

Notice that the set of linear inequalities does not change between the input and output of the function---i.e., there are no new constraints added.

\begin{example}
Consider $f(x,y) = 3x + 2y$ .
Then, $$f^a(\langle 1,2,3 \rangle, \langle 0,1,1 \rangle, \varphi)  =  (\langle 3,8,11 \rangle, \varphi) $$
\end{example}

\subsection*{Rectified Linear Unit}
Let's now look at the abstract transformer for ReLU,
which we illustrated earlier in the chapter:

\begin{center}
\begin{tikzpicture}
    \begin{axis}[
        axis lines=middle,
        xmax=2,
            xmin=-2,
            ymin=-.7,
            ymax=3,
        xlabel={},
        ylabel={},
        ytick={0},
        xtick ={0},width=6cm,
    ]
    \draw [fill=black, opacity=0.2] (axis cs:-1,0) -- (axis cs:0,0)
    --(axis cs:1,1) -- (axis cs:-1,0);
    \addplot [domain=-10:8, samples=500,
              thick, Maroon] {max(0,x)};
     \addplot[Maroon,mark=*] coordinates {(1,0)} node [pos=0,below] {$u$}     ;    
     \addplot[Maroon,mark=*] coordinates {(-1,0)} node [pos=0,below left] {$l$}     ;  
     \addplot[Maroon,mark=*] coordinates {(0,1)} node [pos=0,above right] {$\relu(u)$}     ;    
     \draw [fill=black, dotted] (axis cs:0,1) -- (axis cs:1,1);
     \draw [fill=black, dotted] (axis cs:1,0) -- (axis cs:1,1);
    \end{axis}
\end{tikzpicture}
\end{center}

This is the tightest convex polyhedron we can use to approximate the ReLU function. We can visually verify that tightening the shape any further will either make it not an approximation or not convex---e.g., by bending the top face downwards, we get a better approximation but lose convexity.

Let's see how to formally define $\relu^a$.
The key point is that the top face is the line $$y = \frac{u (x - l)}{u-l}$$
This is easy to check using vanilla geometry.
Now, our goal is to define the shaded region, which is bounded by $y=0$ from below, 
$y = x$ from the right, and $y = \frac{u (x - l)}{u-l}$ from above.

We therefore define $\relu^a$ as follows:
$$\relu^a(\langle c_{i}\rangle_i, \varphi) = (\langle \underbrace{0,0,\ldots,0}_m,1\rangle, \varphi')$$
where 
\begin{align*}
    \varphi' \equiv \varphi & \land \epsilon_{m+1} \leq \frac{u( \langle c_{i}\rangle - l)}{(u-l)} \\
    & \land \epsilon_{m+1} \geq 0 \\
    & \land \epsilon_{m+1} \geq \langle c_{i} \rangle
\end{align*}
There are a number of things to note here:
\begin{itemize}
     \item $l$ and $u$ are the lower and upper bounds of the input polyhedron,
which can be computed using linear programming.
\item  $\langle c_{i}\rangle_i$ is used to denote the full term
$c_0 + \sum_{i=1}^m c_i\epsilon_i$.
\item 
Observe  that we've added a new generator, $\epsilon_{m+1}$.
The new set of constraints $\varphi'$ relate this new generator 
to the input, effectively defining the shaded region.
\end{itemize}

\begin{example}
Consider the 1-dimensional polyhedron 
$$(\langle 0,1\rangle ,\ -1 \leq \epsilon_1 \leq 1)$$
which is the interval between $-1$ and $1$. 
Invoking $$\relu^a(\langle 0,1\rangle , -1 \leq \epsilon_1 \leq 1)$$
results in $(\langle 0,0,1\rangle, \varphi')$,
where 
\begin{align*}
    \varphi' \equiv &-1 \leq \epsilon_1 \leq 1 \\
    & \land \epsilon_{2} \leq \frac{ \epsilon_1  +1}{2} \\
    & \land \epsilon_{2} \geq 0 \\
    & \land \epsilon_{2} \geq \epsilon_1 
\end{align*}
If we plot the region defined by $\varphi'$, 
using $\epsilon_1$ as the $x$-axis and $\epsilon_2$ as the $y$-axis,
we get the shaded region 

\begin{center}
    \begin{tikzpicture}
        \begin{axis}[
            axis lines=middle,
            xmax=2,
                xmin=-2,
                ymin=-.7,
                ymax=3,
            xlabel={},
            ylabel={},
            ytick={0},
            xtick ={0},width=6cm,
        ]
        \draw [fill=black, opacity=0.2] (axis cs:-1,0) -- (axis cs:0,0)
        --(axis cs:1,1) -- (axis cs:-1,0);
        \addplot [domain=-10:8, samples=500,
                  thick, Maroon] {max(0,x)};
         \addplot[Maroon,mark=*] coordinates {(1,0)} node [pos=0,below] {$1$}     ;    
         \addplot[Maroon,mark=*] coordinates {(-1,0)} node [pos=0,below left] {$-1$}     ;  
         \addplot[Maroon,mark=*] coordinates {(0,1)} node [pos=0,above right] {$1$}     ;    
         \draw [fill=black, dotted] (axis cs:0,1) -- (axis cs:1,1);
         \draw [fill=black, dotted] (axis cs:1,0) -- (axis cs:1,1);
        \end{axis}
    \end{tikzpicture}
    \end{center}

\end{example}

\subsection*{Other Activation Functions}
For ReLU, the transformer we presented is the most precise.
For other activation functions, like sigmoid,
there are many ways to define abstract transformers
for the polyhedron domain.
Intuitively, one can keep adding more and more faces to the polyhedron to get a more precise approximation
of the sigmoid curve.

\section{Abstractly Interpreting Neural Networks with Polyhedra}

We can now use our abstract transformers 
to abstractly interpret an entire neural network,
in precisely the same way we did for zonotopes,
except that we're now carrying around a set of constraints.
We review the process here for completeness.

Recall that a neural network is defined as a graph 
$G = (\nodes, \edges)$, giving rise to a function $f_G : \R^n \to \R^m$, where $n = |\inodes|$ and $m = |\onodes|$.
We would like to construct an abstract transformer
$f^a_G$ that takes an $n$-dimensional polyhedron and outputs an $m$-dimensional polyhedron.

We define $f^a_G(\langle c_{1j}\rangle,\ldots\langle c_{nj} \rangle, \varphi)$ as follows:
\begin{itemize}
\item
First, for every input node $v_i$, we define 
$$\outs^a(v_i) = (\langle c_{ij} \rangle_j, \varphi)$$
Recall that we assume a fixed ordering of 
nodes.
\item
Second, for every non-input node $v$,
we define $$\outs^a(v) = f_v^a\left(p_1,\ldots,p_k, \bigwedge_{i=1}^k \varphi_k\right)$$
where $f^a_v$ is the abstract transformer of $f_v$,
 $v$ has the incoming edges $(v_1,v),\ldots,(v_k,v)$,
 and $$\outs^a(v_i) = (p_i, \varphi_i)$$

Observe what is happening here:
we're combining (with $\land$) the constraints from the incoming edges.
This ensures that we capture the relations between incoming values.

\item Finally, the output of $f^a_G$
is the $m$-dimensional polyhedron $$\left(p_1,\ldots,p_m, \bigwedge_{i=1}^m \varphi_i\right)$$
where $v_1,\ldots,v_m$ are the output nodes
and $\outs^a(v_i) = (p_i, \varphi_i)$
\end{itemize}

Some abstract transformers (for activation functions)
add new generators. 
We can assume that all of these generators are already in the input polyhedron but with coefficients set to 0, and they only get non-zero coefficients 
in the outputs of activation function nodes.


\section*{Looking Ahead}
We looked at the polyhedron abstract domain,
which was first introduced by \citet{cousot1978automatic}.
To minimize the size of the constraints,
\citet{singh2019abstract} use a specialized polyhedron restriction
that limits the number of constraints, and apply it to neural-network verification.
Another representation of polyhedra, with specialized abstract transformers
for convolutional neural networks, is ImageStars~\citep{tran2020verification}.
For a good description of efficient polyhedron domain operations and representations, for general programs, please consult~\citet{singh2017fast}.

    \chapter{Verifying with Abstract Interpretation}\label{ch:absver}

We have seen a number of abstract domains that allow us to evaluate a neural network on an infinite set of inputs.
We will now see how to use this idea for verification of specific properties.
While abstract interpretation can be used, in principle, to verify any
property in our language of correctness properties, much of the work 
in the literature is restricted to specific properties of the form:
\[
\begin{array}{c}
    \pre{\emph{precondition}}\\
    \vec{r} \gets f(\vec{x})\\
    \post{\emph{postcondition}}
\end{array}
\]
where the precondition defines a set of possible values for $\vec{x}$, the inputs of the neural network,
and the postcondition defines a set of possible correct values of $\vec{r}$, the outputs of the neural network.
To verify such properties with abstract interpretation,
we need to perform three tasks:
\begin{enumerate}
    \item Soundly represent the set of values of $\vec{x}$ in the abstract domain.
    \item Abstractly interpret the neural network $f$ on all values of $\vec{x}$, resulting in an overapproximation of values of $\vec{r}$.
    \item Check that all values of $\vec{r}$ satisfy the postcondition.
\end{enumerate}
We've seen how to do (2), abstractly interpreting the neural network.
We will now see how to do (1) and (3) for specific correctness properties from the literature.

\section{Robustness in Image Recognition}
In image recognition, we're often interested in ensuring that 
all images similar to some image $\vec{c}$ have the same prediction as the label of $\vec{c}$.
Let's say that the label of $\vec{c}$ is $y$.
Then we can define robustness  using the following property:

\[
\begin{array}{c}
    \pre{\norm{\vec{x} - \vec{c}}_p \leq \epsilon}\\
    \vec{r} \gets f(\vec{x})\\
    \post{\class(\vec{r}) = y}
\end{array}
\]
where 
$\norm{\vec{x}}_p$ is the $\ell_p$ norm of a vector 
and $\epsilon > 0$.
Typically we use the $\ell_2$ (Euclidean) or the $\ell_\infty$ norm
as the distance metric between two images:
$$\norm{\vec{z}}_2 = \sqrt{\sum_i |z_i|^2}$$
$$\norm{\vec{z}}_\infty = \max_i |z_i|$$

Intuitively, the $\ell_2$ norm is the length of the straight-line  between two images in $\R^n$,
while $\ell_\infty$ is the largest discrepancy between two images.
For example, if each element of an image's vector represents one pixel,
then the $\ell_\infty$ norm tells us the biggest difference between two corresponding pixels.

\begin{example}
    \begin{align*}
        \norm{(1,2) - (2,4)}_2 &= \norm{(-1,-2)}_2\\ & = \sqrt{3}\\
    \norm{(1,2) - (2,4)}_\infty &= \norm{(-1,-2)}_\infty\\ & = 2
    \end{align*}
\end{example}

\begin{example}
Consider an image $\vec{c}$ where every element of $\vec{c}$ represents
the brightness of a grayscale pixel, from black to white, say from 0 to 1.
If we want to represent the set of all images that are like $\vec{c}$
but where each pixel differs by a brightness  amount of 0.2, then we can use the $\ell_\infty$ norm in the precondition, i.e., the set of images $\vec{x}$
where $$\norm{\vec{x} - \vec{c}}_\infty \leq 0.2$$
This is because the $\ell_\infty$ norm captures the \emph{maximum} discrepancy a pixel in $\vec{c}$ can withstand.
As an example, consider the handwritten 7 digit on the left and a version of it on the right
where each pixel's brightness was changed by up to 0.2 randomly:
\begin{figure}[h!]
    \centering
    \begin{minipage}[b]{0.3\textwidth}
      \includegraphics[width=\textwidth]{normal7.pdf}
    \end{minipage}
    ~~~
    \begin{minipage}[b]{0.3\textwidth}
      \includegraphics[width=\textwidth]{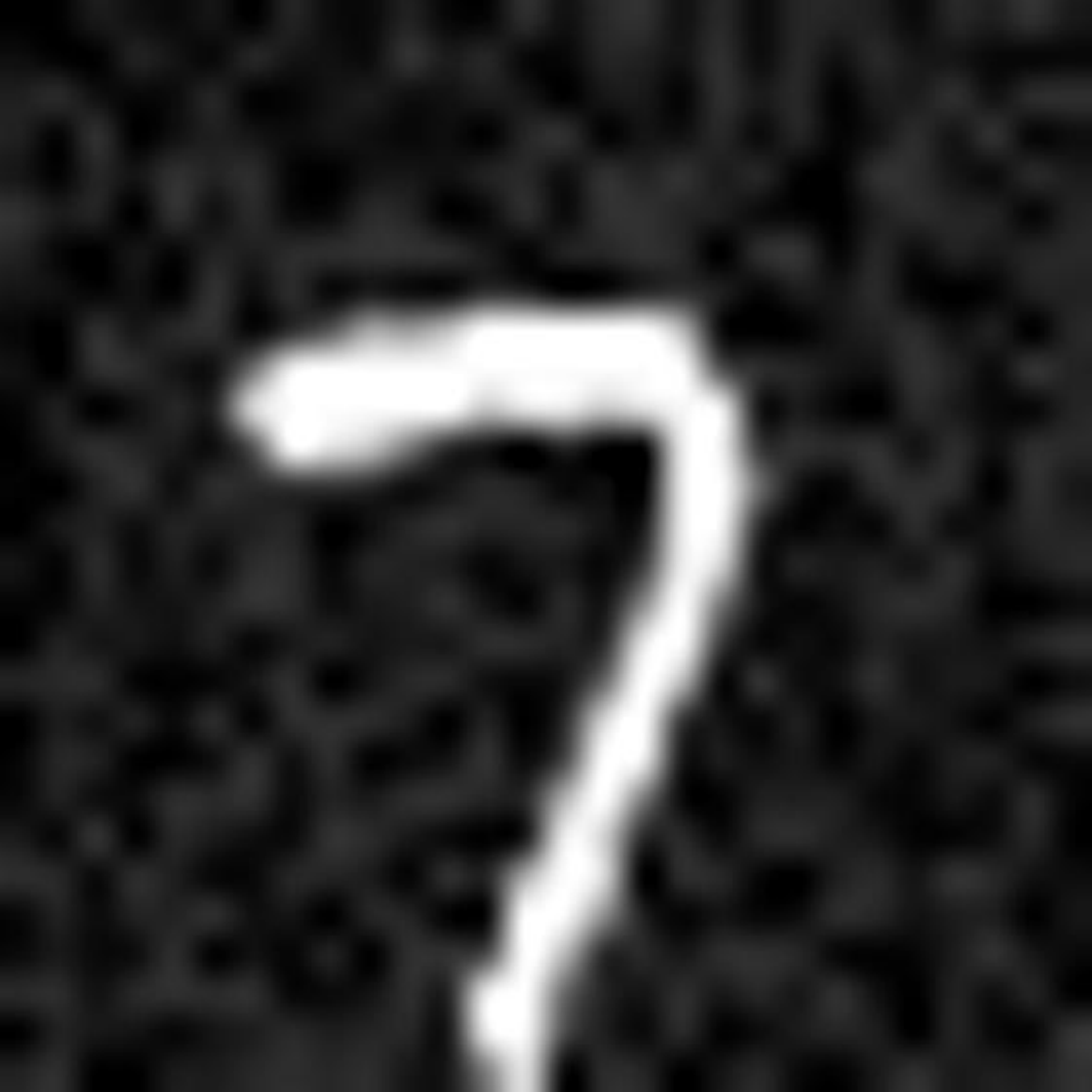}
    \end{minipage}
\end{figure}

Now consider the case where we want to represent all images that are like $\vec{c}$ but where a small region has a very different brightness. For example, 
on the  left we see the handwritten 7 and on the right we see the same handwritten digit but with a small bright dot:
\begin{figure}[h!]
    \centering
    \begin{minipage}[b]{0.3\textwidth}
      \includegraphics[width=\textwidth]{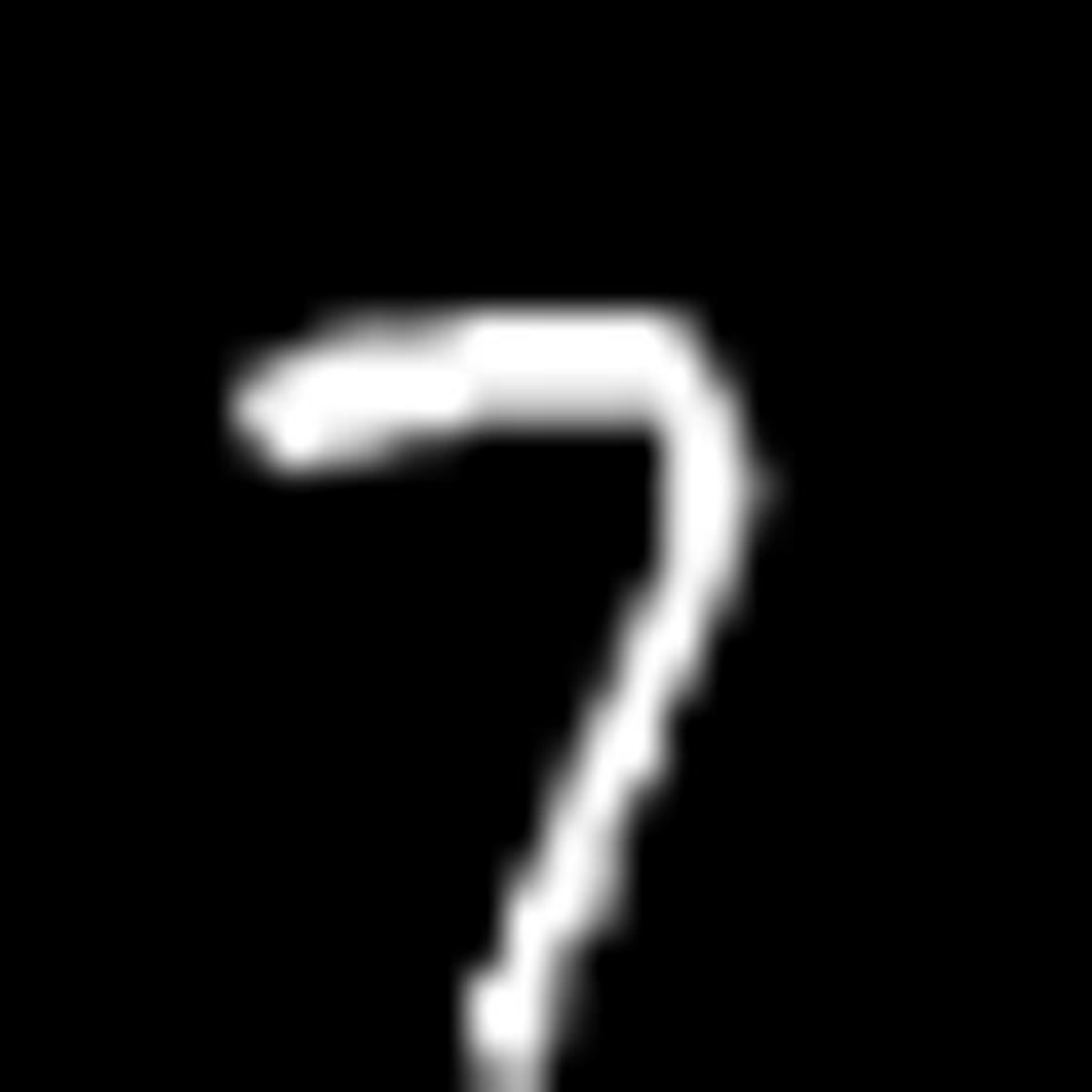}
    \end{minipage}
    ~~~
    \begin{minipage}[b]{0.3\textwidth}
      \includegraphics[width=\textwidth]{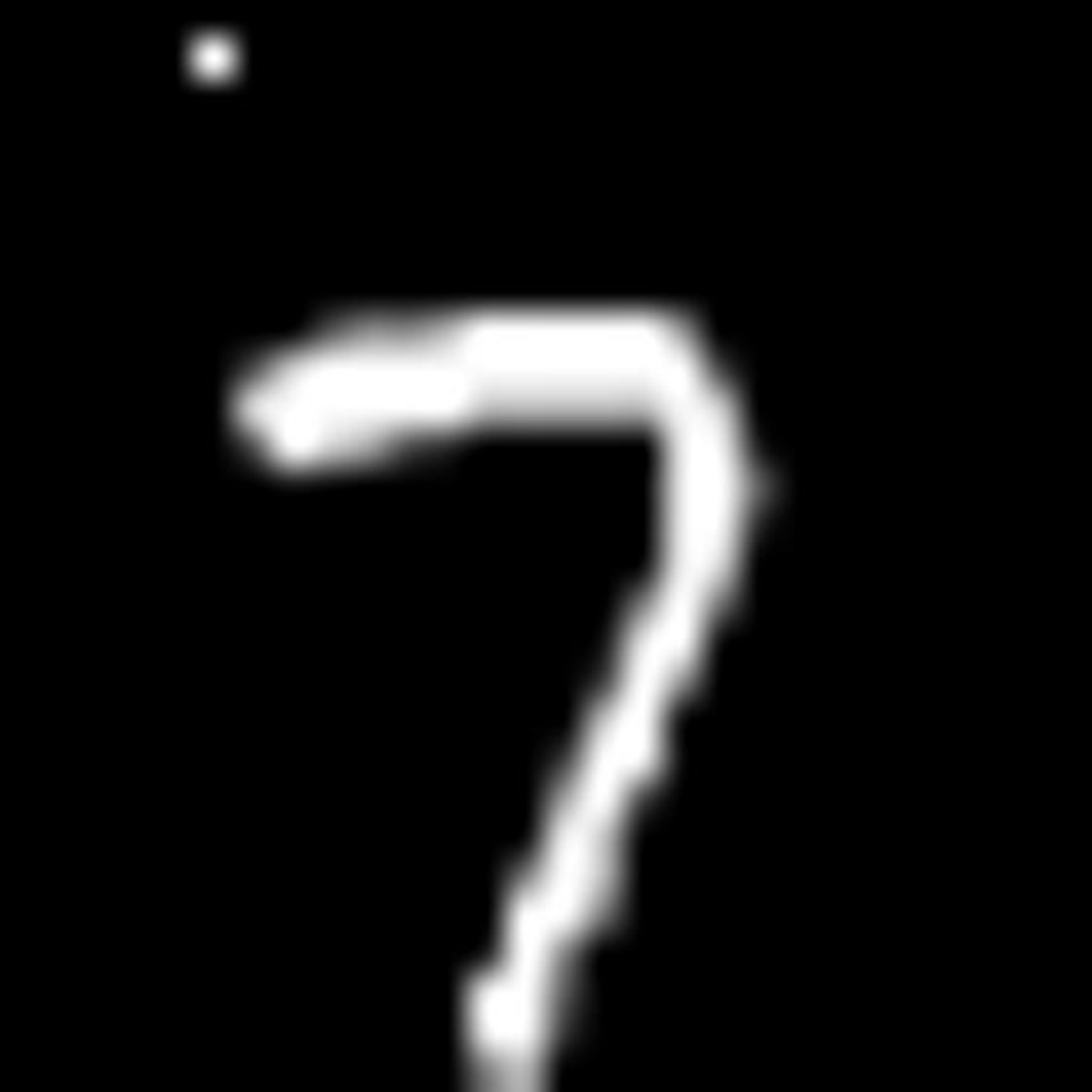}
    \end{minipage}
\end{figure}

To characterize a set of images that have such noise, like the dot above,
we shouldn't use $\ell_\infty$ norm, because $\ell_\infty$ bounds the brightness difference for \emph{all} pixels, but not \emph{some} pixels, and here the brightness difference that results in the white dot is extreme---from 0 (black) to white (1).
Instead, we can use the $\ell_2$ norm.
For the above pair of images,
their $\ell_\infty$-norm distance is 1;
their $\ell_2$-norm distance is also 1,
but the precondition 
$$\{\norm{\vec{x} - \vec{c}}_\infty \leq 1\}$$
includes the images that are all black or all white, which are clearly not the digit 7.
The precondition
$$\{\norm{\vec{x} - \vec{c}}_2 \leq 1\}$$
on the other hand, 
only allows a small number of pixels to significantly differ in brightness.
\end{example}

For verification, 
we will start by focusing on the $\ell_\infty$-norm case
and the interval domain.

\subsection*{Abstracting the Precondition}
Our first goal is to represent the precondition in 
the interval domain.
The precondition is the set of the following images:
$$\{\vec{x} \mid  \norm{\vec{x} - \vec{c}}_\infty \leq \epsilon\}$$

\begin{example}
    Say $\vec{c} = (0,0)$ and $\epsilon = 1$.
    Then the  above set is the following region:
    \begin{center}
        \centering
    \begin{tikzpicture}
        \begin{axis}[
            axis equal image, 
            axis lines=middle,
            xmax=2,
            xmin=-2,
            ymin=-2,
            ymax=2,
            xlabel={},
            ylabel={},
            ytick={-1,0,1},
            xtick ={0,1},width=5cm,
        ]
        \draw [fill=black, opacity=0.2] (axis cs:-1,-1) -- (axis cs:-1,1)
        --(axis cs:1,1) -- (axis cs:1,-1) -- (axis cs:-1,-1);
        \end{axis}
    \end{tikzpicture}
    \end{center}
\end{example}

As the illustration above hints,
it turns out that we can represent the set $\{\vec{x} \mid  \norm{\vec{x} - \vec{c}}_\infty \leq \epsilon\}$
precisely in the interval domain
as 
$$I = ([c_1 - \epsilon,c_1 + \epsilon],\ldots,[c_n - \epsilon,c_n + \epsilon])$$
Informally, this is because the $\ell_\infty$ norm allows us to take any element of $\vec{c}$
and change it by $\epsilon$ independently of other dimensions.

\subsection*{Checking the Postcondition}
Now that we have represented the set of values
that $\vec{x}$ can take in the interval domain as $I$,
we can go ahead and evaluate the abstract transformer
$f^a(I)$,
resulting in an output of the form
$$I' = ([l_1,u_1],\ldots,[l_m,u_m])$$
representing all possible values of $\vec{r}$,
and potentially more.

The postcondition specifies that $\class(\vec{r})=y$.
Recall that $\class(\vec{r})$ is the index of the largest element of $\vec{r}$.
To prove the property, we have to show that for all 
$\vec{r} \in I'$, $\class(\vec{r})=y$.
We make the observation that 
\begin{align*}
&\text{if } l_y > u_i \text{ for all } i \neq y,\\
&\text{then for all } \vec{r} \in I', \class(\vec{r})=y
\end{align*}
In other words, if the $y$th interval is larger than all others,
then we know that the classification is always $y$.
Notice that this is a one-sided check:
if $l_y \leq u_i$ for some $i \neq y$,
then we can't disprove the property.
This is because the set $I'$ overapproximates
the set of possible predictions of the neural network on
the precondition. So $I'$ may include spurious predictions.

\begin{example}
    Suppose that $$f^a(I) = I' = ([0.1,0.2], [0.3,0.4])$$
    Then, $\class(\vec{r}) = 2$ for all $\vec{r} \in I'$.
    This is because the second interval is strictly larger than the 
    first interval.

    Now suppose that $$I' = ([0.1,0.2], [0.15,0.4])$$
    These two intervals overlap in the region $0.15$ to $0.2$.
    This means that we cannot conclusively say that 
    $\class(\vec{r}) = 2$ for all $\vec{r} \in I'$,
    and so verification fails.
\end{example}

\subsection*{Verifying Robustness with Zonotopes}
Let's think of how to check the $\ell_\infty$-robustness property
using the zonotope domain.
Since the precondition is a hyperrectangular set,
we can precisely represent it as a zonotope $Z$.
Then, we evaluate the abstract transformer $f^a(Z)$,
resulting in a zonotope $Z'$.

The fun bit is checking the postcondition.
We want to make sure that dimension $y$ 
is greater than all others.
The problem boils down to checking if a 1-dimensional zonotope
is always $> 0$.
Consider the zonotope 
$$Z' = (\langle c_{1i}\rangle,\ldots\langle c_{mi} \rangle)$$
To check that dimension $y$ is greater than dimension $j$,
we check if the lower bound of the 1-dimensional zonotope $$\langle c_{yi}\rangle - \langle c_{ji}\rangle$$
is $>0$.

\begin{example}
    Suppose that 
    \[Z' = (2 + \epsilon_1,\ 4 + \epsilon_1 + \epsilon_2)\]
    which is visualized as follows, with the center point (2,4) in red:

    \begin{center}
        \begin{tikzpicture}
             \begin{axis}[
                 axis lines=middle,
                 xmax=4,
                 xmin=0,
                 ymin=0,
                 ymax=6.2,
                 xlabel={},
                 ylabel={},
                 ytick={1,2,3,4,5},
                 xtick ={1,2,3,4,5},width=5cm]
                 \draw [fill=black, opacity=0.2] (axis cs:1,2) -- (axis cs:1,4) -- (axis cs:3,6) -- (axis cs:3,4) -- (axis cs:1,2);  
                 \addplot[Maroon,mark=*] coordinates {(2,4)} node [pos=0,left] {}     ;    
        \end{axis}
        \end{tikzpicture}
    \end{center}
    Clearly, for any point $(x,y)$ in this region,
    we have $y > x$.
    To check that $y>x$ mechanically,
    we subtract the $x$ dimension from the $y$ dimension:
    $$ (4 + \epsilon_1 + \epsilon_2) - (2 + \epsilon_1) = 2 + \epsilon_2$$
    The resulting 1-dimensional zonotope ($2+\epsilon_2$) denotes the interval $[1,3]$,
    which is greater than zero.
\end{example}

\subsection*{Verifying Robustness with Polyhedra}
With the polyhedron domain, the story is analogous to zonotopes
but requires invoking a linear-program solver.
We represent the precondition as a hyperrectangular polyhedron
$Y$.
Then, we evaluate the abstract transformer, $f^a(Y)$,
resulting in the polyhedron
$$Y' = (\langle c_{1i}\rangle,\ldots\langle c_{mi} \rangle, \varphi)$$
To check if dimension $y$ is greater than dimension $j$,
we ask a linear-program solver 
if the following constraints are satisfiable
$$F \land \langle c_{yi}\rangle > \langle c_{ji}\rangle$$

\subsection*{Robustness in $\ell_2$ Norm}
Let's now consider the precondition 
with the set of images within an $\ell_2$ norm of $\vec{c}$:
$$\{\vec{x} \mid  \norm{\vec{x} - \vec{c}}_2 \leq \epsilon\}$$

\begin{example}
    Say $\vec{c} = (0,0)$ and $\epsilon = 1$.
    Then the  above set is the following circular region:
    \begin{center}
        \centering
    \begin{tikzpicture}
        \begin{axis}[
            axis equal image, 
            axis lines=middle,
            xmax=2,
            xmin=-2,
            ymin=-2,
            ymax=2,
            xlabel={},
            ylabel={},
            ytick={-1,0,1},
            xtick ={0,1},width=5cm,
        ]
        \addplot [domain=-180:180, samples=100, fill=black, opacity=.2] ({cos(x)},{sin(x)});
    \end{axis}
    \end{tikzpicture}
    \end{center}
\end{example}

This set cannot be represented precisely in the interval domain. 
To ensure that we can verify the property, we need to overapproximate the circle with a box.
The best we can do is using the tightest box around the circle, i.e., $([-1,1],[-1,1])$, shown below in red:
\begin{center}
    \centering
\begin{tikzpicture}
    \begin{axis}[
        axis equal image, 
        axis lines=middle,
        xmax=2,
        xmin=-2,
        ymin=-2,
        ymax=2,
        xlabel={},
        ylabel={},
        ytick={-1,0,1},
        xtick ={0,1},width=5cm,
    ]
    \addplot [domain=-180:180, samples=100, fill=black, opacity=.2] ({cos(x)},{sin(x)});
    \draw [fill=Maroon, opacity=0.2] (axis cs:-1,-1) -- (axis cs:-1,1)
    --(axis cs:1,1) -- (axis cs:1,-1) -- (axis cs:-1,-1);
\end{axis}
\end{tikzpicture}
\end{center}

\newcommand{\polygon}[2]{%
  let \n{len} = {2*#2*tan(360/(2*#1))} in
 ++(0,-#2) ++(\n{len}/2,0) \foreach \x in {1,...,#1} { -- ++(\x*360/#1:\n{len})}}

The zonotope and polyhedron domains also cannot represent the circular set precisely.
However, there isn't a tightest zonotope or polyhedron
that overapproximates the circle.
For example, with polyhedra, one can keep adding more and more faces, getting a better and better approximation, as illustrated below:
\begin{center}
\begin{tikzpicture}
    \draw[fill=black,opacity=0.2] (0,0) circle (1);
    \draw (0,0) \polygon{5}{1};
\end{tikzpicture}
\begin{tikzpicture}
    \draw[fill=black,opacity=0.2] (0,0) circle (1);
    \draw (0,0) \polygon{6}{1};
\end{tikzpicture}
\begin{tikzpicture}
    \draw[fill=black,opacity=0.2] (0,0) circle (1);
    \draw (0,0) \polygon{7}{1};
\end{tikzpicture}
\begin{tikzpicture}
    \draw[fill=black,opacity=0.2] (0,0) circle (1);
    \draw (0,0) \polygon{8}{1};
\end{tikzpicture}
\end{center}

In practice, there is, of course, a precision--scalability tradeoff: more faces mean more complex constraints and therefore slower verification.

\section{Robustness in Natural-Language Processing}
We will now take a look at another robustness
property from natural-language processing.
The goal is to show that replacing words with synonyms
does not change the prediction of the neural network.
For instance, a common task is sentiment analysis, where the neural network predicts whether, say, a movie review is positive or negative. Replacing ``amazing'' with ``outstanding'' should not fool the neural network into thinking a positive review is a negative one.

We assume that the input to the neural network is a vector where  element $i$ is a numerical representation of the $i$th word in the sentence, and that each word $w$ has a finite set of possible synonyms $S_w$, where we assume $w \in S_w$.
Just as with images, we assume a fixed sentence $\vec{c}$
with label $y$
for which we want to show robustness.
We therefore define the correctness property as follows:
\[
    \begin{array}{c}
        \pre{x_i \in S_{c_i} \text{ for all } i}\\
        \vec{r} \gets f(\vec{x})\\
        \post{\class(\vec{r}) = y}
    \end{array}
\]
Intuitively, the precondition defines
all vectors $\vec{x}$ that are like $\vec{c}$ but 
where some words are replaced by synonyms.

The set of possible vectors $\vec{x}$
is finite, but it is exponentially large
in the length of the input sentence.
So it is not wise to verify the property by evaluating the neural network on every possible $\vec{x}$.
We can, however, represent an overapproximation 
of the set of possible sentences in the interval domain. The idea is to take interval between the largest and smallest possible numerical representations of the synonyms of every word, as follows:
$$([\min S_{c_1}, \max S_{c_1}], \ldots,[\min S_{c_n}, \max S_{c_n}] )$$
This set contains all the values of $\vec{x}$, and more, but it is easy to construct, since we need only go through every set of synonyms $S_{c_i}$ individually,
avoiding an exponential explosion.

The rest of the verification process follows that of image robustness. In practice, similar words tend to have close numerical representations, thanks to the power of \emph{word embeddings}~\citep{mikolov2013efficient}. This ensures that the interval is pretty tight.
If words received arbitrary numerical representations, then our abstraction can be arbitrarily bad.

\section*{Looking Ahead}
We saw examples of how to verify properties
via abstract interpretation.
The annoying thing is that for every abstract domain
and every property of interest, we may need custom operations.
Most works that use abstract interpretation so far have focused on the properties I covered in this chapter.
Other properties from earlier in the book can also be verified via the numerical domains we've seen.
For example, the aircraft controller from \cref{ch:correctness}
has properties of the form:
\[
\begin{array}{c}
    \pre{d \geq 55947, \ v_\emph{own} \geq 1145 , \ v_\emph{int} \leq 60}\\
    \vec{r} \gets f(d,v_\emph{own},v_\emph{int}, \ldots )\\
    \post{\text{score of nothing in } \vec{r} \text{ is below 1500}}
\end{array}
\]
Note that the precondition can be captured precisely in the interval domain.

At the time of writing, abstraction-based techniques have been applied successfully to relatively large neural networks, with up to a million neurons along more than thirty layers~\citep{DBLP:journals/corr/abs-2007-10868,DBLP:conf/cav/TranBXJ20}.
Achieving such results requires performant implementations, particularly for more complicated domains like the zonotope and polyhedron domain.
For instance, \citet{DBLP:journals/corr/abs-2007-10868} come up with data-parallel implementations of polyhedron abstract transformers that run on a \textsc{gpu}.
Further, there are heuristics that can be employed to minimize the number of generators in the zonotope domain---limiting the number of generators reduces precision while improving efficiency.
It is also important to note that thus far most of the action in the abstraction-based verification space, and verification of neural networks at large, has been focused on $\ell_p$-robustness properties for images.
(We're also starting to see evidence that the ideas can apply to natural-language robustness~\citep{zhang21}.)
So it's unclear whether verification will work for more complex perceptual notions of robustness---e.g., rotating an image or changing the virtual background on a video---or other more complex properties and domains, e.g., malware detection. 

The robustness properties we discussed check if a fixed region of inputs surrounding a point lead to the same prediction.
Alternatively, we can ask, \emph{how big is the region around a point that leads to the same prediction?} Na\"ively, we can do this by repeatedly performing verification with larger and larger $\ell_2$ or $\ell_\infty$ bounds until verification fails.
Some techniques exploit the geometric structure of a neural network---induced by ReLUs---to grow a robust region around a point~\citep{DBLP:conf/nips/ZhangSS18,DBLP:conf/iclr/FromherzLFPP21}.

As we discussed throughout this part of the book, abstract-interpretation techniques can make stupid mistakes
due to severe overapproximations.
However, abstract interpretation works well in practice for verification. Why? 
Two recent papers shed light on this question from a theoretical perspective~\citep{DBLP:conf/iclr/BaaderMV20,DBLP:journals/corr/abs-2007-06093}.
The papers generalize the \emph{universal approximation} property  of neural networks (\cref{sec:blocks}) to verification with the interval domain or any domain that is more precise.
Specifically, imagine that we have a neural network that is robust
as per the $\ell_\infty$ norm; i.e.,
the following property is true for a bunch of inputs of interest:
\[
\begin{array}{c}
    \pre{\norm{\vec{x} - \vec{c}}_\infty \leq \epsilon}\\
    \vec{r} \gets f(\vec{x})\\
    \post{\class(\vec{r}) = y}
\end{array}
\]
But suppose that abstract interpretation using the interval domain fails to prove robustness for most (or all) inputs of interest.
It turns out that we can always construct a neural network $f'$,
using any realistic activation function (ReLU, sigmoid, etc.),
that is very similar to $f$---as similar as we like---and for which we can prove robustness using abstract interpretation.
The bad news, as per \cite{DBLP:journals/corr/abs-2007-06093},
is that the construction of $f'$ is likely exponential in the size of the domain.


    \chapter{Abstract Training of Neural Networks}
\label{ch:absintnn}

You have reached the final chapter of this glorious journey.
So far on our journey, we have assumed that we're given a neural network
that we want to verify.
These neural networks are, almost always, constructed by learning from data.
In this chapter, we will see how to train a neural network 
that is more amenable to verification via abstract interpretation
for a property of interest.

\section{Training Neural Networks}
We begin by describing neural network training from a data set.
Specifically, we will focus throughout this chapter on a classification setting.

\subsection*{Optimization Objective}
A dataset is of the form $$\{(\vec{x}_1,y_1), \ldots, (\vec{x}_m,y_m)\}$$
where each $\vec{x}_i \in \R^n$ is an \emph{input} to the neural network, e.g., an image or a sentence, and $y_i \in \{0,1\}$ is a binary \emph{label}, 
e.g., indicating if a given image is that of a cat or if a sentence has a positive or negative sentiment.
Each item in the dataset is typically assumed to be sampled independently from a probability distribution, e.g., the distribution of all images of animals.

Given a dataset, we would like to construct a function in $\R^n \to \R$
that makes the right prediction on most of the points in the dataset.
Specifically, we assume that we have a family of functions represented as a parameterized function $f_\theta$, where $\theta$ is a vector of \emph{weights}.
We would like to find the best function by searching the space of $\theta$ values.
For example, we can have the family of affine functions 
$$f_\theta(\vec{x}) = \theta_1 + \theta_2 x_1 + \theta_3 x_2$$  

To find the best function in the function family, we
effectively need to solve an optimization problem like this one:

$$\argmin_\theta \frac{1}{m} \sum_{i=1}^m \mathds{1}[f_\theta(\vec{x}_i) = y_i]$$
where $\mathds{1}[b]$ is 1 if $b$ is  \emph{true} and 0 otherwise.
Intuitively, we want the function that makes the smallest number of prediction mistakes on our dataset $\{(\vec{x}_1,y_1), \ldots, (\vec{x}_m,y_m)\}$.

Practically, this optimization objective is quite challenging to solve, since the objective is non-differentiable---because of the Boolean $\mathds{1}[\cdot]$ operation, which isn't smooth.
Instead, we often solve a relaxed optimization objective like \emph{mean squared error} (\mse), which minimizes \emph{how far} $f_\theta$'s prediction is from each $y_i$. \mse looks like this:
$$\argmin_\theta \frac{1}{m} \sum_{i=1}^m (f_\theta(\vec{x}_i) - y_i)^2$$
Once we've figured out the best values of $\theta$,
we can predict the label of an input $\vec{x}$ by computing $f_\theta(\vec{x})$
and declaring label 1 iff $f_\theta(\vec{x}) \geq 0.5$.

We typically use a general form to describe the optimization objective.
We assume that we're given a \emph{loss} function $L(\theta, \vec{x},y)$ which
measures how \emph{bad} is the prediction $f_\theta(\vec{x})$ is compared to the label $y$.
Formally, we solve 
\begin{align}\label{eq:loss}
    \argmin_\theta \frac{1}{m} \sum_{i=1}^m L(\theta, \vec{x}_i,y_i)
\end{align}

Squared error is one example loss function, but there are others, like \emph{cross-entropy loss.} For our purposes here, we're not interested in what loss function is used.

\subsection*{Loss Function as a Neural Network}
The family of functions $f_\theta$ is represented as a neural network
graph $G_\theta$, where every node $\node$'s function $f_\node$ may be parameterized by $\theta$.
It turns out that we can we represent the loss function $L$ also as a neural network; specifically, we represent $L$ as an extension of the graph $G_\theta$
by adding a node at the very end that computes, for example, the squared difference between $f_\theta(\vec{x})$ and $y$.
By viewing the loss function $L$ as a neural network,
we can abstractly interpret it, as we shall see later in the chapter.

Suppose that $f_\theta : \R^n \to \R$ has a graph of the form
\begin{center}
    \begin{tikzpicture}
        \draw node at (0, 0) [input] (in) {$\node_1$};
        \draw node at (0, -2) [input] (in2) {$\node_n$};
        \draw node at (0, -1) [empty] (mid) {$\vdots$};

        \draw node at (3, -1) [output] (out) {$\node_o$};

        \draw[->,dashed,thick] (in) -- (out);
        \draw[->,dashed,thick] (in2) -- (out);
    \end{tikzpicture}
\end{center}
where the dotted arrows indicate potentially intermediate nodes.
We can construct the graph of a loss function $L(\theta,\vec{x},y)$ by adding an input node $\node_y$ for the label $y$ and creating a new output node $\node_L$ that  compares the output of $f_\theta$ (the node $\node_o$) with $y$.
\begin{center}
    \begin{tikzpicture}
        \draw node at (0, 0) [input] (in) {$\node_1$};
        \draw node at (0, -2) [input] (in2) {$\node_n$};
        \draw node at (0, -3) [input] (in3) {$\node_y$};
        \draw node at (0, -1) [empty] (mid) {$\vdots$};

        \draw node at (3, -1) [oper] (op) {$\node_o$};
        \draw node at (5, -1) [output] (out) {$\node_L$};

        \draw[->,dashed,thick] (in) -- (op);
        \draw[->,dashed,thick] (in2) -- (op);
        \draw[->,thick] (op) -- (out);
        \draw[->,thick] (in3) -- (out);
    \end{tikzpicture}
\end{center}
Here, input node $\node_y$ takes in the label $y$
and
$f_{\node_L}$ encodes the loss function, e.g.,
mean squared error $(f(\vec{x}) - y)^2$.

\subsection*{Gradient Descent}
How do we find values of $\theta$ that minimize the loss?
Generally, this is a hard problem, so we just settle for a good enough set of values.
The simplest thing to do is to randomly sample different values of $\theta$ and return the best one after some number of samples.
But this is a severely inefficient approach.

Typically, neural-network training employs a form of \emph{gradient descent}.
Gradient descent is a very old algorithm, due to Cauchy in the mid 1800s.
It works by starting with a random value of $\theta$ and iteratively
nudging it towards better values by following the gradient of the optimization objective.
The idea is that starting from some point $x_0$, if we want to minimize $g(x_0)$,
then our best bet is to move in the direction of the negative gradient at $x_0$.

The gradient of a function $g(\theta)$ with respect to inputs $\theta$,
denoted $\nabla g$,
is the vector of partial derivatives\footnote{The gradient is typically a column vector, but for simplicity of presentation we treat it as a row vector here.} $$\left(\frac{\partial g}{\partial \theta_1}, \ldots, \frac{\partial g}{\partial \theta_n}\right)$$
The gradient at a specific value $\theta^0$,
denoted $(\nabla g)(\theta^0)$, is  
$$\left(\frac{\partial g}{\partial \theta_1}(\theta^0), \ldots, \frac{\partial g}{\partial \theta_n}(\theta^0)\right)$$
If you haven't played with partial derivatives in a while, I recommend \citet{deisenroth2020mathematics} for a machine-learning-specific refresher.

Gradient descent can be stated as follows:
\begin{enumerate}
    \item Start with $j = 0$ and a random value of $\theta$, called $\theta^0$.
    \item Set $\theta^{j+1}$ to $\theta^j - \eta ((\nabla g)(\theta^i))$.
    \item Set $j$ to $j+1$ and repeat.
\end{enumerate}
Here $\eta > 0$ is the \emph{learning rate}, which constrains the size of
the change of $\theta$: too small a value and you'll make baby steps towards a good solution; too large a value and you'll bounce wildly around unable to catch a good region of solutions for $\theta$, potentially even diverging. The choice of $\eta$ is typically determined empirically by monitoring the progress of the algorithm for a few iterations.
The algorithm is usually terminated when the loss has been sufficiently minimized
or when it starts making tiny steps, asymptotically converging to a solution.

In our setting, our optimization objective is 
$$\frac{1}{m} \sum_{i=1}^m L(\theta, \vec{x}_i,y_i)$$
Following the beautiful properties of derivatives, 
the gradient of this function is 
$$\frac{1}{m} \sum_{i=1}^m \nabla  L(\theta, \vec{x}_i,y_i)$$
It follows that the second step of gradient descent can be rewritten as 
\begin{center}
    Set $\theta^{j+1}$ to $\theta^j - \frac{\eta}{m} \sum_{i=1}^m\nabla  L(\theta^j, \vec{x}_i,y_i)$.
\end{center}
In other words, we compute the gradient for every point in the dataset independently and take the average.

\subsection*{Stochastic Gradient Descent}
In practice, gradient descent is incredibly slow.
So people typically use \emph{stochastic gradient descent} (\sgd).
The idea is that, instead of computing the average gradient in every iteration for the entire dataset,
we use a random subset of the dataset to \emph{approximate} the gradient.
\sgd is also known as \emph{mini-batch gradient descent}.
Specifically, here's how \sgd looks:

\begin{enumerate}
    \item Start with $j = 0$ and a random value of $\theta$, called $\theta^0$.
    \item Divide the dataset into a random set of $k$ \emph{batches}, $B_1,\ldots,B_k$.
    \item For $i$ from $1$ to $k$,  
        \begin{align*}
        \text{Set } & \theta^{j+1} \text{ to  } \theta^j -  \frac{\eta}{m} \sum_{(\vec{x},y) \in B_i}\nabla  L(\theta^j, \vec{x},y)\\
        \text{Set } & j \text{ to } j+1 
        \end{align*}
    \item Go to step 2.
\end{enumerate}
In practice, the number of batches $k$ (equivalently size of the batch) is typically a function of how much data you can cram into the \textsc{gpu} at any one point.\footnote{To readers from the future: In the year 2021, graphics cards and some specialized accelerators used
to be the best thing around for matrix multiplication. What have you folks settled on, quantum or \textsc{dna} computers?}

\section{Adversarial Training with Abstraction}
The standard optimization objective for minimizing 
loss (\cref{eq:loss}) is only interested in, well, minimizing the average loss for the dataset, i.e., getting as many predictions right.
So there is no explicit goal of generating robust neural networks, for any definition of robustness.
As expected, this translates to  neural networks
that are generally not very robust to perturbations in the input.
Furthermore, even if the trained network is robust on some inputs,
verification with abstract interpretation often fails to produce a proof.
This is due to the overapproximate nature of abstract interpretation. One can always rewrite a neural network---or any program for that matter---into one that fools abstract interpretation, causing it to loose a lot of precision and therefore fail to verify properties of interest.
Therefore, we'd like to train neural networks that are \emph{friendly} for abstract interpretation.

We will now see how to change the optimization objective to produce robust networks and how to use abstract interpretation within \sgd to solve this optimization objective.

\subsection*{Robust Optimization Objective}
Let's consider the image-recognition-robustness property from the previous
chapter:
For every $(\vec{x},y)$ in our dataset,
we want the neural network to predict $y$
on all images $\vec{z}$ such that $\norm{\vec{x}-\vec{z}}_\infty \leq \epsilon$.
We can characterize this set as
$$R(\vec{x}) = 
\{\vec{z} \mid \norm{\vec{x}-\vec{z}}_\infty \leq \epsilon\}$$

Using this set, we will rewrite our optimization objective as follows:
\begin{align}\label{eq:robloss}
    \argmin_\theta \frac{1}{m} \sum_{i=1}^m\ \max_{\vec{z} \in R(\vec{x}_i)} L(\theta,\vec{z},y_i)
\end{align}
Intuitively, instead of minimizing the loss for $(\vec{x}_i,y_i)$, we minimize the loss for the worst-case perturbation of $\vec{x}_i$ from the set $R(\vec{x}_i)$.
This is known as a \emph{robust-optimization} problem~\citep{ben2009robust}.
Training the neural network using such objective is known as adversarial training---think of an adversary (represented using the  \emph{max})  that's always trying to mess with your dataset to maximize the loss as you are performing the training~\citep{madry2017towards}.

\subsection*{Solving Robust Optimization via Abstract Interpretation}
We will now see how to solve the robust-optimization problem
using \sgd and abstract interpretation!

Let's use the interval domain.
The set $R(\vec{x})$ can be defined in the interval domain precisely, as we saw in the last chapter, since it defines a set of images within an $\ell_\infty$-norm bound.  
Therefore, we can overapproximate the inner maximization 
by abstractly interpreting $L$ on the entire set $R(\vec{x}_i)$.
(Remember that $L$, as far as we're concerned, is just a neural network.)
Specifically, by virtue of soundness of the abstract transformer $L^a$, we know that 
 $$\left(\max_{\vec{z} \in R(\vec{x}_i)} L(\theta,\vec{z},y_i)\right) \leq u $$
where  $$L^a(\theta, R(\vec{x}_i), y_i) = [l,u]$$
In other words, we can overapproximate the inner maximization
by abstractly interpreting the loss function on the set  $R(\vec{x}_i)$
and taking the upper bound.

We can now rewrite our robust-optimization objective as follows:
\begin{align}\label{eq:absloss}
    \argmin_\theta \frac{1}{m} \sum_{i=1}^m\ 
    \text{upper bound of } L^a(\theta,R(\vec{x}_i),y_i)
\end{align}
Instead of thinking of $L^a$ as an abstract transformer in the interval domain,
we can think of it as a function 
that takes a vector of inputs, denoting lower and upper bounds of $R(\vec{x})$, and returns the pair of lower and upper bounds.
We call this idea \emph{flattening} the abstract transformer; we illustrate flattening with a simple example:
\begin{example}
    Consider the ReLU function $\relu(x) = \max(0,x)$.
    The interval abstract transformer is 
    $$\relu^a([l,u]) = [\max(0,l), \max(0,u)]$$
    We can flatten it into a function $\relu^{\emph{af}} : \R^2 \to \R^2$
    as follows:
    $$\relu^{\emph{af}}(l,u) = (\max(0,l), \max(0,u))$$
    Notice that $\relu^\emph{af}$ returns a pair in $\R^2$ as opposed to an interval.
\end{example}

With this insight, we can flatten the abstract loss function $L^a$ into
$L^\emph{af}$.
Then, we just invoke \sgd on the following optimization problem,

\begin{align}
    \argmin_\theta \frac{1}{m} \sum_{i=1}^m\ 
     L^{\emph{af}}_u(\theta,l_{i1},u_{i1},\ldots,l_{in},u_{in},y_i)
\end{align}
where $L_u^\emph{af}$ is only the upper bound of the output of $L^\emph{af}$,
i.e., we throw away the lower bound (remember \cref{eq:absloss}),
and $R(\vec{x}_i) = ([l_{i1},u_{i1}],\ldots,[l_{in},u_{in}])$.

\sgd can optimize such objective 
because all of the abstract transformers 
of the interval domain that are of interest for neural networks are differentiable (almost everywhere).
The same idea can be adapted to the zonotope domain, but it's a tad bit uglier.
\begin{example}
    Given a function $f: \R \to \R$,
     its zonotope abstract transformer
    $f^a$ is one that takes as input a 1-dimensional input zonotope with $m$ generator variables,
    $\langle c_0, \ldots, c_m\rangle$, and outputs
    a 1-dimensional zonotope also with $m$ generators, $\langle c_0', \ldots, c_m'\rangle$.
    We can flatten 
    $f^a$ by treating it as a function in $$f^\emph{af}: \R^{m+1} \to \R^{m+1}$$
    where the $m+1$ arguments and outputs are the coefficients of the $m$ generator variables and the center point.   
\end{example}

Flattening does not work for the polyhedron domain, because it invokes a black-box linear-programming solver for activation functions, which is not differentiable.

\section*{Looking Ahead}
We saw how to use abstract interpretation to train (empirically) more robust neural networks.
It has been shown that neural networks
trained with abstract interpretation tend to be 
(1) more robust to perturbation attacks
and (2) are verifiably robust using abstract interpretation.
The second point is subtle: 
You could have a neural network that satisfies a correctness property of interest, but that does not mean that an abstract domain will succeed at verifying that the neural network
satisfies the property.
By incorporating abstract interpretation into training,
we guide \sgd towards neural networks that are amenable to verification.

The first use of abstract interpretation within the training loop
came in 2018~\citep{mirman2018differentiable,gowal2018effectiveness}.
Since then, many approaches have used abstract interpretation
to train robust image-recognition as well as natural-language-processing models~\citep{zhang20,zhang21,jia2019certified,NEURIPS2020_0cbc5671,huang2019achieving}.
Robust optimization is a rich field~\citep{ben2009robust};
to my knowledge, \citet{madry2017towards} were the first
to pose training of $\ell_p$-robust neural networks as a robust-optimization problem.

There are numerous techniques for producing neural networks 
that are amenable to verification.
For instance, \cite{DBLP:conf/nips/SivaramanFMB20} use constraint-based verification to verify that a neural network is monotone.
Since constraint-based techniques can be complete, they can produce counterexamples, which are then used to retrain the neural network, steering it towards monotonicity.
Another interesting direction in the constraint-based world
is to train neural networks towards ReLUs whose inputs are always positive
or always negative~\citep{xiao2018training}. This ensures that the generated constraints have as few disjunctions as possible, because the encoding of the ReLU will be linear (i.e., no disjunction).

$\ell_p$-robustness properties are closely related to the notion of  \emph{Lipschitz continuity}.
For instance, a network $f : \R^n \to \R^m$ is $K$-Lipschitz under the $\ell_2$
norm if
$$\norm{f(\vec{x}) - f(\vec{y})}_2 \leq K\norm{\vec{x}-\vec{y}}_2$$ 
The smallest $K$ satisfying the above is called the Lipschitz constant of $f$.
If we can bound $K$, then we can prove $\ell_2$-robustness of $f$.
A number of works aim to construct networks with constrained Lipschitz constants, e.g., by adding special layers to the network architecture or modifying the training procedure~\citep{DBLP:conf/iclr/TrockmanK21,DBLP:conf/icml/LeinoWF21,DBLP:conf/nips/LiHALGJ19} 

    \chapter{The Challenges Ahead}
\label{ch:ahead}

My goal with this book is to give an introduction to two salient neural-network verification approaches. 
But, as you may expect, there are many interesting ideas, issues, and prospects that we did not discuss.

\subsection*{Correctness Properties}
In Part I of the book, we saw a general language of correctness properties,
and saw a number of interesting examples across many domains.
One of the hardest problems in the field verification---and the one that is least discussed---is how to actually come up with such properties (also known as \emph{specifications}).
For instance, we saw forms of the robustness property many times throughout the book. 
Robustness, at a high level, is very desirable. You expect an \emph{intelligent}
system to be robust in the face of silly transformations to its input.
But how exactly do we define robustness? Much of the literature focuses on $\ell_p$ norms, which we saw in \cref{ch:absver}.
But one can easily perform transformations that lie outside $\ell_p$ norms,
e.g., rotations to an image, or work in domains where $\ell_p$ norms don't make much sense, e.g., natural language, source code, or other structured data.

Therefore, coming up with the right properties to verify and enforce is a challenging, domain-dependent problem requiring a lot of careful thought.

\subsection*{Verification Scalability}
Every year, state-of-the-art neural networks blow up in size,
gaining more and more parameters.
We're talking about billions of parameters.
There is no clear end in sight.
This poses incredible challenges for verification.
Constraint-based approaches are already not very scalable,
and abstraction-based approaches tend to lose precision
with more and more operations.
So we need creative ways to make sure that verification technology
keeps up with the parameter arms race.

\subsection*{Verification Across the Stack}
Verification research has focused on checking properties of neural networks in 
isolation.
But neural networks are, almost always, a part of a bigger more complex system.
For instance, a neural network in a self-driving car receives a video stream from multiple cameras and makes decisions on how to steer, speed up, or brake.
These video streams run through layers of encoding, and the decisions made by the neural network go through actuators with their own control software and sensors.
So, if one wants to claim any serious correctness property of a neural-network-driven car, one needs to look at all of the software components together as a system.
This makes the verification problem challenging for two reasons:
(1) The size of the entire stack is clearly bigger than just the neural network, so scalability can be an issue.
(2) Different components may require different verification techniques,
e.g., abstract domains.

Another issue with verification approaches is the lack of emphasis on 
the training algorithms that produce neural networks.
For example, training algorithms may themselves not be robust:
a small corruption to the data may create vastly different neural networks.
For instance, a number of papers have shown that \emph{poisoning} the dataset
through minimal manipulation can cause a neural network to pick up on spurious correlations that can be exploited by an attacker.
Imagine a neural network that detects whether a piece of code is malware. This network can be trained using a dataset of malware and non-malware. By adding silly lines of code to some of the non-malware code in the dataset, like 
\lstinline{print("LOL")}, we can force the neural network to learn a correlation between the existence of this print statement and the fact that a piece of code is not malware~\citep{DBLP:journals/corr/abs-2006-06841}. This can then be exploited by an attacker.
This idea is known as installing a \emph{backdoor} in the neural network.

So it's important to prove that our training algorithm is not susceptible to small perturbations in the input data. This is a challenging problem, but researchers have started to look at it for simple models~\citep{drews2020proving,rosenfeld2020certified}.

\subsection*{Verification in Dynamic Environments}
Often, neural networks are deployed in a dynamic setting, where the neural network interacts 
with the environment, e.g.,  a self-driving car.
Proving correctness in this setting is rather challenging.
First, one has to understand the interaction between the neural network
and the environment---the \emph{dynamics}. This is typically hard to pin down precisely, as real-world physics may not be as clean as textbook formulas.
Further, the world can be uncertain, e.g., we have to somehow reason about other crazy drivers on the road.
Second, in such settings, one needs to verify that a neural-network-based
controller maintains the system in a safe state (e.g., on the road, no crash, etc.). This requires an inductive proof, as one has to reason about arbitrarily many time steps of control.
Third, sometimes the neural network is learning on-the-go, using \emph{reinforcement learning}, where the neural network tries things to see how the environment responds, like a toddler stumbling around. So we have to ensure that the neural network does not do stupid things as it is learning.

Recently, there have been a number of approaches attempting to verify properties of neural networks in dynamic and reinforcement-learning settings~\citep{bastani2018verifiable,zhu2019inductive,ivanov2019verisig,DBLP:conf/nips/AndersonVDC20}.

\subsection*{Probabilistic Approaches}
The verification problems we covered are hard, yes-or-no problems.
A recent approach, called \emph{randomized smoothing}~\citep{cohen2019certified,lecuyer2019certified},
has shown that one can get probabilistic guarantees, at least for some robustness properties~\citep{ye2020safer,bojchevski2020efficient}. Instead of saying a neural network is robust or not
around some input, we say it is robust with a high probability.

\printbibliography

\end{document}